\documentclass[10pt,twocolumn,letterpaper]{article}

\usepackage[pagenumbers]{cvpr} 

\usepackage{url}
\usepackage{booktabs}       
\usepackage{multirow}
\usepackage{epsfig}
\usepackage{marvosym}
\usepackage{threeparttable}
\usepackage{algorithm}
\usepackage{algorithmic}

\usepackage{bm}
\usepackage{makecell}
\usepackage{amsmath}
\usepackage{mathrsfs}
\usepackage{graphicx}
\usepackage{natbib}
\usepackage{color}
\usepackage{amsthm}
\usepackage{amsfonts}
\usepackage[most]{tcolorbox}
\usepackage{colortbl}

\newcommand{\myformer}{TimeArtist }

\newcommand{\Rmnum}[1]{\expandafter\@slowromancap\romannumeral #1@}

\definecolor{cvprblue}{rgb}{0.21,0.49,0.74}
\usepackage[pagebackref,breaklinks,colorlinks,allcolors=cvprblue]{hyperref}

\def\paperID{1599}
\def\confName{CVPR}
\def\confYear{2026}

\title{Temporal-Visual Semantic Alignment: A Unified Architecture for Transferring Spatial Priors from Vision Models to Zero-Shot Temporal Tasks}

\author{Xiangkai Ma, Han Zhang, Wenzhong Li\textsuperscript{\Letter}, Sanglu Lu\\
State Key Laboratory for Novel Software Technology, Nanjing University\\
Nanjing, China\\
{\tt\small {xiangkai.ma,zhanh}@smail.nju.edu.cn, {sanglu,lwz}@nju.edu.cn}
\thanks{
The corresponding author is Wenzhong Li.
}
}

\begin{document}


\twocolumn[{
\renewcommand\twocolumn[1][]{#1}
\vspace{-40pt}
\maketitle
\begin{center}
\vspace{-25pt}
\captionsetup{type=figure}
\includegraphics[width=0.95\linewidth]{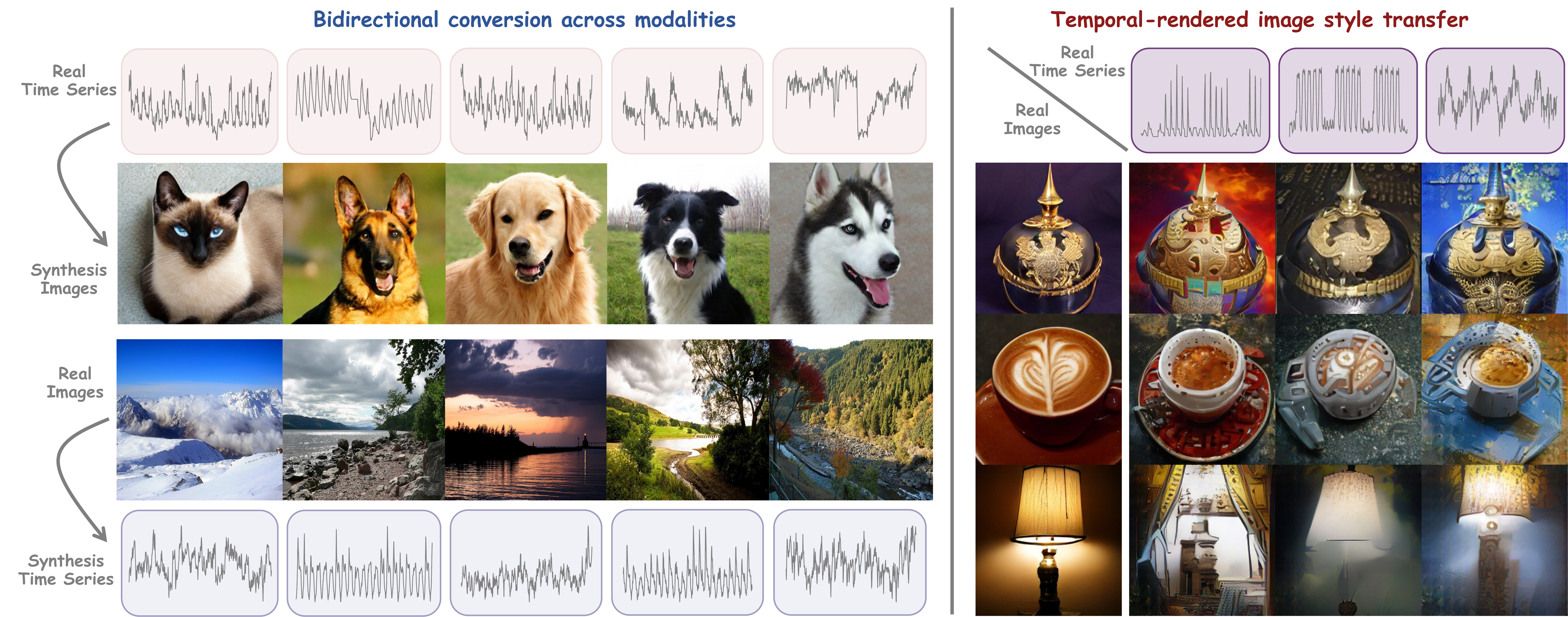}
\vspace{-1pt}
\caption{We propose \textbf{TimeArtist}, a groundbreaking temporal-visual multimodal framework that pioneers the representational alignment between temporal fluctuation and visual semantic information. \myformer establishes a bidirectional conversion pipeline between time series and image, while learning temporal patterns to render temporally-styled image.}\label{fig:abstract}
\end{center}
}]

\begin{abstract}
Large Multimodal Models (LMMs) have achieved remarkable progress in aligning and generating content across text and image modalities. 
However, the potential of using non-visual, continuous sequential, as a conditioning signal for high-fidelity image generation remains largely unexplored. 
Furthermore, existing methods that convert series into ``pseudo-images'' for temporal forecasting fail to establish semantic-level alignment.
In this paper, we propose TimeArtist, a temporal-visual conversion framework that pioneers semantic-level alignment between time series fluctuations and visual concepts. 
It pioneers a ``warmup-align'' paradigm: first, a dual-autoencoder and shared quantizer are self-supervised trained on large-scale datasets to learn modality-shared representations. Then, the encoders and quantizer are frozen, and a projection is introduced to align temporal and visual samples at the representation level.
TimeArtist establishes a versatile cross-modal framework, enabling high-quality, diverse image generation directly from time series, while capturing temporal fluctuation patterns to render images as styles transfer. 
Extensive experiments show that TimeArtist achieves satisfactory performance in image generation metrics, while also attaining superior results in zero-shot temporal tasks. Our work establishes a new paradigm for cross-modal generation, bridging the gap between temporal dynamics and visual semantics.

\end{abstract}
\vspace{-1pt}

\section{Introduction} \label{section:introduction}
The field of image generation has witnessed remarkable progress, driven by advancements in GANs~\cite{goodfellow2020gan,gui2021review}, VQVAE~\cite{van2017neural,chen2018pixelsnail,esser2021taming,yu2021vector,ramesh2021zero}, diffusion models~\cite{croitoru2023diffusion,yang2023diffusion}, and autoregressive models~\cite{xiong2024autoregressive,tian2024visual}. 
These models can synthesize high-fidelity and diverse images from various conditioning signals such as text descriptions~\cite{deng2025words} or sketches~\cite{wu2024cadvlm}. 
However, a significant and largely unexplored frontier remains: how to generate semantically meaningful images from non-visual, continuous sequential data, like time series? 
Ubiquitous in the real world, time series data, from sensor readings and physiological signals to financial charts, encodes rich dynamic patterns~\cite{jin2023time,wu2022timesnet,wang2024timemixer++} (e.g., trends, periodicity, anomalous fluctuations). 
The ability to render these temporal patterns into discernible visual concepts would unlock novel applications, such as generating cardiac imagery from ECG signals~\cite{bai2024dreamdiffusion}, visualizing market sentiment from stock volatility, or creating landscape art from climate sequences.


In addition to synthesizing images from time series, our another focus lies in leveraging pre-trained Large Vision Models (LVMs) on temporal scenarios, such as time series forecasting (TSF)~\citep{li2020forecasting,wang2024timemixer++} and time series classification (TSC)~\citep{hatami2018classification}.
Recent studies~\citep{li2023time,chen2024visionts,zhong2025time} have attempted to convert 1D time series into 2D ``pseudo-images'', via statistical methods~\citep{wang2015imaging} or periodic folding~\citep{wu2022timesnet,wang2024timemixer++}. 
Nevertheless, due to the incompatibility between the highly abstract pixel arrangements in pseudo-images and the statistical priors learned during pretraining on natural image datasets (e.g., ImageNet~\citep{deng2009imagenet}), the learned inductive biases of LVM can't be effectively transferred to TSF and TSC. This fundamental mismatch severely limits generalization capability and zero-shot performance. 
Different from previous attempts, our work focuses on generating semantically explicit temporal-image by establishing a general-purpose Temporal-Visual Conversion Framework (TVCF), enabling two critical tasks: (i) zero-shot time series analysis with pretrained vision models, and (ii) synthesizing high-quality images from time series.


The key challenge in temporal-visual conversion lies in the fundamental modality gap: unlike images, which often have natural linguistic descriptions, real-world time series data typically lack corresponding visual or textual annotations. 
This absence of aligned multimodal counterparts makes it infeasible to bridge the two domains as in the CLIP-style approaches~\citep{radford2021learning,Li2022BLIP,li2023blip,zhai2023sigmoid}, by simply minimizing representation distances between paired samples. 
To address this discrepancy, we propose constructing a unified temporal-visual latent space through a shared codebook, enabling joint representation learning. 
Within this shared space, we establish cross-modal projections to achieve fine-grained, representation-level alignment between temporal dynamics and visual semantics.

\begin{figure*}[t]
\begin{center}
	\centerline{\includegraphics[width=1.95\columnwidth]{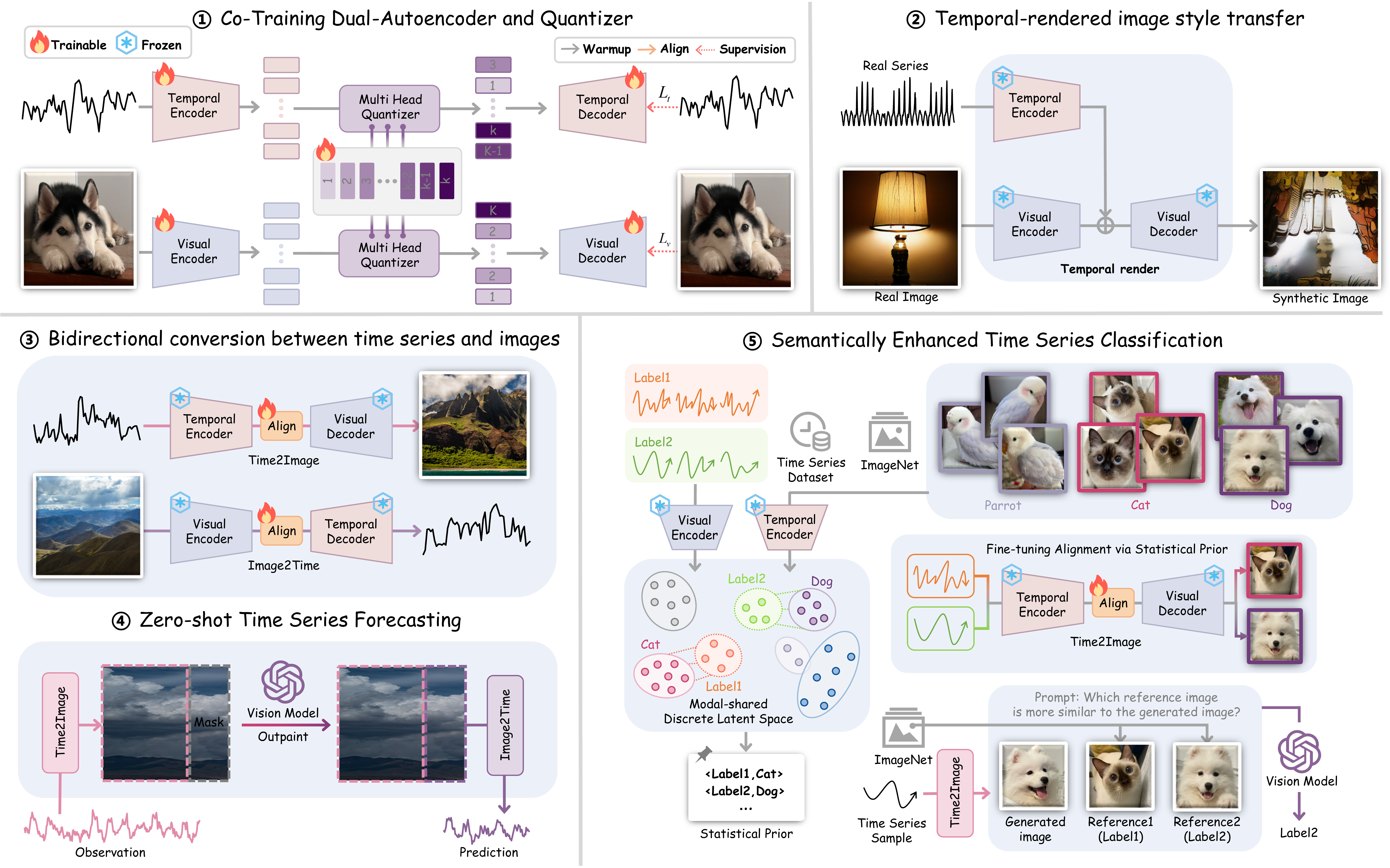}}
	\vspace{-5pt}
	\caption{The overview of \myformer. We first self-supervised train the model in large-scale temporal and visual datasets, then apply the pre-trained model directly or fine-tuned to diverse downstream tasks.}\label{fig:method1}
\end{center}
\vspace{-30pt}
\end{figure*}

To this end, we introduce \myformer to implement TVCF in Figure~\ref{fig:method1}.
We first employ self-supervised pre-training to optimize dual-encoders and shared quantizer across extensive temporal and visual domains, thereby learning discrete representations. 
Each time series and image is encoded and quantized into a sequence of quantified indices.
Benefiting from the unified discrete latent space, the hybrid architecture enables free conversion between arbitrary time series and image. 
Furthermore, to ensure the synthesized images exhibit clear visual semantics, we freeze the autoencoder and codebook while training the alignment module to establish a projection between temporal and visual modalities. 
By utilizing the shared codebook, the prior distribution of the discrete representations is learned dynamically rather than being static.
The alignment module innovatively parameterizes the cross-modal joint posterior distribution without requiring human-annotated supervision. It achieves this through the projection between the underlying distributions of code indices, thereby aligning temporal fluctuation patterns with visual semantic information at the representation level. As a result, it effectively captures dynamic feature from time series and establishes discrete representation level connections with semantic information derived from image modality. 

The advantages of the proposed \myformer are twofold:
First, \myformer overcomes the fundamental discrepancy between temporal and visual modalities through self-supervised training on extensive time series domains and large-scale image datasets (e.g., ImageNet-1k).
Second, the alignment occurs in the discrete latent space rather than the original space. This encourages the model to initially learn universal high-level semantic information from images before progressively refining fine-grained pixel features across diverse images. 
These advantages empower \myformer to produce high-quality and diverse images from time series, as shown in Figure~\ref{fig:abstract}. Moreover, the semantic-level alignment supports TimeArtist's strong cross-modal generalization and zero-shot capability in temporal scenarios.

Our contributions can be summarized as follows. 
\begin{itemize}
    \item We propose \myformer, a novel framework that projects temporal and visual modalities via representation-level alignment. This enables strong cross-modal generalization and supports zero-shot temporal tasks with LVM.
    \item We introduce a warmup-align paradigm with dual-autoencoders and shared quantizer, alongside multi-head quantization, to establish a unified representation space for temporal-visual cross-modal conversion.
    \item \myformer achieves 2.30 rFID and 3.29 gFID in image generation with 145× faster and one-step synthesis, while improving time series forecasting by 13.7\% and classification by 8.5\%.
\end{itemize}


\section{Related Work} \label{section:relatedwork}

\subsection{Vector Quantization for Visual Generation}
Vector quantization have achieved satisfactory success in the vision generation~\citep{razavi2019generating,yu2021vector,ramesh2021zero}. VQVAE~\citep{van2017neural} pioneered the VQ paradigm that encoding an image into discrete tokens sequence, followed by directly modeling the underlying distribution of token sequences via an autoregressive model~\citep{chen2018pixelsnail}. 
Recent advances in VQ tokenizers have focused on three main directions: improving reconstruction fidelity and generation quality~\citep{lee2022autoregressive}, enhancing codebook utilization~\citep{yu2021vector,zhang2023regularized,zhu2024scaling}, and exploring novel architectures such as the multi-scale VQVAE~\citep{tian2024visual} for next-scale prediction of images.
Surpassing existing methodologies, \myformer proposes an innovative Multi-Head Quantization (MHQ) strategy. For image, MHQ concurrently models high-level semantic information and low-level pixel-wise features. For time series, MHQ effectively disentangles and captures trend-seasonal components along with multi-periodic patterns. By introducing a joint representation space rather than relying on a single distribution, \myformer successfully improves codebook utilization efficiency.

\subsection{Alignment of Visual-Temporal Modalities}
Pioneering TimesNet~\citep{wu2022timesnet} and TimeMixer++~\citep{wang2024timemixer++} attempt to decompose raw sequences into periodic components and reorganize them as 2d grayscale images for forecasting. 
DreamDiffusion~\citep{bai2024dreamdiffusion} introduces pretrained CLIP~\citep{radford2021learning} supervision to facilitate multimodal alignment across EEG, text, and vision. 
Recent studies have explored leveraging the cross-modality generalization of pre-trained LMMs to enhance zero-shot time series analysis. 
For instance, TimeVLM\citep{zhong2025time} and VisionTS~\citep{chen2024visionts} employ MAE~\cite{he2022masked} to reconstruct masked portions of grayscale images, ultimately flattening the output to predict. 
However, existing researches have not yet explored to establish semantic-level feature alignment. Such pseudo images lack explicit semantic information, creating incompatibility with the LVMs' prior knowledge on real-world images, which severely limits the generalizability and zero-shot performance.

Surpassing existing approaches, \myformer pioneers the alignment between temporal fluctuations and visual semantics
This comprehensive alignment enables unprecedented capabilities, including bidirectional transformation between 1d time series and 2d image, as well as temporal-pattern-guided image rendering. Furthermore, the semantically enriched representations fully unleash the potential of LVM, significantly enhancing zero-shot temporal tasks.


\section{\myformer Architecture} \label{section:pets}
The overview of TimeArtist in Figure~\ref{fig:method1}, which consists of three key components: dual-autoencoder, multi-head quantizer, and alignment module. We designed a self-supervised ``warmup-align'' training strategy as~(\textcircled{1}). 
During the warm-up phase, we employ batch-level mixing of multimodal inputs, where each batch contains only unimodal time-series or image data. Subsequently, we compute the reconstruction error for each modality to optimize its corresponding autoencoder and the shared quantizer.
In the alignment phase, we freeze auto-encoders and quantizer while training the alignment module to learn the projection between the two discrete distributions. Specifically, model quantizes samples from diverse modalities into discrete indices distribution, then establishes projection between these cross-modal indices distributions.
After warmup-align, TimeArtist achieves seamless cross-modal conversion between temporal and visual data in \textcircled{3}.
Utilizing the conversion, we can redefinite of TSF as an outpainting task to leverage pre-trained vision models as \textcircled{4}. Furthermore, we can transform of TSC into an image classification task through converting labeled time-series data into images containing specific visual content corresponding to their respective classes in \textcircled{5}. 
Beyond temporal tasks, TimeArtist can inject temporal patterns into images within representation space to generate stylistical images in \textcircled{2}.


\subsection{Warmup-Alignment Self-Supervised Training}
\subsubsection{Dual-Autoencoder and Shared Quantizer}
We first tokenize images and time series into latent tokens, and subsequently utilize quantizer and decoder to reconstruct the original images and time series from these tokens. 

We begin by tokenizing multimodal inputs into a common latent representation space. Given an input time series \begin{small}$\mathbf{x}_{\mathrm{time}}\in\mathbb{R}^L$\end{small} of length \begin{small}$L$\end{small}, and an image \begin{small}$\mathbf{x}_{\mathrm{image}}$\end{small} of spatial dimensions \begin{small}$H\times W$\end{small}, we first transform them into sequences of latent tokens.
For the image, a patchification operation divides it into non-overlapping patches, which are then linearly embedded and flattened into a sequence of visual tokens \begin{small}$\mathbf{V}\in\mathbb{R}^{N\times D}$\end{small}, where \begin{small}$N=\frac{H}{f}\times\frac{W}{f}$\end{small} is the number of patches (with downsampling factor \begin{small}$f$\end{small}) and \begin{small}$D$\end{small} is the embedding dimension. 
For the time series, it is partitioned into \begin{small}$P_L$\end{small} segments of fixed length \begin{small}$l$\end{small} (\begin{small}$P_L\times l=L$\end{small}), and these segments are transformed into temporal tokens \begin{small}$\mathbf{T}\in\mathbb{R}^{N\times D}$\end{small}. 
The use of \begin{small}$N$\end{small} ensures the output sequence length is consistent across modalities for subsequent processing.
These token sequences are processed by separate encoders: a visual encoder \begin{small}$\varepsilon_{\mathrm{vis}}$\end{small} and a temporal encoder \begin{small}$\varepsilon_{\mathrm{temp}}$\end{small}. They produce continuous latent embedding:
\begin{small}
\begin{equation}
\vspace{-5pt}
  \begin{split}
    \mathbf{e}_v=\mathcal{E}_\mathrm{vis}(\mathbf{V})\in\mathbb{R}^{N\times D},~\mathbf{e}_t=\mathcal{E}_\mathrm{temp}(\mathbf{T})\in\mathbb{R}^{N\times D}.
  \end{split}
\vspace{-15pt}
\end{equation}
\end{small}

A shared multi-head vector quantizer \begin{small}$\mathcal{Q}$\end{small} then maps these continuous embeddings into a discrete latent space. 
This quantizer projects \begin{small}$\mathbf{e}_{v}$\end{small} and \begin{small}$\mathbf{e}_{t}$\end{small} into quantized representations \begin{small}$\mathbf{q}_{v}$\end{small} and \begin{small}$\mathbf{q}_{t}$\end{small}, both in \begin{small}$\mathbb{R}^{N\times D}$\end{small}.
These discrete codes serve as the bridge between modalities. Finally, dedicated decoders reconstruct the original inputs from these codes:
\begin{small}
\begin{equation}
\vspace{-5pt}
  \begin{split}
    \hat{\mathbf{x}}_{\mathrm{image}}=\mathcal{D}_{\mathrm{vis}}(\mathbf{q}_{v}),~\hat{\mathbf{x}}_{\mathrm{time}}=\mathcal{D}_{\mathrm{temp}}(\mathbf{q}_{t}).
  \end{split}
  \vspace{-15pt}
\end{equation}
\end{small}

This dual-autoencoder structure is trained on large-scale, unpaired time series and image datasets (e.g., Monash~\cite{godahewa2021monash}, ImageNet~\cite{deng2009imagenet}) to learn robust, modality-invariant features.

\subsubsection{Multi-Head Quantization (MHQ)}
Given that both temporal and visual data exhibit hierarchical representational structures~\citep{redmon2018yolov3,wang2024timemixer++,liu2022pyraformer}, we propose a Multi-Head Quantization (MHQ) strategy employing multi-head codebook to model discrete representation spaces at varying scales. Specifically, the quantizer independently retrieves code vectors from each codebook and concatenates them together. 
The MHQ architecture yields several fundamental advantages: First, it significantly improves codebook utilization by introducing a joint representation distribution that enables potential combinations of multi-scale features. 
For instance, diverse codebooks allow the model to simultaneously decompose and learn high-level semantic information and fine-grained pixel features from images, as well as capture trend, seasonal, and multi-periodic patterns in time series. 
This strategy substantially enhances reconstruction fidelity for both modalities, and ensures the generation of diverse synthetic images. 
MHQ incorporates multiple heads that learn temporal and visual features at diverse scales, improving utilization rate by introducing the joint representation space rather than a single distribution.

Instead of a single codebook, MHQ employs \begin{small}$M$\end{small} independent codebooks \begin{small}$\mathcal{Q}=\{\mathcal{Q}_m\}_{m=1}^M$\end{small}, each representing a different scale or aspect of the data.
Each codebook \begin{small}$\mathcal{Q}_m$\end{small} contains \begin{small}$K$\end{small} codes \begin{small}$\frac{D}{M}$\end{small}: \begin{small}$\mathcal{Q}_{m}=\{\mathbf{z}_{m,k}\in\mathbb{R}^{\frac{D}{M}}\}_{k=1}^{K}$\end{small}. 
The continuous latent embedding \begin{small}$\mathbf{e}\in\mathbb{R}^{N\times D}$\end{small} (from either encoder) is split along the feature dimension into \begin{small}$M$\end{small} sub-vectors \begin{small}$\{\mathbf{e}_m\in\mathbb{R}^{N\times \frac{D}{M}}\}_{m=1}^M$\end{small}. 
For each head \begin{small}$m$\end{small}, a nearest-neighbor search is performed within its corresponding codebook \begin{small}$\mathcal{Q}_{m}$\end{small}:
\begin{small}
\begin{equation}
\vspace{-5pt}
  \begin{split}
    \mathbf{q}_{m,i}=\arg\min_{\mathbf{z}_{m,k}\in\mathcal{Q}_{m}}\parallel\mathbf{e}_{m,i}-\mathbf{z}_{m,k}\parallel_{2},
  \end{split}
  \vspace{-15pt}
\end{equation}
\end{small}
where \begin{small}$\mathbf{q}_{m,i}$\end{small} and \begin{small}$\mathbf{e}_{m,i}$\end{small} denote the i-th token of \begin{small}$\mathbf{q}_{m}$\end{small} and \begin{small}$\mathbf{e}_{m}$\end{small}, respectively. 
The final quantized representation \begin{small}$\mathbf{q}$\end{small} is formed by concatenating the outputs from all heads:
\begin{small}
\begin{equation}
\vspace{-5pt}
  \begin{split}
    \mathbf{q}=[\mathbf{q}_1;\mathbf{q}_2;...;\mathbf{q}_M]\in\mathbb{R}^{N\times D}.
  \end{split}
  \vspace{-15pt}
\end{equation}
\end{small}
This architecture enhances codebook utilization and reconstruction fidelity by allowing the model to simultaneously capture coarse semantic concepts and fine-grained details.

\subsubsection{Optimization and Training Objective}
During the warmup phase, the entire system, the dual-autoencoder and shared-quantizer, is trained end-to-end using a combination of reconstruction and quantization losses. The total loss for each modality is:
\begin{small}
\begin{equation}
  \begin{split}
    \mathcal{L}_{\mathrm{rec}}^{\mathrm{temp}}\text{=}\underbrace{\|\mathbf{x}_{\mathrm{time}}\text{-}\hat{\mathbf{x}}_{\mathrm{time}}\|_{2}^{2}}_{\text{Reconstruction}}\text{+}\underbrace{\|\mathrm{sg}[\mathbf{e}_{t}]\text{-}\mathbf{q}_{t}\|_{2}^{2}}_{\text{Quantization Loss}}\text{+}\underbrace{\|\mathrm{sg}[\mathbf{q}_{t}]\text{-}\mathbf{e}_{t}\|_{2}^{2}}_{\text{Commitment Loss}}, \\
    \mathcal{L}_{\mathrm{rec}}^{\mathrm{vis}}\text{=}\|\mathbf{x}_{\mathrm{image}}\text{-}\hat{\mathbf{x}}_{\mathrm{image}}\|_2^2\text{+}\|\mathrm{sg}[\mathbf{e}_v]\text{-}\mathbf{q}_v\|_2^2\text{+}\|\mathrm{sg}[\mathbf{q}_v]\text{-}\mathbf{e}_v\|_2^2.
  \end{split}
\end{equation}
\end{small}
Here, \begin{small}$\operatorname{sg}[\cdot]$\end{small} denotes the stop-gradient operator, \begin{small}$\|\mathbf{x}_\mathrm{time}-\hat{\mathbf{x}}_\mathrm{time}\|_2^2$\end{small} and \begin{small}$\|\mathbf{x}_\mathrm{image}-\hat{\mathbf{x}}_\mathrm{image}\|_2^2$\end{small} as reconstruction loss to optimize dual-autoencoder.
The gradients are backpropagated through the non-differentiable quantization step using the straight-through estimator~\citep{bengio2013estimating}. 
To learn the embedding space, the quantization loss \begin{small}$\|\mathrm{sg}[\mathbf{e}_v]-\mathbf{q}_v\|_2^2$\end{small} is utilized to fit the embedding vector \begin{small}$\mathbf{q}_{v}$\end{small} towards the encoder output, and \begin{small}$\|\mathrm{sg}[\mathbf{q}_{v}]-\mathbf{e}_{v}\|_{2}^{2}$\end{small} is the commitment los~\citep{van2017neural}.
The total training objective is \begin{small}$\mathcal{L}=\mathcal{L}_{\mathrm{rec}}^{\mathrm{temp}}+\mathcal{L}_{\mathrm{rec}}^{\mathrm{vis}}$\end{small}.

\subsubsection{Alignment from Fluctuation to Semantics}
Upon completion of the warmup phase, the dual encoders and the shared multi-head quantizer are frozen, having established a unified discrete latent space. 
In this space, both time series and images are encoded and quantized into temporal and visual discrete index probability distributions.
We define this learned set of encoded vectors as the multimodal joint prior distribution. 
Following the paradigm of MaskGIT~\citep{chang2022maskgit} modeling visual probability distributions, we train an additional alignment module to establish a projection between temporal-visual distribution.

Formally, let \begin{small}$\mathbf{e}_{v}$\end{small} and \begin{small}$\mathbf{e}_{t}$\end{small} be the continuous latent embeddings from the visual and temporal encoders. 
The multi-head quantizer \begin{small}$\mathcal{Q}=\{\mathcal{Q}_m\}_{m=1}^M$\end{small} maps these embeddings to quantized codes \begin{small}$\mathbf{q}_{v}$\end{small} and \begin{small}$\mathbf{q}_{t}$\end{small}.
Critically, each token in \begin{small}$\mathbf{q}_{v}$\end{small} and \begin{small}$\mathbf{q}_{t}$\end{small} is a direct lookup from the same set of codebooks, which induces a joint prior distribution over the two modalities.

We then extract the corresponding index sequences \begin{small}$\mathbf{I}_v,\mathbf{I}_t\in\mathbb{Z}^N$\end{small}, by recording the codebook indices selected during quantization. Specifically, for the \begin{small}$m$\end{small}-th head and the \begin{small}$i$\end{small}-th token, the index is:
\begin{small}
\begin{equation}
  \begin{split}
    (\mathbf{I}_{v,m})_i=\arg\min_{k\in[1,K]}\|(\mathbf{e}_{v,m})_i-\mathbf{z}_{m,k}\|_2,
  \end{split}
\end{equation}
\end{small}
with an analogous definition for \begin{small}$(\mathbf{I}_{t,m})_i$\end{small}.

The alignment phase aims to learn a bidirectional mapping between these index sequences. 
We introduce a transformer-based alignment module \begin{small}$\mathcal{A}$\end{small}, which is trained to predict visual indices from temporal ones, and vice versa:
\begin{small}
\begin{equation}\label{eq:A}
  \begin{split}
    \mathbf{I}_v=\mathcal{A}(\mathbf{I}_t)~\mathrm{and}~\mathbf{I}_t=\mathcal{A}^{-1}(\mathbf{I}_v).
  \end{split}
\end{equation}
\end{small}

By operating directly on the discrete indices, \begin{small}$\mathcal{A}$\end{small} learns a cross-modal joint posterior distribution. 
This representation-level alignment effectively bridges temporal fluctuation patterns (e.g., trends and periodicities) with visual semantic concepts (e.g., objects and textures), enabling high-fidelity conversion.

\subsection{Zero-shot TSF and TSC with Vision Models}
Benefiting from the seamless cross-modal conversion capability, we can effectively leverage the prior knowledge acquired by vision models during pre-training. 
We first introduce the bidirectional transformation mechanism between time series and image, followed by an explanation of how time series forecasting (TSF) and time series classification (TSC) can be ``translated'' into visual tasks.

\subsubsection{Bidirectional conversion across modalities}
In cross-modal bidirectional conversion, we reorganize the TimeArtist's hybrid architecture and data flow, establishing a flexible bidirectional conversion pipeline between 1d time series \begin{small}$x^{time}$\end{small} and 2d images \begin{small}$x^{image}$\end{small}, as:
\begin{small}
\begin{equation}\label{eq4}
  \begin{split}
    \hat{\mathbf{x}}_{\mathrm{image}}=\mathcal{D}_{\mathrm{vis}}\circ\mathcal{Q}\circ\mathcal{E}_{\mathrm{temp}}(\mathbf{x}_{\mathrm{time}}), \\
    \hat{\mathbf{x}}_{\mathrm{time}}=\mathcal{D}_{\mathrm{temp}}\circ\mathcal{Q}\circ\mathcal{E}_{\mathrm{vis}}\left(\mathbf{x}_{\mathrm{image}}\right).
  \end{split}
\end{equation}
\end{small}

\subsubsection{Zero-shot TSF with outpainting}
By aligning at the semantic level, \myformer constructs a projection between temporal fluctuation segments and objects in landscape paintings, thereby bridging TSF with visual outpainting. 
Our key insight focus that critical fluctuation patterns in the observation series are aligned with landscape objects (e.g., rivers, mountains, and clouds), where the composition of outpainted objects sharing consistent stylistic features is subsequently converted back to the temporal modality, as shown in Figure~\ref{fig:method1} \textcircled{2}.
Concretely, the observed series is first projected into discrete representation space and subsequently decoded directly into landscape imagery. 
We then employ pretrained LVMs to outpaint that maintains the style of the synthesized landscape. 
Finally, the results of outpainting are inversely transformed to yield predicted series. 
Leveraging the robust semantic alignment, we achieve zero-shot forecasting with arbitrary lengths. Details in B.1 of appendix.

\subsubsection{Semantically enhanced TSC}
Let \begin{small}$\{\mathcal{T}^{(m)}\}_{m=1}^{M}$\end{small} and \begin{small}$\{\mathcal{V}^{(n)}\}_{n=1}^N$\end{small} denote \begin{small}$M$\end{small} time series subsets amd \begin{small}$N$\end{small} image subsets.
Each sample is encoded into the shared discrete latent space via frozen encoders and quantizer, as:
\begin{small}
\begin{equation}
  \begin{split}
    \mathbf{e}_t=\mathcal{E}_{\mathrm{temp}}(\mathbf{x}_{\mathrm{time}}),~\mathbf{e}_v=\mathcal{E}_\mathrm{vis}(\mathbf{x}_\mathrm{image~}).
  \end{split}
\end{equation}
\end{small}
hese embeddings are then quantized via the shared multi-head quantizer to obtain index sequences \begin{small}$\mathbf{I}_{t}$\end{small} and \begin{small}$\mathbf{I}_{v}$\end{small}.

For each subset \begin{small}$\mathcal{T}^{(m)}$\end{small}, we estimate its empirical discrete distribution \begin{small}$\hat{p}_t^{(m)}$\end{small} over the joint index space \begin{small}$\{1,\ldots,K\}^{M}$\end{small} by aggregating the quantized indices of all its samples across all quantization heads and token positions. Analogously, we construct \begin{small}$\hat{p}_{v}^{(n)}$\end{small} for each image subset \begin{small}$\mathcal{V}^{(n)}$\end{small}.

We then compute the Jensen–Shannon divergence~\cite{fuglede2004jensen} between every pair of temporal and visual distributions:
\begin{small}
\begin{equation}
  \begin{split}
    C_{mn}=\mathcal{D}_{\mathrm{JS}}\left(\hat{p}_{t}^{(m)}\parallel\hat{p}_{v}^{(n)}\right).
  \end{split}
\end{equation}
\end{small}
This yields a cost matrix \begin{small}$\mathbf{C}\in\mathbb{R}^{M\times N}$\end{small}. To establish a semantically coherent one-to-one correspondence between temporal classes and visual labels, we solve the following optimization problem:
\begin{small}
\begin{equation}
  \begin{split}
    \min_{\Pi\in\mathcal{P}}\sum_{m=1}^M\sum_{n=1}^N\Pi_{mn}C_{mn},
  \end{split}
\end{equation}
\end{small}
where \begin{small}$\mathcal{P}=\left\{\Pi\in\{0,1\}^{M\times N}|\sum_{n=1}^{N}\Pi_{mn}=1,\sum_{m=1}^{M}\Pi_{mn}\leq1\right\}$\end{small} enforces an injective mapping from temporal classes to visual labels. This assignment problem is solved optimally using the Hungarian algorithm~\cite{mills2007hungarian}, yielding an optimal binary mapping \begin{small}$\Pi^{*}$\end{small}.
We then construct a paired dataset:
\begin{small}
\begin{equation}
  \begin{split}
    \mathcal{D}_{\mathrm{align}}=\bigcup_{m=1}^{M}\left\{(x_{\mathrm{time}},x_{\mathrm{image}})|\Pi_{mn}^{*}=1\right\}.
  \end{split}
\end{equation}
\end{small}
This dataset is used to fine-tune the alignment \begin{small}$\mathcal{A}$\end{small} to learn the cross-modal index mapping as in Eq.~\ref{eq:A}. 
Once trained, any time series \begin{small}$\mathbf{x}_{\mathrm{time}}$\end{small} is converted into an image via:
\begin{small}
\begin{equation}
  \begin{split}
    \hat{\mathbf{x}}_{\mathrm{image}}=\mathcal{D}_{\mathrm{vis}}\circ \mathcal{A}\circ\mathcal{E}_{\mathrm{temp}}(\mathbf{x}_{\mathrm{time}})..
  \end{split}
\end{equation}
\end{small}
The resulting image is classified by a frozen vision model using a prompt-based template that compares it against reference images from the \begin{small}$M$\end{small} visual categories selected by \begin{small}$\Pi^{*}$\end{small}, thereby inferring the original time series class.


\subsection{Training Details}
TimeArtist introduces a unified cross-modal architecture that employs a warmup-align strategy: 

\subsubsection{Warmup Strategy}
During the warmup phase, we self-supervised train dual-autoencoder and quantizer to capture universal visual-temporal patterns. 
Specifically, we utilizing independent time series Monash~\cite{godahewa2021monash} datasets and visual ImageNet~\cite{deng2009imagenet} datasets train the system. 
As illustrated in Figure 7 of appendix, this stage does not require paired time-series and image data.
These pre-trained modules remain frozen in subsequent stages, enabling the following alignment phase to focus exclusively on task-specific semantic alignment without relearning foundational features, thereby ensuring both parameter efficiency and task adaptability. 

\subsubsection{Align Strategy for Forecasting and Classification}
We adopted distinct alignment strategies for diverse temporal tasks. In zero-shot forecasting, we aligned time-series data from the Monash~\cite{godahewa2021monash} repository with web-collected landscape images. Specifically, we employed sliding windows to synchronously segment sub-series and sub-images, subsequently training the alignment module on the constructed paired dataset as illustrated in Appendix Figures 8 and 9. The model's zero-shot performance was ultimately evaluated on unseen time-series datasets, with rigorous dataset partitioning implemented to prevent any information leakage.
For classification tasks, we compute the spatial distance between the temporal distribution and visual distribution in the latent space. Based on these statistical priors, we subsequently identify an image subset with designated labels for each temporal category subset that minimizes the spatial distance between these paired subsets. Here, time series from UEA classification archive~\cite{bagnall2018uea}, while images from the ImageNet~\cite{deng2009imagenet}. Finally, we train alignment on this paired dataset, and evaluate on UEA benchmarks.

The model was trained using 16 A100 (80G) GPUs, with computational time requirements of 45 hours for the warm-up phase, 12 hours for forecasting alignment, and 4 hours for classification alignment. Details in B.1 of appendix.



\section{Experiments} \label{section:experiments}

\subsection{Experimental Setup}
As a cross-modal framework, we employ large-scale time series and image datasets to train \myformer. 
For visual inputs, we select ImageNet-1k as train set, with an image resolution of H=W=256, a downsampling rate of f=16, and a codebook containing 4096 discrete codes. 
For temporal inputs, we adhere to the benchmarking protocol established by Chronos~\citep{ansari2024chronos}, utilizing 28 datasets comprising 890K univariate time series with a total of 84B observed tokens as train set. 
The zero-shot TSF performance is evaluated on the widely adopted Monash benchmark~\citep{godahewa2021monash} encompassing 29 datasets, while both zero-shot and full-shot TSF capabilities are assessed on 6 TFB benchmarks~\citep{qiu2024tfb}. 
To ensure experimental fairness and reproducibility, we follow VisionTS's data configuration, including context length and dataset split protocols. 
We strictly adhere to the configuration of FoundTS~\citep{qiu2024foundts} and refrain from the ``Drop Last'' trick to ensure result fairness.
Comprehensive details regarding baseline and benchmark configurations are provided in the appendix B.

\subsection{Visual and Temporal Generation}
Based on bidirectional conversion, \myformer can sample visual distributions from real-world time series and directly reconstruct high-quality images without requiring a dedicated model to explicitly model visual discrete distributions. 
Similarly, \myformer can also generate time series from images, thereby establishing an efficient dual-modal generative model.
To thoroughly assess the model, evaluation includes both time series generation metrics Context-FID Score~\citep{yuan2024diffusion}, and image generation metrics FID~\citep{heusel2017gans}. 
Details of metrics in the appendix B.6.

\begin{table}[ht]
  \caption{The Performance comparison of \myformer and other baselines on visual reconstruction and generation.}\label{tab:task1_visual}
  \vspace{-8pt}
  \centering
  \resizebox{1.0\columnwidth}{!}{
  \begin{small}
  \renewcommand{\multirowsetup}{\centering}
  \tabcolsep=0.2cm
  \renewcommand\arraystretch{1.1}
  \begin{tabular}{cccccccc}
    \toprule
    & tokens & codes & rFID$\downarrow$ & P$\downarrow$ & S$\downarrow$ & T$\uparrow$ & gFID$\downarrow$ \\
    \midrule

    LDM-4 & 1024 & 16k & 1.14 & 400M & 250 & 0.3 & 3.60 \\
    DiT-XL & - & - & - & 675M & 150 & 0.4 & 2.27 \\
    VQGAN & 256 & 1k & 7.94 & 1.4B & 256 & 5.8 & 15.78 \\
    RQVAE & 256 & 16k & 3.20 & 3.8B & 64 & 7.5 & 7.55 \\
    MaskGIT & 256 & 1k & 2.28 & 177M & 8 & 35.9 & 6.18 \\
    VIM & 1024 & 8k & 1.28 & 1.7B & 1024 & 0.2 & 4.17 \\

    \rowcolor{lightgray}
    \myformer & 64 & 4k & 2.30 & - & 1 & 43.7 & 3.29 \\

    \bottomrule
  \end{tabular}
  \end{small}
}
\end{table}

We summarize all results on ImageNet-1K generation benchmark of resolution \begin{small}$256\times256$\end{small} in Table \ref{tab:task1_visual}. 
Here, LDM-4 and DiT-X are diffusion-based generative models, while the remaining approaches are autoregressive-based generative models. 
Specifically, rFID and gFID denote FID for reconstruction and generation; ``tokens’’ and ``codes’’ represent the number and dimension of code vector; ``P’’ indicates the generator’s parameters; ``S’’ refers to sampling steps in generation; and ``T’’ measures throughput, expressed as samples per seconds on A100. 
\myformer achieves generation quality (gFID) comparable to other VQ models without requiring any additional generator training. 
By directly reconstructing images from time series, \myformer attains an rFID of 2.30, on par with the well-trained MaskGIT (rFID 2.28), while using only one-eighth of the parameter scale. 
Compared to other diffusion models, \myformer also delivers competitive performance while supporting 1-step image generation rather than multi-step denoising through its dual-autoencoder architecture. 
Specifically, \myformer achieves superior gFID (3.29 $vs.$ 3.60) to LDM-4~\citep{rombach2022high} while generating $145\times$ faster (43.7 $vs.$ 0.3 samples/s). 
Visualization is in Figure 10 and 11 of appendix F.

\begin{figure}[t]
\begin{center}
\centerline{\includegraphics[width=1.0\columnwidth]{images/exper7.pdf}}
\vspace{5pt}
\centerline{\includegraphics[width=1.0\columnwidth]{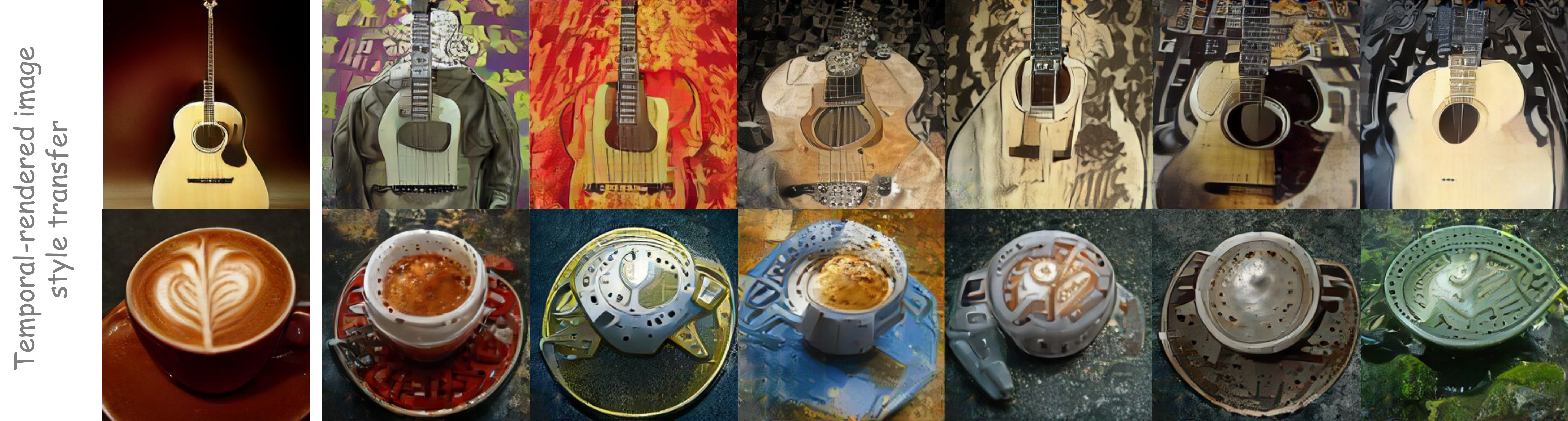}}
\vspace{-5pt}
\caption{
The visualization demonstrates bidirectional cross-modal synthesis between time-series data and images.
}\label{fig:exper7}
\end{center}
\vspace{-20pt}
\end{figure}

\begin{table}[ht]
  \vspace{-2pt}
  \caption{Comparison on time series generation. \myformer achieves SOTA performance across all evaluations.}\label{tab:task1_temporal1}
  \vspace{-8pt}
  \centering
  \resizebox{1.0\columnwidth}{!}{
  \begin{threeparttable}
  \begin{small}
  \renewcommand{\multirowsetup}{\centering}
  \tabcolsep=0.8mm
  \renewcommand\arraystretch{1.0}
  \begin{tabular}{ccccccccc}
  \toprule
  \multicolumn{1}{c}{Dataset} &
  \multicolumn{1}{c}{Metric} &
  \multicolumn{1}{c}{{{\myformer}}} & 
  \multicolumn{1}{c}{{DiffusionTS}} & 
  \multicolumn{1}{c}{{TimeGAN}} & 
  \multicolumn{1}{c}{{TimeVAE}} & 
  \multicolumn{1}{c}{{Diffwave}} & 
  \multicolumn{1}{c}{{DiffTime}} \\
  
  \hline
  \multirow{4}{*}{ETTh} 
  
  & Context-FID &
  \textbf{0.318} & 0.423 & 5.872 & 0.826 & 2.899 & 3.524 \\
  

  & Correlational & 
  \textbf{0.047} & 0.064 & 0.522 & 0.046 & 0.199 & 0.135 \\
  
  
  & Discriminative & 
  \textbf{0.041} & 0.060 & 0.442 & 0.178 & 0.304 & 0.243 \\
  

  & Predictive &
  \textbf{0.076} & 0.109 & 0.220 & 0.110 & 0.132 & 0.118 \\
  
  \hline
  \multirow{4}{*}{Energy} 
  
  & Context-FID &
  \textbf{0.098} & 0.126 & 5.032 & 3.768 & 5.572 & 4.735 \\
  
  
  & Correlational &
  \textbf{0.317} & 0.361 & 4.487 & 1.279 & 5.690 & 1.800 \\
  
  
  & Discriminative &
  \textbf{0.228} & 0.290 & 0.499 & 0.499 & 0.499 & 0.437 \\
  
  
  & Predictive & 
  \textbf{0.192} & 0.245 & 0.351 & 0.353 & 0.251 & 0.251 \\

  \bottomrule
  \end{tabular}
  \end{small}
  \end{threeparttable}
}
\vspace{-2pt}
\end{table}



    

    
    
    
    
    


Refer to the generative metrics in DiffusionTS~\citep{yuan2024diffusion} to design experiments for evaluating the performance of the proposed \myformer on the temporal generation. 
The comparison of \myformer and other diffusion baselines on ETTh and Energy benchmarks~\citep{yuan2024diffusion} are shown in table \ref{tab:task1_temporal1}.
Specifically, the four generation metrics are reduced by 24.8\%, 26.6\%, 31.7\% and 30.3\% on ETTh dataset compared to the state-of-the-art approach, which demonstrates the superior prediction capability of the proposed method.
Besides, compared with DiffusionTS, the average improvement of all indicators of \myformer on ETTh and Energy datasets is 28.4\% and 19.4\%, respectively.
Details are in appendix D.

\subsection{Visualization of temporal-style image, generated images and time series}
Through cross-modal transformation, we present visualizations of synthetic images and generated time series in Figures~\ref{fig:exper7}.
More visualization of generated images is in Figure 10 and 11 of appendix F.
More visualization of Temporal-rendered image is in Figure 16, 17 and 18 of appendix F.

\begin{table*}[ht]
  \vspace{-5pt}
  \caption{
Comparison of the performance of zero-shot and few-shot results on the long-term TSF benchmark. Results are averaged across prediction lengths {96, 192, 336, 720}. The complete results are presented in Table 10 of appendix.}\label{tab:task3_zero_longtsf}

  \vspace{-8pt}
  \centering
  \resizebox{2.0\columnwidth}{!}{
  \begin{small}
  \renewcommand{\multirowsetup}{\centering}
  \tabcolsep=0.06cm
  \renewcommand\arraystretch{0.9}

\begin{tabular}{ccccccccccccccccccccccccccccccccccccc}
\toprule
\multicolumn{1}{c}{} & \quad\quad & 
\multicolumn{17}{c}{ {Zero-Shot}} & \quad\quad & 
\multicolumn{17}{c}{ {Few-Shot (10\% Downstream Dataset)}} \\

\cmidrule{3-19}\cmidrule{21-37}

\multicolumn{1}{c}{ {Pretrain}} & & 
\multicolumn{8}{c}{\textit{ {Images}}} & & 
\multicolumn{8}{c}{{ {\textit{Time-series}}}} & & 
\multicolumn{2}{c}{\textit{ {Images}}} & & 
\multicolumn{5}{c}{{ {\textit{Text}}}} & & 
\multicolumn{8}{c}{ {\textit{No Pretrain}}} \\

\cmidrule{3-10}\cmidrule{12-19}\cmidrule{21-22}\cmidrule{24-28}\cmidrule{30-37}

\multicolumn{1}{c}{ {Method}} & & 
\multicolumn{2}{c}{ {\myformer}} & & 
\multicolumn{2}{c}{ {VisionTS}} & & 
\multicolumn{2}{c}{ {TimeVLM}} & & 
\multicolumn{2}{c}{ {Moirai(S)}} & & 
\multicolumn{2}{c}{ {Moirai(B)}} & & 
\multicolumn{2}{c}{ {Moirai(L)}} & & 
\multicolumn{2}{c}{ {TimeVLM}} & & 
\multicolumn{2}{c}{ {TimeLLM}} & & 
\multicolumn{2}{c}{ {GPT4TS}} & & 
\multicolumn{2}{c}{ {DLinear}} & & 
\multicolumn{2}{c}{ {PatchTST}} & & 
\multicolumn{2}{c}{ {TimesNet}} \\

\multicolumn{1}{c}{ {Metric}} & & MSE & MAE & & MSE & MAE & & MSE & MAE & & MSE & MAE & & MSE & MAE & & MSE & MAE & & MSE & MAE & & MSE & MAE & & MSE & MAE & & MSE & MAE & & MSE & MAE & & MSE & MAE \\

\midrule

ETTh1
     & & \textbf{0.379} & \textbf{0.405} &      &  {0.390} &  {0.414} & & 0.496 & 0.480 &      & 0.400  & 0.424  &      & 0.434  & 0.439  &      & 0.510  & 0.469  &   & 0.431  & 0.422  &   & 0.556  & 0.522  &      & 0.590  & 0.525  &      & 0.691  & 0.600  &      & 0.633  & 0.542  &      & 0.869  & 0.628  \\
\midrule

ETTh2
     & & \textbf{0.308} & \textbf{0.362} &      &  {0.333} &  {0.375} & & 0.346 & 0.391 &      & 0.341  & 0.379  &      & 0.346  & 0.382  &      & 0.354  & 0.377  &   & 0.361  & 0.405  &   & 0.370  & 0.394  &      & 0.397  & 0.421  &      & 0.605  & 0.538  &      & 0.415  & 0.431  &      & 0.479  & 0.465  \\
\midrule

ETTm1
     & & \textbf{0.367} & \textbf{0.365} &      &  {0.374} &  {0.372} & & 0.432 & 0.426 &      & 0.448  & 0.410  &      & 0.382  & 0.388  &      & 0.390  & 0.389  &   & \textbf{0.360}  & 0.382  &   & 0.404  & 0.427  &      & 0.464  & 0.441  &      & 0.411  & 0.429  &      & 0.501  & 0.466  &      & 0.677  & 0.537  \\
\midrule

ETTm2
     & & \textbf{0.264} & \textbf{0.291} &      & 0.282  & 0.321 & & 0.284 & 0.340  &      & 0.300  & 0.341  &      &  {0.272} & 0.321  &      & 0.276  &  {0.320} &   & \textbf{0.263}  & 0.323  &   & 0.277  & 0.323  &      & 0.293  & 0.335  &      & 0.316  & 0.368  &      & 0.296  & 0.343  &      & 0.320  & 0.353  \\
\midrule

Electricity
     & & 0.193 & \textbf{0.272} &      & 0.207  & 0.294 & & 0.243 & 0.327  &      & 0.233  & 0.320  &      & \textbf{0.188}  & 0.274  &      & 0.188  & 0.273  &   & 0.198  & 0.291  &   &  \textbf{0.175} & 0.270  &      & 0.176  &  \textbf{0.269} &      & 0.180  & 0.280  &      & 0.180  & 0.273  &      & 0.323  & 0.392  \\
\midrule

Weather
     & & \textbf{0.221} & \textbf{0.249} &      & 0.269  & 0.292 & & 0.281 & 0.304  &      & 0.242  & 0.267  &      & 0.238  &  {0.261} &      & 0.260  & 0.275  &   & 0.233  & 0.274  &   &  {0.234} & 0.273  &      & 0.238  & 0.275  &      & 0.241  & 0.283  &      & 0.242  & 0.279  &      & 0.279  & 0.301  \\
\midrule
Average 
     & & \textbf{0.288} & \textbf{0.324} &      &  {0.309} & 0.345 & & 0.347 & 0.378  &      & 0.327  & 0.357  &      & 0.310  &  {0.344} &      & 0.329  & 0.350  &   & 0.307  & 0.352  &   & 0.336  & 0.368  &      & 0.360  & 0.378  &      & 0.407  & 0.416  &      & 0.378  & 0.389  &      & 0.491  & 0.446  \\
\bottomrule
\end{tabular}%
  \end{small}
}
\vspace{-10pt}
\end{table*}

\begin{table}[ht]
\caption{
Comparison of computational costs (s) for forward propagation in the TSF benchmarks during the inference
.
}\label{tab:task3_computation_cost}
\vspace{-8pt}
\centering
\resizebox{1.0\columnwidth}{!}{
\begin{small}
\renewcommand{\multirowsetup}{\centering}
\tabcolsep=0.05cm
\renewcommand\arraystretch{1.0}
\begin{tabular}{cccccccc}
\toprule
ContextLen & \myformer & VisionTS & LLMTime & Moirai(B) & GPT4TS & DeepAR & PatchTST \\
\midrule
1k & 0.10 & 0.07 & $>$200 & 0.05 & 0.02 & 0.44 & 0.02 \\
2k & 0.11 & 0.07 & $>$200 & 0.06 & 0.05 & 5.92 & 0.03 \\
3k & 0.11 & 0.08 & $>$200 & 0.07 & 0.06 & 8.37 & 0.05 \\

\bottomrule
\end{tabular}
\end{small}
}
\vspace{-5pt}
\end{table}
\subsection{Zero-shot TSF with Outpainting}
Table~\ref{tab:task3_zero_longtsf} demonstrates that \myformer achieves superior forecasting performance in nearly all scenarios. 
Compared to TimeLLM and LLMTime which leverage language models with few-shot (10\%) fine-tuning, \myformer demonstrates a relative average MSE reduction ranging from 14.3\% to 20\%, indicating that visual-to-temporal knowledge transfer proves more effective than textual approaches. 
Against VisionTS which similarly employs vision models, \myformer achieves 6.7\% reduction in MSE. 
Results validate that representation-level alignment effectively activates vision models' cross-domain generalization and zero-shot.

We evaluated the computational cost at diverse prediction lengths with fixed horizon 1024 (1k). Table~\ref{tab:task3_computation_cost} presents the averaged results from 90 runs on NVIDIA A100. 
By converting to fixed-size image, \myformer achieves $O(1)$ complexity and demonstrates significantly superior efficiency compared to language-aligned approaches.

Furthermore, we present more zero-shot performance on the Monash TSF benchmark and full-shot results on the LongTerm TSF benchmark in Table 10 and 12 of appendix.

\begin{figure}[t]
\begin{center}
\centerline{\includegraphics[width=1.0\columnwidth]{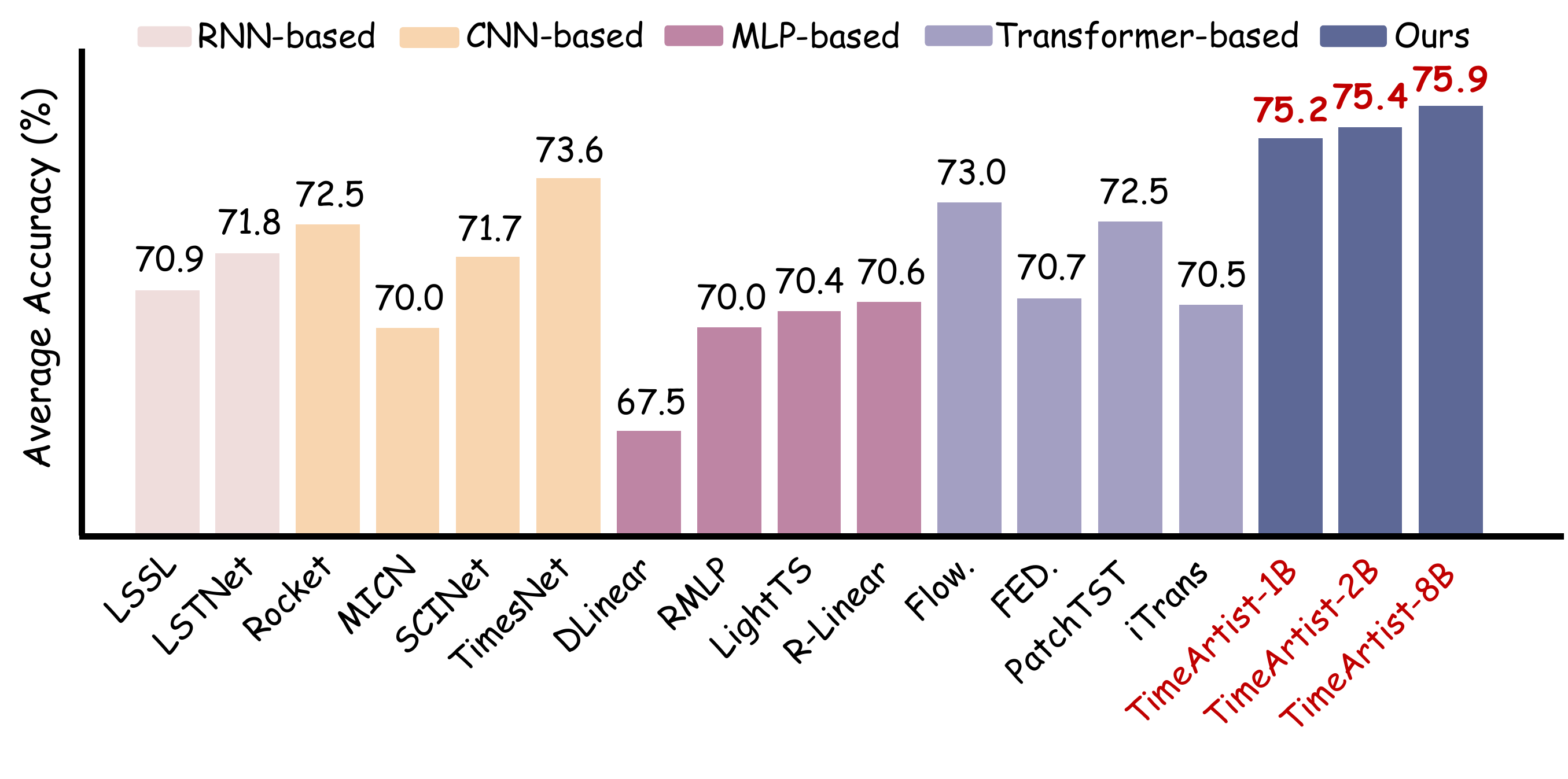}}
\vspace{-13pt}
\caption{Results of classification are averaged from UEA datasets. The detailed results are in the Table 15 of appendix.}\label{fig:exper3}
\end{center}
\vspace{-25pt}
\end{figure}

\subsection{Semantically enhanced TSC}
The difference at inductive biases between discriminative and generative tasks imposes significant challenges for developing universal models, as model must simultaneously learn pattern recognition and data distribution. 
\myformer achieves preliminary unification by converting time series into distinct semantic objects. 
Extensive experiments were conducted on the UEA classification archive \cite{bagnall2018uea}. 
Specifically, we utilized InternVL2.5 as the LMM, and demonstrated multiple variants (1B, 2B, 8B) in the Figure~\ref{fig:exper3}, which summarizes the baseline's average accuracy across all 30 datasets. 
Compared to prevalent Transformer architectures, \myformer demonstrates a performance improvement of 5.6\%-11.4\% in classification accuracy.

\begin{figure}[t]
\begin{center}
\centerline{\includegraphics[width=1.0\columnwidth]{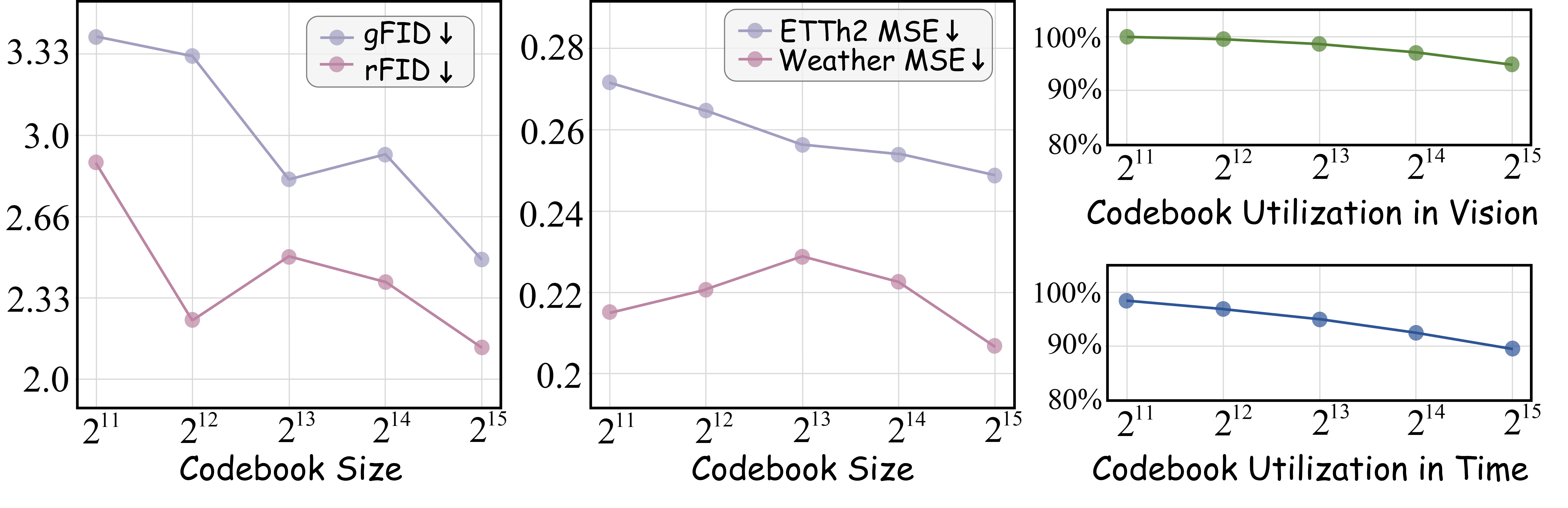}}
\vspace{-5pt}
\caption{Impact of codebook size on reconstruction quality, generation, and zero-shot forecasting benchmarks.}\label{fig:exper4}
\end{center}
\vspace{-28pt}
\end{figure}

\subsection{Ablation Studies of TimeArtist}
\subsubsection{Utilization of Codebook}
Figure~\ref{fig:exper4} demonstrates the performance across codebook sizes ranging from \begin{small}$2^{11}$\end{small}(1k) to \begin{small}$2^{15}$\end{small}(32k). Our approach demonstrates consistent performance improvements in both reconstruction and zero-shot TSF tasks as the codebook size increases, since large codebook size offers more high- and low-level feature combination possibilities. 
Notably, \myformer maintains over 90\% utilization efficiency in the temporal-visual domain even at maximum scale, which is attributed to proposed multi-head quantization method's remarkable scalability that significantly beyond prior approaches~\citep{lee2023vector,rasul2024vq} in both capacity and efficiency. 
All Results were obtained by keeping the codebook size at 4,096.

\begin{table}[ht]
  \caption{Ablation Studies. Results present the mean MSE values across 4 prediction lengths for diverse model variants.}\label{tab:task5_ablation}
  \vspace{-8pt}
  \centering
  \resizebox{0.9\columnwidth}{!}{
  \begin{small}
  \renewcommand{\multirowsetup}{\centering}
  \tabcolsep=0.3cm
  \renewcommand\arraystretch{1.2}
  \begin{tabular}{cccccc}
    \toprule
    Dataset & \myformer & w/o VM & VM2Attn & VM2Trsf & RandVM \\
    \midrule

    ETTh1 & 0.379 & 0.873 & 0.438 & 0.452 & 0.619 \\
    ETTh2 & 0.308 & 0.461 & 0.392 & 0.424 & 0.530 \\
    ETTm1 & 0.368 & 0.758 & 0.415 & 0.431 & 0.519 \\
    ETTm2 & 0.264 & 0.411 & 0.309 & 0.308 & 0.372 \\
    
    \bottomrule
  \end{tabular}
  \end{small}
}
\vspace{-10pt}
\end{table}

\subsubsection{Effectiveness of Visual Model}
To rigorously validate the effectiveness of the proposed alignment strategy, we adhere to the ablation protocol proposed by~\citet{tan2024language} and~\citet{chen2024visionts}. 
Specifically, we substitute the proposed dual-autoencoder and quantizer components with the VM (vision model). Table~\ref{tab:task5_ablation} presents multiple ablation variants, including: (1) w/o VM (removing VM entirely), (2) VM2Attn/VM2Trsf (replacing the VM with a single attention or transformer layer), and (3) RandVM (randomly initializing the VM). 
The significant performance degradation observed in these variants conclusively demonstrates that representation-level alignment is critical for establishing a unified temporal-visual multimodal model.

\begin{table}[ht]
\caption{
We conducted an ablation on image category pairs and demonstrated the accuracy rate on the UEA dataset.
}\label{tab:ablation_tsc_brief}
\vspace{-8pt}
\centering
\resizebox{1.0\columnwidth}{!}{
\begin{small}
\renewcommand{\multirowsetup}{\centering}
\tabcolsep=0.05cm
\renewcommand\arraystretch{1.0}
\begin{tabular}{ccccc}
\toprule
{categorys} & {Face.(\%)} & {Heart.(\%)} & {SCP1.(\%)} & {SCP2.(\%)} \\
\midrule
Persian Cat $vs.$ School Bus & 71.9 & 79.3 & 88.1 & 56.8 \\
Golden Retriever $vs.$ Siamese Cat & 72.1 & 78.5 & 87.8 & 57.0 \\
Gray Wolf $vs.$ Siberian Husky & 73.4 & 78.8 & 87.6 & 57.2 \\
Coffee Cup $vs.$ Mug & 72.8 & 79.1 & 87.3 & 57.0 \\
Elephant $vs.$ Ant & 71.2 & 78.5 & 87.9 & 58.9 \\

\bottomrule
\end{tabular}
\end{small}
}
\vspace{-8pt}
\end{table}
\subsubsection{Image categories in TSC}
To investigate the impact of predefined label-category pairs on temporal-visual alignment training in TSC, we construct five ImageNet-1k image category pairs varying difficulty levels (from trivial to challenging) and diverse semantic domains (animals, artifacts, natural objects) in Table~\ref{tab:ablation_tsc_brief}. 
The ablation study shows classification performance remains stable across visual semantic variations, confirming the model's ability to learn discriminative features during alignment. Details in Table 18 of appendix E.

\subsubsection{Hyperparameter analysis}
We demonstrates the full-shot forecasting's performance of TimeArtist under diverse model configurations, in and Table 16 of appendix E.

\subsubsection{Memory usage in Finetuning}
In the appendix E, Table 17 serves as a reference for the computational complexity of model fine-tuning in the full-shot forecasting scenario.

\section{Conclusion and Future Work}\label{section:conclusion}
\vspace{-3pt}
\paragraph{Conclusion.}
We establishes a versatile cross-modal framework, enabling free conversion between time series and images, while capturing temporal fluctuations to render and synthesise images. 
TimeArtist enables semantically enhanced temporal tasks, via pre-trained vision models. 


\vspace{-13pt}
\paragraph{Future Work.}
This work enhances time series tasks via vision models and shows potential for video generation by incorporating temporal dependencies, enabling coherent motion and long-term sequence consistency.



\clearpage
{
    \small
    \bibliographystyle{ieeenat_fullname}
    \bibliography{main}
}

\clearpage
\newpage
\section{A. Further Related Work} \label{section:further_relatedwork}
\subsection{A.1 Applications of Vision Models in TSF}
The application of visual techniques to time series analysis has evolved through diverse distinct phases. Pioneering attempts involved preprocessing time series utilizing statistical transformation. 
Notably, \citet{wang2015imaging} introduced gramian angular and markov transition to reconstruct time series into 2d spectrograms, enabling subsequent CNN-based classification and imputation tasks. 
Parallel work by \citet{li2020forecasting} and \citet{hatami2018classification} employed recurrence plot to generate 2d texture-like matrix from time series for forecasting and classification. Time-frequency time series analysis has been widely witnessed significant advancement. 
Existing approaches like TimesNet~\citep{wu2022timesnet} and TimeMixer++\citep{wang2024timemixer++} attempt to decompose raw sequences into periodic components and reorganize them as 2d grayscale images. 
Then leveraging convolutional kernels' receptive fields to simultaneously capture intra- and inter- period dependencies.

Recent studies have explored leveraging the cross-modality generalization capabilities of pre-trained large vision models (LVMs) to enhance zero-shot potential in time-series tasks. 
For instance, ViTST~\citep{li2023time} and VisionTS~\citep{chen2024visionts} employ ViT-based~\citep{dosovitskiy2020image} MAE~\citep{he2022masked} architectures to reconstruct masked portions of grayscale images, ultimately flattening the output to generate predictive sequences. 
Similarly, TimeVLM~\citep{zhong2025time} investigates the incorporation of supplementary textual and visual modalities to enhance TSF.
In addition, ViTime~\citep{yang2024vitime} generated synthetic time series data to pre-train a vision model for the TSF task. CLIP-LSTM~\citep{wimmer2023leveraging} and InsightMiner~\citep{zhang2023insight} employed vision-language multimodal pre-trained models to extract predictive features and generate text descriptions. 

However, These approaches~\citep{li2020forecasting,li2023time,chen2024visionts,zhong2025time} fail to establish semantic-level feature alignment. 
The synthesized grayscale images generated through periodic partitioning and stacking operations~\citep{wu2022timesnet,wang2024timemixer++} merely exhibit structural similarity to natural images while lacking explicit semantic information (e.g., cat or dog categories). 
Crucially, mainstream vision models~\citep{radford2021learning,liu2024llavanext,zhu2025internvl3} are exclusively pretrained on authentic natural image datasets (e.g., ImageNet~\citep{deng2009imagenet}, LAION~\citep{schuhmann2021laion} and COYO~\cite{kakaobrain2022coyo-700m}). 
Due to the incompatibility between the highly abstract pixel arrangements in grayscale images (from time series) and the statistical priors learned during visual pretraining, the learned inductive biases of large vision models (LVMs) can't be effectively transferred to tasks of grayscale domains (time series forecasting and classification). 
This fundamental mismatch severely limits generalization capability and zero-shot performance.

Surpassing existing approaches, \myformer pioneers the alignment between temporal fluctuations and visual semantics at the representation level, where wave bands in time series are effectively ``translated'' into discernible visual concepts such as forests and mountain peaks. 
This comprehensive alignment enables unprecedented capabilities, including bidirectional transformation between 1d time series and 2d image, as well as style transfer through temporal-pattern-guided image rendering. 
Furthermore, the semantically enriched representations fully unleash the potential of LVMs, significantly enhancing zero-shot performance in time series forecasting (TSF) and classification (TSC).

\subsection{A.2 Aligning Temporal and Visual Representation}
Fundamentally, temporal and visual modalities share intrinsic representational homology. 
Distinct from linguistic systems emergent through complex neural interactions in brain, both time series and image constitute direct observational samples from the natural world. 
The difference in information density necessitates distinct tokenization for modeling: textual data employs word embedding to derive latent tokens, while images and time series obtain latent tokens directly through patchification operations. 
Furthermore, both modalities exhibit hierarchical representational structures: images contain semantic and pixel-level features, while time series exhibit trend-seasonal and multi-periodic coupling. 
These inherent parallels motivate our approach of establishing a shared representation space as prior knowledge, subsequently learning joint posterior distributions of temporal and visual data to achieve cross-modal alignment.

Recent advances have demonstrated remarkable achievements in generating images from brain signals, including fMRI~\citep{bird2014categorical} and EEG~\citep{kavasidis2017brain2image}. 
Early studies~\citep{mozafari2020reconstructing,ozcelik2022reconstruction} employed regression models to extract latent representations and fine-tuned BigGAN~\citep{brock2018large} to reconstruct natural scenes. 
MinD-VIS~\citep{chen2023seeing} integrated the LDM~\citep{rombach2022high} architecture to preserve rich semantic information and generate higher-quality images. 
Brain2Image~\citep{kavasidis2017brain2image} generates visual stimuli capable of eliciting specific neural responses directly from EEG data. BrainEdit~\citep{davis2022brain} utilizes EEG signals as supervisory inputs to learn semantic representations for image editing applications. 
DreamDiffusion~\citep{bai2024dreamdiffusion} introduces pretrained CLIP~\citep{radford2021learning} supervision to facilitate multimodal alignment across EEG, text, and vision. 

These alignment strategies assume a strict one-to-one mapping between time-series samples and images. However, unlike images, which can naturally be interpreted by human language~\citep{schuhmann2021laion,kakaobrain2022coyo-700m}, real-world time-series data often lack corresponding visual counterparts. 
Consequently, this CLIP-style alignment strategy restricts the development of general-purpose temporal-visual multimodal models. 
In contrast to existing approaches limited to EEG signals, \myformer aims to leverage broader temporal data domains to establish a universal temporal-visual multimodal framework.

\begin{figure*}[ht]
\begin{center}
\centerline{\includegraphics[width=1.9\columnwidth]{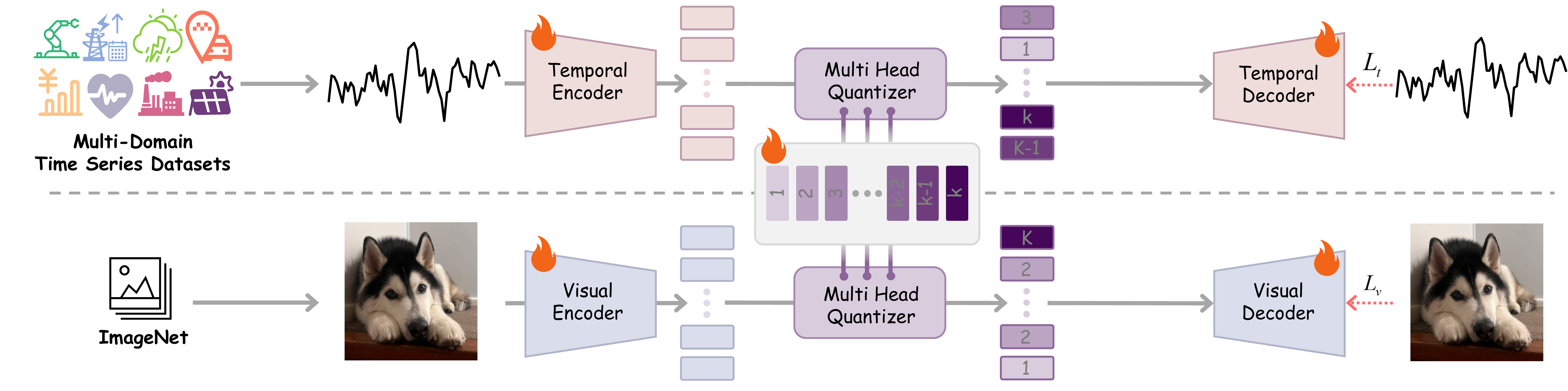}}
\caption{
Self-supervised pretraining of Dual-autoencoder and MH-Quantizer without paired time-series and image data.
}\label{fig:training_0}
\end{center}
\vspace{-20pt}
\end{figure*}

\subsection{A.3 Vector Quantization in Vision and Time Series}
Vector quantization (VQ) strategies have achieved remarkable success in the field of visual generation~\citep{qu2025tokenflow,yu2024image}. 
VQVAE~\citep{van2017neural} pioneered the paradigm that first training a discrete tokenizer through reconstruction tasks to encode an image into discrete tokens sequence, followed by directly modeling the underlying distribution of token sequences via an autoregressive model~\citep{chen2018pixelsnail}. 
VQVAE-2~\citep{razavi2019generating} advanced this framework through exponential moving average updates and a hierarchical multi-scale approach. 
Subsequent image generation studies adopted this two-stage paradigm, where DALL-E~\citep{ramesh2021zero} improves token prediction in second stage by using Transformers. 
VQGAN~\citep{esser2021taming} further enhanced the architecture by incorporating adversarial and perceptual losses~\citep{johnson2016perceptual,zhang2018unreasonable}, yielding more precise and detailed representations. 
Recent advances in VQ tokenizers have focused on three main directions: improving reconstruction fidelity and generation quality~\citep{lee2022autoregressive}, enhancing codebook utilization~\citep{yu2021vector,zhang2023regularized,zhu2024scaling}, and exploring novel architectures such as the multi-scale VQVAE~\citep{tian2024visual} for next-scale prediction of images. 

Recent studies have begun exploring the potential value of VQ in time series analysis. TimeVQVAE~\citep{lee2023vector} introduces VQ in the time-frequency domain to capture global temporal consistency, for time series generation. 
SparseVQ~\citep{zhao2024sparse} employs VQ to capture sufficient statistics for forecasting, serving as an alternative to the FFN. 
VQ-TR~\citep{rasul2024vq} utilizes VQ to compress long sequences into multiple discrete representations, thereby enabling a transformer-based model with linear complexity. Surpassing existing methodologies, \myformer proposes an innovative Multi-Head Quantization (MHQ) strategy. 
For image, MHQ concurrently models high-level semantic information and low-level pixel-wise features; for time series, it effectively disentangles and captures trend-seasonal components along with multi-periodic patterns. 
By introducing a joint representation space rather than relying on a single distribution, \myformer successfully addresses the critical challenges of codebook underutilization and collapse in discrete representation spaces.

\section{B. Implementation Details}\label{sec:detail}
As a powerful and versatile cross-modal paradigm, TimeArtist has undergone comprehensive evaluation across diverse critical scenarios. 
Concretely, we utilize the ``warmup-align'' strategy to train \myformer on large scale time series datasets~\citep{ansari2024chronos} and image datasets~\citep{deng2009imagenet}, where \myformer simultaneously accepts both temporal and visual data as input. 
For visual inputs, the image resolution is H=256 and W=256, with the downsampling rate f=16. The codebook contains 4096 discrete codes with each entry a vector with 12 channels. 
For temporal inputs, we strictly adhere to the experimental configuration established by A et al., maintaining identical parameters including: (1) observation sequence length (context window), (2) train-test split methodology, and (3) evaluation protocols. 
This design ensures fair comparison with state-of-the-art benchmarks while eliminating potential confounding factors from preprocessing variations. 
Furthermore, we validated the utilization rate of encoded vectors across codebook sizes ranging from 2,048 (2k) to 32,768 (32k), to assess the effectiveness of the multi-head quantization mechanism. 
Finally, we conducted extensive ablation studies on the model configuration. Detailed experimental setups and dataset information will be provided.

\subsection{B.1 Training Details}\label{sec:more_training}
TimeArtist serves as a unified architecture for both visual and temporal tasks. We design distinct training and inference strategies tailored to different application scenarios, including (1) pre-training for time series and image generation, (2) time series forecasting  and (3) time series classification. 

The dual-autoencoder module obtained during the pre-training stage is reused in both subsequent forecasting and classification tasks. 
Stage 1 enables the model to capture general visual semantics and temporal fluctuation patterns through large-scale self-supervised learning, while stages 2 and 3 focus specifically on learning fine-grained semantic alignment relationships between time series samples and images, without the need to relearn low-level features.

\subsubsection{B.1.1 Pre-training for Image Generation}
During the pre-training phase, we collect a large-scale corpus of unpaired time series and visual samples from publicly available sources (Monash~\cite{godahewa2021monash} time series datasets and ImageNet~\cite{deng2009imagenet}) to self-supervised pretrain the dual-autoencoder and shared quantizer. As illustrated in Figure~\ref{fig:training_0}, this stage does not require paired time-series and image data. The objective is to endow the model with strong generalization capabilities through large-scale representation learning. Once pretrained, the model enables seamless bidirectional conversion between time series and images, thereby facilitating cross-modal generation.

\subsubsection{B.1.2 Time Series Forecasting}
\begin{figure*}[ht]
\begin{center}
\centerline{\includegraphics[width=1.9\columnwidth]{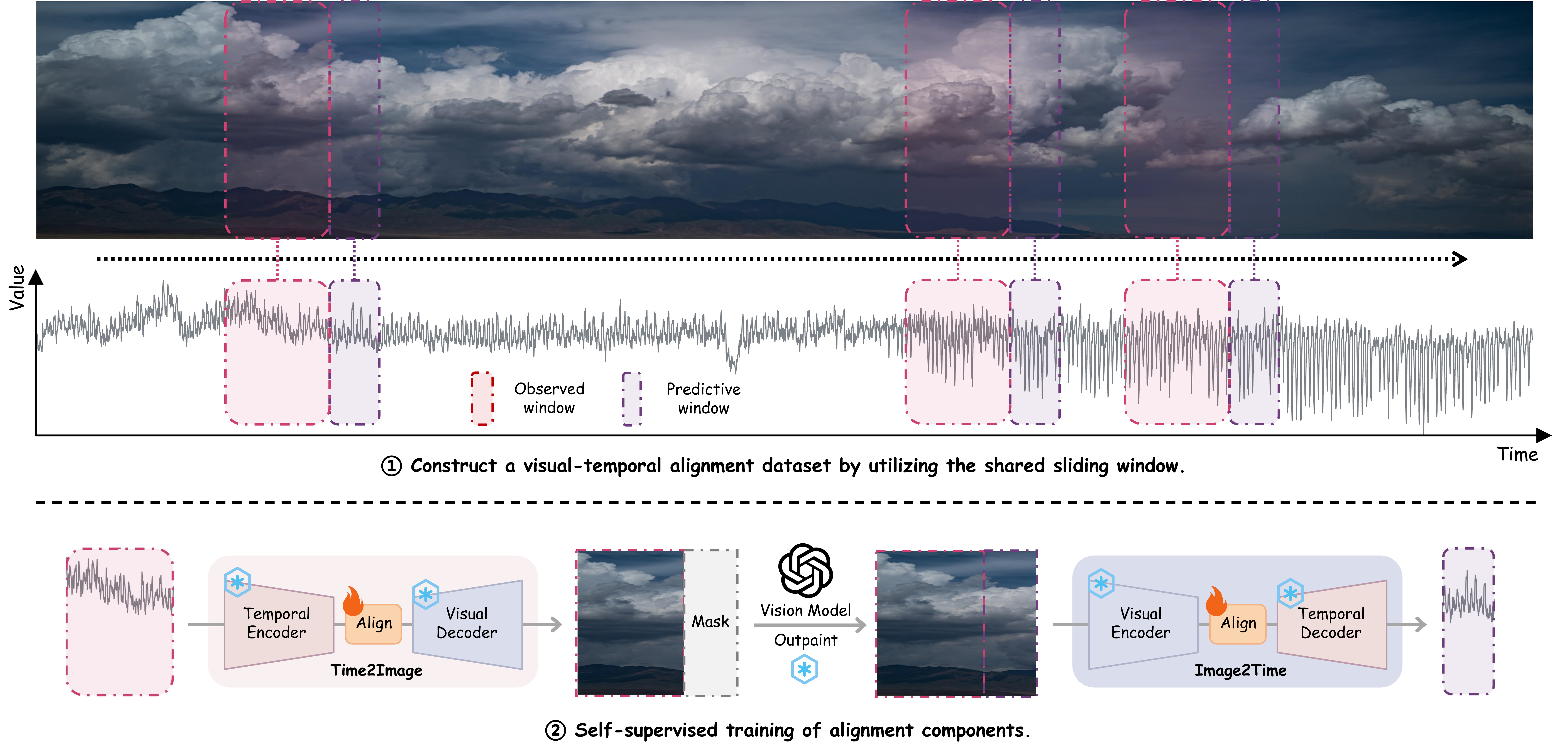}}
\caption{
Utilizing the shared sliding time window, self-supervise the training of the modal alignment component.
}\label{fig:training_1}
\end{center}
\vspace{-20pt}
\end{figure*}

\begin{figure*}[ht]
\begin{center}
\centerline{\includegraphics[width=1.9\columnwidth]{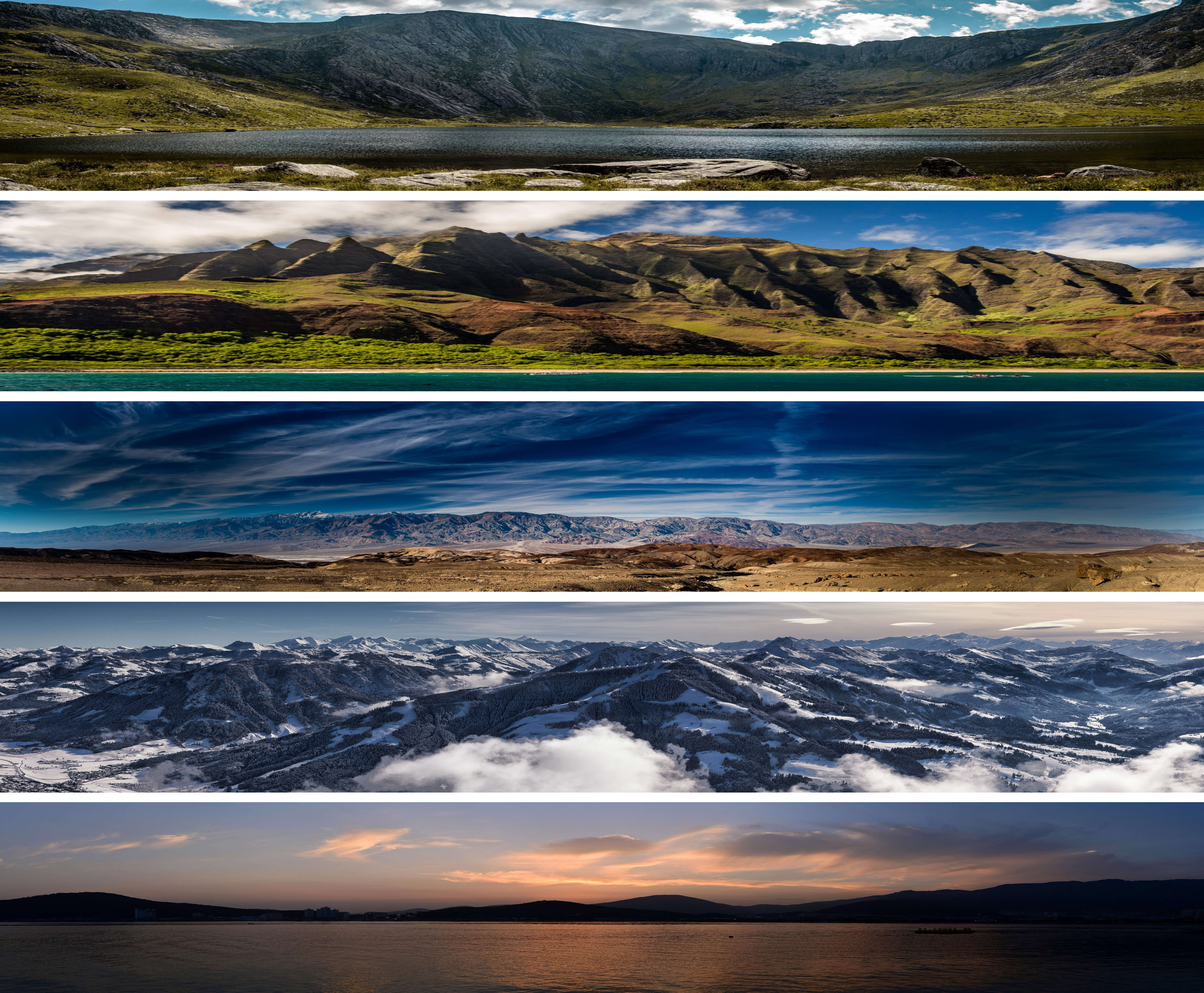}}
\caption{
We have collected different landscape images for each time series dataset in each domain to avoid duplicate pairings.
}\label{fig:training_2}
\end{center}
\vspace{-20pt}
\end{figure*}

In this stage, we align time series samples with designated landscape images. Specifically, we assign one landscape image per temporal domain from the Monash time series datasets and fine-tune the alignment module end-to-end via sliding-window strategy, as depicted in Figure~\ref{fig:training_1}. 
Throughout this process, we freeze the parameters of the dual-autoencoder and quantizer to preserve the pixel-level visual semantics and temporal fluctuation patterns learned during pre-training. 
Figure~\ref{fig:training_2} showcases a subset of the landscape images used. Our experiments reveal that images with higher resolution retain more temporal information and yield improved forecasting accuracy. Moreover, landscapes with multi-scale visual elements, such as foreground objects aligned with high-frequency components and distant scenery corresponding to low-frequency trends, demonstrate superior capability in capturing hierarchical temporal patterns.

\subsubsection{B.1.3 Time Series Classification}
Similar to the forecasting, we freeze the dual-autoencoder and quantizer and train the alignment module on labeled time-series and image pairs. For each classification dataset, a dedicated alignment module must be trained, with the number of classes matching between the temporal and visual domains. Furthermore, multiple sets of visual categories can be designed for the same classification task to enhance robustness. Table~\ref{tab:ablation_tsc} presents an ablation study on the impact of different visual label sets on classification performance.

\subsubsection{B.1.3 Training Costs}
During the pre-training phase, we leveraged large-scale time series datasets such as Monash (time series datasets) and ImageNet (vision datasets) to self-supervise the training of the temporal encoder. This process was conducted on 8 A100 GPUs for 45 hours, while the visual tokenizer, pre-trained using TiTok was loaded.

In the time series forecasting alignment phase, we collected landscape images depicting natural scenery from the web to train the alignment module. This training stage required 12 hours on 8 A100 GPUs.

For the time series classification alignment phase, we selected 12,000 images with specified labels from ImageNet to train the alignment module. Across all 30 UEA time series classification datasets, the total training time amounted to 8 hours on 4 A100 GPUs.

Notably, using only a small subset (10\%) of images for training was observed to reduce the diversity of the generated images, though it did not compromise the accuracy of the time series classification task. In contrast, the performance of the forecasting task was more significantly influenced by the quality of the training images.

\subsection{B.2 Visual Benchmarks}
We employ ImageNet-1K~\citep{deng2009imagenet} as the large-scale image dataset to train \myformer. 
This dataset represents a foundational computer vision recognition project designed to emulate human visual cognition by enabling object identification from images. 
The ImageNet-1K comprises 1,000 object categories spanning animals, plants, man-made objects, and natural scenes. 
The train set contains 1,281,167 images with approximately 1,300 samples per class, while the validation and test sets include 50,000 and 100,000 images respectively. 
Originally web-crawled and manually annotated, the dataset exhibits variable native resolutions (typically $469\times 387$ pixels), with all images uniformly resized to $256\times 256$ pixels during preprocessing for model training and evaluation.

\subsection{B.3 Visual Baselines}
We select vision generation baselines for comparison, including \textbf{diffusion-based} and \textbf{autoregression-based} vision generation model. The baseline models selected for comparison are briefly described below:

\paragraph{1.LDM~\citep{rombach2022high}} is a hierarchical generative framework that implements stable diffusion processes in compressed latent spaces, achieving superior sample quality with significantly reduced computational costs compared to pixel-space diffusion models.

\paragraph{2.DiT~\citep{peebles2023scalable}} is a scalable transformer-based diffusion architecture that replaces traditional U-Net backbones with vision transformers.

\paragraph{3.VQGAN~\citep{esser2021taming}} is a two-stage generative model that combines VQVAE with adversarial training, achieving high-fidelity image synthesis through discrete codebook representations and patch-based discriminators.

\paragraph{4.RQVAE~\citep{lee2022autoregressive}} is a residual-quantized variational autoencoder framework that progressively discretizes latent representations through multi-stage residual quantization.

\paragraph{5.MaskGIT~\citep{chang2022maskgit}} is a bidirectional transformer-based image generation framework that predicts multiple masked tokens simultaneously through masked visual token modeling.

\subsection{B.4 Temporal Benchmarks}
In accordance with the evaluation benchmarking protocol established by Chronos~\citep{ansari2024chronos}, we employ 55 datasets aggregated from multiple sources, including the Monash Time Series Forecasting Repository~\citep{godahewa2021monash}, M-competitions~\citep{makridakis2020m4}, and public-domain datasets from Kaggle, for comprehensive training and evaluation of \myformer. 
These datasets encompass diverse application domains such as energy, transportation, healthcare, retail, networking, meteorology, and finance, with sampling frequencies ranging from 5-minute intervals to annual recordings. 
This rich statistical diversity ensures \myformer develops robust prior knowledge across extensive temporal patterns. The complete dataset inventory, including detailed source attributions and specifications, is documented in table~\ref{tab:dataset}.

To evaluate TimeArtist's zero-shot performance on the Long-Term TSF benchmark, we implemented strict segregation between training and evaluation datasets to prevent data leakage. 
Specifically, we selected six widely-used datasets from the Long-Term TSF benchmark that were not included in Chronos' pretraining corpus as our zero-shot TSF evaluation set. 
Furthermore, since most baseline models lack native zero-shot prediction capability, we fine-tune these baselines on 10\% of each target dataset to enable comparative analysis under few-shot settings.
Our comprehensive evaluation protocol consists of two components: (1) zero-shot TSF performance assessment on Monash benchmarks, and (2) zero-shot versus full-shot TSF evaluation on Long-Term TSF benchmarks, where complete results provided in section~\ref{sec:full_results}. Below we detail the benchmark specifications:

\paragraph{Long-Term TSF Benchmark} We evaluate our model on 7 widely used long-term TSF datasets~\citep{wu2022timesnet}, including ETTh1, ETTh2, ETTm1, ETTm2, Electricity, Traffic, and Weather. Performance is assessed using Mean Squared Error (MSE) and Mean Absolute Error (MAE), with lower values indicating better forecasting accuracy. 

\paragraph{Monash Benchmark} Following~\citet{woo2024unified}, we tested 29 Monash datasets~\citep{godahewa2021monash} using GluonTS~\citep{alexandrov2020gluonts}, including M1 Monthly, M3 Monthly, M3 Other, M4 Monthly, M4 Weekly, M4 Daily, M4 Hourly, Tourism Quarterly, Tourism Monthly, CIF 2016, Australian Electricity Demand, Bitcoin, Pedestrian Counts, Vehicle Trips, KDD Cup, Weather, NN5 Daily, NN5 Weekly, Carparts, FRED-MD, Traffic Hourly, Traffic Weekly, Rideshare, Hospital, COVID Deaths, Temperature Rain, Sunspot, Saugeen River Flow, and US Births. Performance is assessed using MAE.

\subsection{B.5 Temporal Baselines}
We select representative baselines for comparison, including \textbf{Vision-based}, \textbf{TS-based} and \textbf{Text-based} foundation models, and \textbf{other popular TSF baselines} covering both Transformer-based, MLP-based and CNN-based architectures. The baseline models selected for comparison are briefly described below:

\paragraph{1.VisionTS~\citep{chen2024visionts}} is a vision-based zero-shot TSF model, utilizing the periodic folding strategy, the time series is transformed into grayscale image, and the MAE is employed to generate the time series.

\paragraph{2.TimeVLM~\citep{zhong2025time}} utilizes a multimodal model to integrate temporal, visual and linguistic inputs, and enhances the model's predictive performance through few-shot fine-tuning.

\paragraph{3.Moirai~\citep{woo2024unified}} is a TSF foundation model trained on the Large-scale Open Time Series Archive (LOTSA), with over 27B observations across nine domains. Moirai achieve zero-shot forecasting across diverse domains, frequencies and variables in a zero-sample manner.

\paragraph{4.TimeLLM~\citep{jin2023time}} is a text-based TSF foundation model. By reprogramming time series data to introduce prompt information, the temporal modalities are aligned with the text, thereby enhancing the reasoning ability of LLM in the temporal domain.

\paragraph{5.GPT4TS~\citep{zhou2023one}} is a text-based TSF foundation model, which freezing attention and FFN modules from the pre-trained GPT2, and fine-tuned linear layer to adapt to downstream tasks.

\paragraph{6.DLinear~\citep{zeng2023dlinear}} proposes a linear forecasting model, enhanced by seasonal-trend decomposition or normalization.

\paragraph{7.PatchTST~\citep{Nie2023PatchTST}} uses Transformer encoders with patching and channel independence techniques for improved predictions.

\paragraph{8.TimesNet~\citep{wu2022timesnet}} applies convolution kernels along the time dimension, using temporal decomposition and periodical segmentation to capture temporal patterns.

\paragraph{9.FEDformer~\citep{zhou2022fedformer}} employs a sparse frequency domain representation, using frequency-enhanced blocks for cross-time dependency.

\paragraph{10.Autoformer~\citep{wu2021autoformer}} uses series decomposition blocks and Auto-Correlation to capture cross-time dependency.

\paragraph{11.Stationary~\citep{Liu2022NonstationaryTR}} introduces stationarization and de-stationary attention mechanisms.

\paragraph{12.ETSformer~\citep{woo2022etsformer}} leverages exponential smoothing principles, including exponential smoothing and frequency attention mechanisms.

\paragraph{13.Informer~\citep{zhou2021informer}} proposes ProbSparse self-attention and distillation operations.

\paragraph{14.DiffusionTS~\citep{yuan2024diffusion}} proposes a novel diffusion-based framework that generates high quality time series by utilizing the autoencoder transformer with disentangled temporal representations.

\paragraph{15.TimeGAN~\citep{yoon2019timegan}} generate realistic time series that combines the flexibility of the unsupervised paradigm with the control afforded by supervised training.

\paragraph{16.TimeVAE~\citep{desai2021timevae}} propose a novel architecture for synthetically generating time-series data with the use of variational autoencoders.

\paragraph{17.Diffwave~\citep{kong2020diffwave}} is a versatile diffusion probabilistic model for conditional and unconditional waveform generation.


\paragraph{Summary.} For the long-term TSF benchmark, we include TS-based foundation model results from their original papers, Vision-based model results from~\citet{chen2024visionts,zhong2025time}, Text-based model results from~\citet{tan2024language}, and other baseline results from~\citet{zhou2023one}. 
For the Monash and PF benchmark, we include results from~\citet{woo2024unified}.

\paragraph{Environment} All experiments are conducted using \textit{Time-Series-Library}(\url{https://github.com/thuml/Time-Series-Library}) and GluonTS library \citep{alexandrov2020gluonts} on an NVIDIA V100 GPU.

\subsection{B.6 Evaluation Metrics}
\subsubsection{\textbf{Discriminative and Predictive score.}}
The discriminative score is calculated as $\mid$accuracy $-$ 0.5$\mid$, while the predictive score is the mean absolute error (MAE) evaluated between the predicted values and the ground-truth values in test data. For a fair comparison, we reuse the experimental settings of TimeGAN \cite{yoon2019timegan} for the discriminative and predictive score. Both the classifier and sequence-prediction model use a 2-layer GRU-based neural network architecture. 

\subsubsection{\textbf{Context-FID score.}}
A lower FID score means the synthetic sequences are distributed closer to the original data. \cite{kong2020diffwave} proposed a Frechet Inception distance (FID)-like score, Context-FID (Context-Frechet Inception distance) score by replacing the Inception model of the original FID with a time series representation learning method called TS2Vec \citep{desai2021timevae}. They have shown that the lowest scoring models correspond to the best-performing models in downstream tasks and that the Context-FID score correlates with the downstream forecasting performance of the generative model. Specifically, we first sample synthetic time series and real-time series respectively. Then we compute the FID score of the representation after encoding them with a pre-trained TS2Vec model.

\subsubsection{\textbf{Correlational score.}}
Following \cite{yuan2024diffusion}, we estimate the covariance of the $i^{th}$ and $j^{th}$ feature of time series as follows:
\begin{equation}
    {{\mathop{\rm cov}} _{i,j}} = \frac{1}{T}\sum\limits_{t = 1}^T {X_i^tX_i^t}  - \left( {\frac{1}{T}\sum\limits_{t = 1}^T {X_i^t} } \right)\left( {\frac{1}{T}\sum\limits_{t = 1}^T {X_{j}^t}} \right).
\end{equation}
Then the metric on the correlation between the real data and synthetic data is computed by
\begin{equation}
    \frac{1}{{10}}\sum\limits_{i,j}^d {\left| {\frac{{{\mathop{\rm cov}} _{i,j}^r}}{{\sqrt {{\mathop{\rm cov}} _{i,i}^r} {\mathop{\rm cov}} _{j,j}^r}} - \frac{{{\mathop{\rm cov}} _{i,j}^f}}{{\sqrt {{\mathop{\rm cov}} _{i,i}^f} {\mathop{\rm cov}} _{j,j}^f}}} \right|},
\end{equation}

\begin{table*}[thbp]
\caption{
All the datasets required for time series forecasting. 
We divide these datasets according to the usage during the training and evaluation of \myformer. 
The data of \emph{Pretraining} is only used for training model. 
In the zero-shot TSF scenario, \emph{Evaluation}'s test set will be utilize to evaluate the model. In the full-shot TSF, we freeze the dual-autoencoder and quantizer structures and use the \emph{Evaluation}'s train set to fine-tune the alignment model to align the observed time series to the landscape paintings. 
Finally, the \emph{Evaluation}'s test set is used to evaluate the model.
}\label{tab:dataset}
\centering
\resizebox{1.2\columnwidth}{!}{
\begin{threeparttable}
\begin{small}
\renewcommand{\multirowsetup}{\centering}
\setlength{\tabcolsep}{3.8pt}
\begin{tabular}{ccccccc}
\toprule

\multirow{2}{*}{\rotatebox{0}{Dataset}} & 
\multirow{2}{*}{\rotatebox{0}{Domain}} & 
\multirow{2}{*}{\rotatebox{0}{Frequency}} & 
\multirow{2}{*}{\rotatebox{0}{Series Number}} & 
\multicolumn{3}{c}{Series Length} \\
& & & & min & avg & max \\ 
\midrule

Pretraining & & & & & & \\ 
\midrule

Brazilian Cities Temperature & nature & M & 12 & 492 & 757 & 1320 \\ 
Mexico City Bikes & transport & 1H & 494 & 780 & 78313 & 104449 \\ 
Solar (5 Min.) & energy & 5min & 5166 & 105120 & 105120 & 105120 \\ 
Solar (Hourly) & energy & 1H & 5166 & 8760 & 8760 & 8760 \\ 
Spanish Energy and Weather & energy & 1H & 66 & 35064 & 35064 & 35064 \\ 
Taxi (Hourly) & transport & 1H & 2428 & 734 & 739 & 744 \\ 
USHCN & nature & 1D & 6090 & 5906 & 38653 & 59283 \\ 
Weatherbench (Daily) & nature & 1D & 225280 & 14609 & 14609 & 14610 \\ 
Weatherbench (Hourly) & nature & 1H & 225280 & 350633 & 350639 & 350640 \\ 
Weatherbench (Weekly) & nature & 1W & 225280 & 2087 & 2087 & 2087 \\ 
Wiki Daily (100k) & web & 1D & 100000 & 2741 & 2741 & 2741 \\ 
Wind Farms (Daily) & energy & 1D & 337 & 71 & 354 & 366 \\ 
Wind Farms (Hourly) & energy & 1H & 337 & 1715 & 8514 & 8784 \\ 
Electricity (15 Min.) & energy & 15min & 370 & 16032 & 113341 & 140256 \\ 
Electricity (Hourly) & energy & 1H & 321 & 26304 & 26304 & 26304 \\ 
Electricity (Weekly) & energy & 1W & 321 & 156 & 156 & 156 \\ 
KDD Cup 2018 & nature & 1H & 270 & 9504 & 10897 & 10920 \\ 
London Smart Meters & energy & 30min & 5560 & 288 & 29951 & 39648 \\ 
M4 (Daily) & various & 1D & 4227 & 107 & 2371 & 9933 \\ 
M4 (Hourly) & various & 1H & 414 & 748 & 901 & 1008 \\ 
M4 (Monthly) & various & 1M & 48000 & 60 & 234 & 2812 \\ 
M4 (Weekly) & various & 1W & 359 & 93 & 1035 & 2610 \\ 
Pedestrian Counts & transport & 1H & 66 & 576 & 47459 & 96424 \\ 
Rideshare & transport & 1H & 2340 & 541 & 541 & 541 \\ 
Taxi (30 Min.) & transport & 30min & 2428 & 1469 & 1478 & 1488 \\ 
Temperature-Rain & nature & 1D & 32072 & 725 & 725 & 725 \\ 
Uber TLC (Daily) & transport & 1D & 262 & 181 & 181 & 181 \\ 
Uber TLC (Hourly) & transport & 1H & 262 & 4344 & 4344 & 4344 \\ 
\midrule

Evaluation & & & & & & \\ 
\midrule

Australian Electricity & energy & 30min & 5 & 230736 & 231052 & 232272 \\ 
CIF 2016 & banking & 1M & 72 & 28 & 98 & 120 \\ 
Car Parts & retail & 1M & 2674 & 51 & 51 & 51 \\ 
Covid Deaths & healthcare & 1D & 266 & 212 & 212 & 212 \\ 
Dominick & retail & 1D & 100014 & 201 & 296 & 399 \\ 
ERCOT Load & energy & 1H & 8 & 154854 & 154854 & 154854 \\ 
ETT (15 Min.) & energy & 15min & 14 & 69680 & 69680 & 69680 \\ 
ETT (Hourly) & energy & 1H & 14 & 17420 & 17420 & 17420 \\ 
Exchange Rate & finance & 1B & 8 & 7588 & 7588 & 7588 \\ 
FRED-MD & economics & 1M & 107 & 728 & 728 & 728 \\ 
Hospital & healthcare & 1M & 767 & 84 & 84 & 84 \\ 
M1 (Monthly) & various & 1M & 617 & 48 & 90 & 150 \\ 
M1 (Quarterly) & various & 3M & 203 & 18 & 48 & 114 \\ 
M1 (Yearly) & various & 1Y & 181 & 15 & 24 & 58 \\ 
M3 (Monthly) & various & 1M & 1428 & 66 & 117 & 144 \\ 
M3 (Quarterly) & various & 3M & 756 & 24 & 48 & 72 \\ 
M3 (Yearly) & various & 1Y & 645 & 20 & 28 & 47 \\ 
M4 (Quarterly) & various & 3M & 24000 & 24 & 100 & 874 \\ 
M4 (Yearly) & various & 1Y & 23000 & 19 & 37 & 841 \\ 
M5 & retail & 1D & 30490 & 124 & 1562 & 1969 \\ 
NN5 (Daily) & finance & 1D & 111 & 791 & 791 & 791 \\ 
NN5 (Weekly) & finance & 1W & 111 & 113 & 113 & 113 \\ 
Tourism (Monthly) & various & 1M & 366 & 91 & 298 & 333 \\ 
Tourism (Quarterly) & various & 1Q & 427 & 30 & 99 & 130 \\ 
Tourism (Yearly) & various & 1Y & 518 & 11 & 24 & 47 \\ 
Traffic & transport & 1H & 862 & 17544 & 17544 & 17544 \\ 
Weather & nature & 1D & 3010 & 1332 & 14296 & 65981 \\ 

\bottomrule
\end{tabular}
\end{small}
\end{threeparttable}
}
\end{table*}

\begin{table}
  \centering
  \caption{MAE results of TimesFM and LLMTime for zero-shot forecasting, on the last test window of the original test.}\label{tab:task3_zero_timesfm_llmtime}
  \small
  \resizebox{0.8\linewidth}{!}{
  \begin{tabular}{cccccc}
    \toprule
    \multicolumn{2}{c}{Method} & \myformer & VisionTS & TimesFM & LLMTime \\
    \midrule
    
    \multirow{2}[1]{*}{ETTh1} 
    & 96 & 0.35 & 0.35 & 0.45 & 0.42 \\
    & 192 & 0.43 & 0.45 & 0.53 & 0.50 \\
    \midrule
    
    \multirow{2}[0]{*}{ETTh2} 
    & 96 & 0.21 & 0.24 & 0.35 & 0.33 \\
    & 192 & 0.53 & 0.60 & 0.62 & 0.70 \\
    \midrule
    
    \multirow{2}[0]{*}{ETTm1} 
    & 96 & 0.11 & 0.12 & 0.19 & 0.37 \\
    & 192 & 0.21 & 0.23 & 0.26 & 0.71 \\
    \midrule
    \multirow{2}[0]{*}{ETTm2} 
    & 96 & 0.17 & 0.19 & 0.24 & 0.29 \\
    & 192 & 0.22 & 0.24 & 0.27 & 0.31 \\
    \midrule
    
    \multicolumn{2}{c}{ {Average}} 
    & 0.27 & 0.30 & 0.36 & 0.45 \\
    \bottomrule
    \end{tabular}%
    }
\end{table}%

\section{C. Warmup Strategy with Auxiliary Codes} 
The training process of vector-quantized models is notably sensitive, and their performance is significantly influenced by the adopted training paradigm. 
For instance, compared to DALL-E~\citep{ramesh2021zero}, which employs the original VQVAE~\citep{van2017neural} training strategy, VQGAN~\citep{esser2021taming} achieves substantial improvements in reconstruction FID on the ImageNet validation set. 
This enhancement benefits from advancements in perceptual loss~\citep{johnson2016perceptual,zhang2018unreasonable} and adversarial loss~\citep{goodfellow2020gan}.

Surpassing conventional training strategies~\citep{yu2021vector,yu2024image}, we introduce a novel ``warmup-align'' paradigm with improved efficiency and performance. 
Specifically, during the ``prior warmup'' stage, instead of directly training from scratch by reconstructing time series and images, we propose leveraging the discrete code of the standard MaskGIT-VQGAN~\citep{chang2022maskgit} to train TimeArtist's dual-autoencoder structure, referred to as auxiliary codes. 
Subsequently, in the ``semantic align'' stage, we freeze the dual-autoencoder and shared quantizer, designing a novel indices projection to align temporal fluctuation patterns with visual semantic representations. 
The proposed ``Warmup-Align'' strategy bypasses intricate loss functions and GAN framework, concentrating TimeArtist’s efforts on aligning temporal and visual modalities.

Critically, this modification fully preserves the functionality of the original VQGAN's image tokenizer and de-tokenization, while introducing an additional temporal modality and aligning it with visual semantics. 
These designs indicates \myformer can seamlessly integrate with state-of-the-art vision generation and understanding models~\citep{tian2024visual,qu2025tokenflow,kim2025tatitok}, demonstrating strong scalability. 
Results show that this two-stage strategy significantly improves training stability and downstream task performance, including: bidirectional conversion between time series and images, temporal-rendered image style transfer, zero-shot time series forecasting and time series classification based on vision models.

Furthermore, we follow the VQGAN's design during visual and temporal tokenization. 
Concretely, the encoder processes ViT output features through a $1\times1$ convolution to reduce token dimension. 
This dimensionality compression promotes semantically similar features to cluster into identical auxiliary codes, effectively mitigating two critical challenges: (1) code fragmentation caused by minor variations in high-dimensional space, and (2) codebook collapse. Consequently, this design achieves dual benefits of stabilized training dynamics and reduced computational complexity in code space search.

\section{D. Complete Result of \myformer}\label{sec:full_results}

\begin{table*}[ht]
  \centering
  \caption{Configurations of \myformer utilized in our zero-shot TSF by outpainting.}\label{tab:task3_zero_hyper}
  \small
    \begin{tabular}{ccccccc}
    \toprule
    & \textbf{ETTh1} & \textbf{ETTh2} & \textbf{ETTm1} & \textbf{ETTm2} & \textbf{Weather} & \textbf{Electricity} \\
    \midrule
    Observation-image Width & 256 & 256 & 256 & 256 & 256 & 256 \\
    Forecasting-image Width & 768 & 768 & 768 & 768 & 768 & 768 \\
    Context length $L$ & 2880 & 1728 & 2304 & 4032 & 4032 & 2880 \\
    \bottomrule
    \end{tabular}
\end{table*}%

\paragraph{Hyperparameters of zero-shot forecasting.}
To ensure the fairness of the comparison, we adopted the exact same context length settings as those of VisionTS. Furthermore, we convert the time series forecasting into an image outpainting. We fixed the height of the input and output images for the visual model to 256. The width of the input and output images are summarized in table~\ref{tab:task3_zero_hyper}.

\begin{table*}
\centering
\caption{The complete zero-shot forecasting results of \myformer on the Long-term TSF benchmark, which some LLM-based baselines were fine-tuned utlizing few-shot 10\% data for fine-tuning.}\label{tab:task3_zero_longtsf_all}
\tabcolsep=0.7mm
\renewcommand\arraystretch{1.5}
\resizebox{\linewidth}{!}{

\begin{tabular}{ccccccccccccccccccccccccccccccccccccc}
\toprule
\multicolumn{2}{c}{} & \quad\quad & 
\multicolumn{13}{c}{ {Zero-Shot}} & \quad\quad & 
\multicolumn{20}{c}{ {Few-Shot (10\% Downstream Dataset)}} \\

\cmidrule{4-16}\cmidrule{18-37}
\multicolumn{2}{c}{ {Pretrain}} & & 
\multicolumn{4}{c}{\textit{ {Images}}} & & 
\multicolumn{8}{c}{{ {\textit{Time-series}}}} & & 
\multicolumn{5}{c}{{ {\textit{Text}}}} & & 
\multicolumn{14}{c}{ {\textit{No Pretrain}}} \\

\cmidrule{4-7}\cmidrule{9-16}\cmidrule{18-22}\cmidrule{24-37}

\multicolumn{2}{c}{ {Method}} & & 
\multicolumn{2}{c}{ {\myformer}} & & 
\multicolumn{2}{c}{ {VisionTS}} & & 
\multicolumn{2}{c}{ {Moirai$_\textit{S}$}} & & 
\multicolumn{2}{c}{ {Moirai$_\textit{B}$}} & & 
\multicolumn{2}{c}{ {Moirai$_\textit{L}$}} & & 
\multicolumn{2}{c}{ {TimeLLM}} & & 
\multicolumn{2}{c}{ {GPT4TS}} & & 
\multicolumn{2}{c}{ {DLinear}} & & 
\multicolumn{2}{c}{ {PatchTST}} & & 
\multicolumn{2}{c}{ {TimesNet}} & & 
\multicolumn{2}{c}{ {Autoformer}} & 
\multicolumn{2}{c}{ {Informer}} \\

\cmidrule{4-5}\cmidrule{7-8}\cmidrule{10-11}\cmidrule{13-14}\cmidrule{16-17}\cmidrule{19-20}\cmidrule{22-23}\cmidrule{25-26}\cmidrule{28-29}\cmidrule{31-32}\cmidrule{34-37}

\multicolumn{2}{c}{ {Metric}} 

& & MSE & MAE & & MSE & MAE & & MSE & MAE & & MSE & MAE & & MSE & MAE & & MSE & MAE & & MSE & MAE & & MSE & MAE & & MSE & MAE & & MSE & MAE & & MSE & MAE & MSE & MAE \\

\midrule
\multirow{5}[2]{*}{\rotatebox{90}{$ETTh1$}} & 
\multicolumn{1}{r}{96} & & 
0.358 & 0.383 & &
 {0.353} &  {0.383} & & 
0.375 & 0.402 & & 
0.384 & 0.402 & & 
0.380 & 0.398 & & 
0.448 & 0.460 & & 
0.458 & 0.456 & & 0.492  & 0.495  &      & 0.516  & 0.485  &      & 0.861  & 0.628  &      & 0.613  & 0.552  & 1.179  & 0.792  \\
     
     & \multicolumn{1}{r}{192} & & 0.373 & 0.395 & &  {0.392} &  {0.410} &      & 0.399  & 0.419  &      & 0.425  & 0.429  &      & 0.440  & 0.434  &      & 0.484  & 0.483  &      & 0.570  & 0.516  &      & 0.565  & 0.538  &      & 0.598  & 0.524  &      & 0.797  & 0.593  &      & 0.722  & 0.598  &       1.199  & 0.806  \\
     & \multicolumn{1}{r}{336} & & 0.391 & 0.408 & &  {0.407} &  {0.423} &      & 0.412  & 0.429  &      & 0.456  & 0.450  &      & 0.514  & 0.474  &      & 0.589  & 0.540  &      & 0.608  & 0.535  &      & 0.721  & 0.622  &      & 0.657  & 0.550  &      & 0.941  & 0.648  &      & 0.750  & 0.619  &       1.202  & 0.811  \\
     & \multicolumn{1}{r}{720} & & 0.394 & 0.430 & &  {0.406} &  {0.441} &      & 0.413  & 0.444  &      & 0.470  & 0.473  &      & 0.705  & 0.568  &      & 0.700  & 0.604  &      & 0.725  & 0.591  &      & 0.986  & 0.743  &      & 0.762  & 0.610  &      & 0.877  & 0.641  &      & 0.721  & 0.616  &       1.217  & 0.825  \\
     & \multicolumn{1}{r}{avg} & & 0.379 & 0.405 & &  {0.390} &  {0.414} &      & 0.400  & 0.424  &      & 0.434  & 0.439  &      & 0.510  & 0.469  &      & 0.556  & 0.522  &      & 0.590  & 0.525  &      & 0.691  & 0.600  &      & 0.633  & 0.542  &      & 0.869  & 0.628  &      & 0.702  & 0.596  &       1.199  & 0.809  \\
\midrule
\multirow{5}[2]{*}{\rotatebox{90}{$ETTh2$}} & 
\multicolumn{1}{r}{96} & & 0.267 & 0.317 & &  {0.271} & 0.328  &      & 0.281  & 0.334  &      & 0.277  & 0.327  &      & 0.287  &  {0.325} &      & 0.275  & 0.326  &      & 0.331  & 0.374  &      & 0.357  & 0.411  &      & 0.353  & 0.389  &      & 0.378  & 0.409  &      & 0.413  & 0.451  &  3.837  & 1.508  \\
     & \multicolumn{1}{r}{192} & & 0.298 & 0.353 & &  {0.328} &  {0.367} &      & 0.340  & 0.373  &      & 0.340  & 0.374  &      & 0.347  &  {0.367} &      & 0.374  & 0.373  &      & 0.402  & 0.411  &      & 0.569  & 0.519  &      & 0.403  & 0.414  &      & 0.490  & 0.467  &      & 0.474  & 0.477  &    3.856  & 1.513  \\
     & \multicolumn{1}{r}{336} & & 0.323 & 0.374 & &  {0.345} &  {0.381} &      & 0.362  & 0.393  &      & 0.371  & 0.401  &      & 0.377  & 0.393  &      & 0.406  & 0.429  &      & 0.406  & 0.433  &      & 0.671  & 0.572  &      & 0.426  & 0.441  &      & 0.537  & 0.494  &      & 0.547  & 0.543  &       3.952  & 1.526  \\
     & \multicolumn{1}{r}{720} & & 0.346 & 0.405 & &  {0.388} & 0.422  &      & 0.380  & 0.416  &      & 0.394  & 0.426  &      & 0.404  &  {0.421} &      & 0.427  & 0.449  &      & 0.449  & 0.464  &      & 0.824  & 0.648  &      & 0.477  & 0.480  &      & 0.510  & 0.491  &      & 0.516  & 0.523  &     3.842  & 1.503  \\
     & \multicolumn{1}{r}{avg} & & 0.308 & 0.362 & &  {0.333} &  {0.375} &      & 0.341  & 0.379  &      & 0.346  & 0.382  &      & 0.354  & 0.377  &      & 0.370  & 0.394  &      & 0.397  & 0.421  &      & 0.605  & 0.538  &      & 0.415  & 0.431  &      & 0.479  & 0.465  &      & 0.488  & 0.499  &      3.872  & 1.513  \\
\midrule
\multirow{5}[2]{*}{\rotatebox{90}{$ETTm1$}} & 
\multicolumn{1}{r}{96} & & 0.343 & 0.345 & &  {0.341} &  {0.347} &      & 0.404  & 0.383  &      & 0.335  & 0.360  &      & 0.353  & 0.363  &      & 0.346  & 0.388  &      & 0.390  & 0.404  &      & 0.352  & 0.392  &      & 0.410  & 0.419  &      & 0.583  & 0.501  &      & 0.774  & 0.614  &       1.162  & 0.785  \\
     & \multicolumn{1}{r}{192} & & 0.358 & 0.357 & &  {0.360} &  {0.360} &      & 0.435  & 0.402  &      & 0.366  & 0.379  &      & 0.376  & 0.380  &      & 0.373  & 0.416  &      & 0.429  & 0.423  &      & 0.382  & 0.412  &      & 0.437  & 0.434  &      & 0.630  & 0.528  &      & 0.754  & 0.592  &       1.172  & 0.793  \\
     & \multicolumn{1}{r}{336} & & 0.372 & 0.364 & &  {0.377} &  {0.374} &      & 0.462  & 0.416  &      & 0.391  & 0.394  &      & 0.399  & 0.395  &      & 0.413  & 0.426  &      & 0.469  & 0.439  &      & 0.419  & 0.434  &      & 0.476  & 0.454  &      & 0.725  & 0.568  &      & 0.869  & 0.677  &       1.227  & 0.908  \\
     & \multicolumn{1}{r}{720} & & 0.401 & 0.393 & &  {0.416} &  {0.405} &      & 0.490  & 0.437  &      & 0.434  & 0.419  &      & 0.432  & 0.417  &      & 0.485  & 0.476  &      & 0.569  & 0.498  &      & 0.490  & 0.477  &      & 0.681  & 0.556  &      & 0.769  & 0.549  &      & 0.810  & 0.630  &       1.207  & 0.797  \\
     & \multicolumn{1}{r}{avg} & & 0.368 & 0.365 & &  {0.374} &  {0.372} &      & 0.448  & 0.410  &      & 0.382  & 0.388  &      & 0.390  & 0.389  &      & 0.404  & 0.427  &      & 0.464  & 0.441  &      & 0.411  & 0.429  &      & 0.501  & 0.466  &      & 0.677  & 0.537  &      & 0.802  & 0.628  &       1.192  & 0.821  \\
\midrule
\multirow{5}[2]{*}{\rotatebox{90}{$ETTm2$}} & 
\multicolumn{1}{r}{96} & & 0.220 & 0.264 & & 0.228  & 0.282  &      & 0.205  & 0.282  &      & 0.195  & 0.269  &      &  {0.189} &  {0.260} &      & 0.177  & 0.261  &      & 0.188  & 0.269  &      & 0.213  & 0.303  &      & 0.191  & 0.274  &      & 0.212  & 0.285  &      & 0.352  & 0.454  &       3.203  & 1.407  \\
     & \multicolumn{1}{r}{192} & & 0.244 & 0.276 & & 0.262  & 0.305  &      & 0.261  & 0.318  &      & 0.247  & 0.303  &      &  {0.247} &  {0.300} &      & 0.241  & 0.314  &      & 0.251  & 0.309  &      & 0.278  & 0.345  &      & 0.252  & 0.317  &      & 0.270  & 0.323  &      & 0.694  & 0.691  &       3.112  & 1.387  \\
     & \multicolumn{1}{r}{336} & & 0.268 & 0.293 & & 0.293  &  {0.328} &      & 0.319  & 0.355  &      &  {0.291} & 0.333  &      & 0.295  & 0.334  &      & 0.274  & 0.327  &      & 0.307  & 0.346  &      & 0.338  & 0.385  &      & 0.306  & 0.353  &      & 0.323  & 0.353  &      & 2.408  & 1.407  &       3.255  & 1.421  \\
     & \multicolumn{1}{r}{720} & & 0.325 & 0.331 & &  {0.343} &  {0.370} &      & 0.415  & 0.410  &      & 0.355  & 0.377  &      & 0.372  & 0.386  &      & 0.417  & 0.390  &      & 0.426  & 0.417  &      & 0.436  & 0.440  &      & 0.433  & 0.427  &      & 0.474  & 0.449  &      & 1.913  & 1.166  &       3.909  & 1.543  \\
     & \multicolumn{1}{r}{avg} & & 0.264 & 0.291 & & 0.282  & 0.321  &      & 0.300  & 0.341  &      &  {0.272} & 0.321  &      & 0.276  &  {0.320} &      & 0.277  & 0.323  &      & 0.293  & 0.335  &      & 0.316  & 0.368  &      & 0.296  & 0.343  &      & 0.320  & 0.353  &      & 1.342  & 0.930  &       3.370  & 1.440  \\
\midrule
\multirow{5}[2]{*}{\rotatebox{90}{$Electricity$}} & 
\multicolumn{1}{r}{96} & & 0.166 & 0.251 & & 0.177  & 0.266  &      & 0.205  & 0.299  &      & 0.158  & 0.248  &      & 0.152  & 0.242  &      &  {0.139} & 0.241  &      &  {0.139} &  {0.237} &      & 0.150  & 0.253  &      & 0.140  & 0.238  &      & 0.299  & 0.373  &      & 0.261  & 0.348  &       1.259  & 0.919  \\
     & \multicolumn{1}{r}{192} & & 0.174 & 0.270 & & 0.188  & 0.277  &      & 0.220  & 0.310  &      & 0.174  & 0.263  &      & 0.171  & 0.259  &      &  {0.151} &  {0.248} &      & 0.156  & 0.252  &      & 0.164  & 0.264  &      & 0.160  & 0.255  &      & 0.305  & 0.379  &      & 0.338  & 0.406  &       1.160  & 0.873  \\
     & \multicolumn{1}{r}{336} & & 0.198 & 0.272 & & 0.207  & 0.296  &      & 0.236  & 0.323  &      & 0.191  & 0.278  &      & 0.192  & 0.278  &      &  {0.169} &  {0.270} &      & 0.175  &  {0.270} &      & 0.181  & 0.282  &      & 0.180  & 0.276  &      & 0.319  & 0.391  &      & 0.410  & 0.474  &       1.157  & 0.872  \\
     & \multicolumn{1}{r}{720} & & 0.231 & 0.295 & & 0.256  & 0.337  &      & 0.270  & 0.347  &      & 0.229  &  {0.307} &      & 0.236  & 0.313  &      & 0.240  & 0.322  &      &  {0.233} & 0.317  &      & 0.223  & 0.321  &      & 0.241  & 0.323  &      & 0.369  & 0.426  &      & 0.715  & 0.685  &       1.203  & 0.898  \\
     & \multicolumn{1}{r}{avg} & & 0.193 & 0.272 & & 0.207  & 0.294  &      & 0.233  & 0.320  &      & 0.188  & 0.274  &      & 0.188  & 0.273  &      &  {0.175} & 0.270  &      & 0.176  &  {0.269} &      & 0.180  & 0.280  &      & 0.180  & 0.273  &      & 0.323  & 0.392  &      & 0.431  & 0.478  &       1.195  & 0.891  \\
\midrule
\multirow{5}[2]{*}{\rotatebox{90}{$Weather$}} & 
\multicolumn{1}{r}{96} & & 0.157 & 0.197 & & 0.220  & 0.257  &      & 0.173  & 0.212  &      & 0.167  &  {0.203} &      & 0.177  & 0.208  &      &  {0.161} & 0.210  &      & 0.163  & 0.215  &      & 0.171  & 0.224  &      & 0.165  & 0.215  &      & 0.184  & 0.230  &      & 0.221  & 0.297  &       0.374  & 0.401  \\
     & \multicolumn{1}{r}{192} & & 0.193 & 0.226 & & 0.244  & 0.275  &      & 0.216  & 0.250  &      & 0.209  &  {0.241} &      & 0.219  & 0.249  &      &  {0.204} & 0.248  &      & 0.210  & 0.254  &      & 0.215  & 0.263  &      & 0.210  & 0.257  &      & 0.245  & 0.283  &      & 0.270  & 0.322  &       0.552  & 0.478  \\
     & \multicolumn{1}{r}{336} & & 0.241 & 0.262 & & 0.280  & 0.299  &      & 0.260  & 0.282  &      &  {0.256} &  {0.276} &      & 0.277  & 0.292  &      & 0.261  & 0.302  &      &  {0.256} & 0.292  &      & 0.258  & 0.299  &      & 0.259  & 0.297  &      & 0.305  & 0.321  &      & 0.320  & 0.351  &       0.724  & 0.541  \\
     & \multicolumn{1}{r}{720} & & 0.293 & 0.314 & & 0.330  & 0.337  &      & 0.320  & 0.322  &      & 0.321  &  {0.323} &      & 0.365  & 0.350  &      &  {0.309} & 0.332  &      & 0.321  & 0.339  &      & 0.320  & 0.346  &      & 0.332  & 0.346  &      & 0.381  & 0.371  &      & 0.390  & 0.396  &       0.739  & 0.558  \\
     & \multicolumn{1}{r}{avg} & & 0.221 & 0.249 & & 0.269  & 0.292  &      & 0.242  & 0.267  &      & 0.238  &  {0.261} &      & 0.260  & 0.275  &      &  {0.234} & 0.273  &      & 0.238  & 0.275  &      & 0.241  & 0.283  &      & 0.242  & 0.279  &      & 0.279  & 0.301  &      & 0.300  & 0.342  &       0.597  & 0.495  \\
\midrule
\multicolumn{2}{c}{ {Average}} & & 0.288 & 0.324 & &  {0.309} & 0.345  &      & 0.327  & 0.357  &      & 0.310  &  {0.344} &      & 0.329  & 0.350  &      & 0.336  & 0.368  &      & 0.360  & 0.378  &      & 0.407  & 0.416  &      & 0.378  & 0.389  &      & 0.491  & 0.446  &      & 0.678  & 0.579  &       1.904  & 0.995  \\
\bottomrule
\end{tabular}%
}
\end{table*}%

\paragraph{Zero-shot forecasting results on the long-term TSF benchmark.}
Table~\ref{tab:task3_zero_longtsf_all} shows the full results of zero-shot/few-shot long-term forecasting performance. To ensure the fairness of the experiment, we adopted the context configuration of VisionTS and also presented the results reported by them from previous paper.

\paragraph{Comparison of TimesFM and LLMTime on the long-term TSF benchmark.}
We also compared two decoder-only baselines, including TimesFM trained end-to-end from the temporal domain and LLMTime based on corpus pretraining. Due to the step-by-step output nature of autoregressive forecasting, both TimesFM and LLMTime exhibit slower prediction efficiency, and thus we only report partial test window results in the table~\ref{tab:task3_zero_timesfm_llmtime}. The proposed TimeArtist demonstrates superior performance across most metrics.

\begin{table*}[htpb]
  \centering
  \caption{Comparison of traditional approaches on the zero-shot forecasting benchmarks.}\label{tab:task3_zero_tradition}
  \tabcolsep=1.8mm
  \resizebox{0.8\linewidth}{!}{
    \begin{tabular}{ccccccccccccccccccccc}
    \toprule
    \multicolumn{2}{c}{ {Method}} & & 
    \multicolumn{2}{c}{ {TimeArtist}} & & 
    \multicolumn{2}{c}{ {VisionTS}} & & 
    \multicolumn{2}{c}{ {ETS}} & & 
    \multicolumn{2}{c}{ {ARIMA}} & & 
    \multicolumn{2}{c}{ {Seasonal Naive}} & & 
    \multicolumn{2}{c}{ {Seasonal Avg}} \\
    
    \cmidrule{4-5}\cmidrule{7-8}\cmidrule{10-11}\cmidrule{13-14}\cmidrule{16-17}\cmidrule{19-20}
    
    \multicolumn{2}{c}{ {Metric}} & & 
    MSE & MAE & & MSE & MAE & & MSE & MAE & & MSE & MAE & & MSE & MAE & & MSE & MAE \\
    \midrule
    \multirow{5}[1]{*}{\rotatebox{90}{$ETTh1$}} 
    & \multicolumn{1}{r}{96} & & 0.358 & 0.383 & & 0.353 & 0.383 & & 1.289  & 0.710  &      & 0.900  & 0.719  &      & 0.512  & 0.433  &      & 0.589  & 0.585  \\
         & \multicolumn{1}{r}{192} & & 0.373 & 0.395 &      &  {0.392} &  {0.410} &      & 1.319  & 0.730  &      & 0.906  & 0.724  &      & 0.581  & 0.469  &      & 0.598  & 0.590  \\
         & \multicolumn{1}{r}{336} & & 0.391 & 0.408 &      &  {0.407} &  {0.423} &      & 1.324  & 0.742  &      & 0.908  & 0.731  &      & 0.650  & 0.501  &      & 0.610  & 0.597  \\
         & \multicolumn{1}{r}{720} & & 0.394 & 0.430 &      &  {0.406} &  {0.441} &      & 1.329  & 0.751  &      & 0.932  & 0.753  &      & 0.655  & 0.514  &      & 0.656  & 0.624  \\
         & \multicolumn{1}{r}{avg} & & 0.379 & 0.405 &      &  {0.390} &  {0.414} &      & 1.315  & 0.733  &      & 0.912  & 0.732  &      & 0.600  & 0.479  &      & 0.613  & 0.599  \\
    \midrule
    \multirow{5}[1]{*}{\rotatebox{90}{$ETTh2$}} 
    & \multicolumn{1}{r}{96} & & 0.267 & 0.317 &      &  {0.271} &  {0.328} &      & 0.399  & 0.408  &      & 0.488  & 0.508  &      & 0.391  & 0.380  &      & 0.457  & 0.494  \\
         & \multicolumn{1}{r}{192} & & 0.298 & 0.353 &      &  {0.328} &  {0.367} &      & 0.500  & 0.459  &      & 0.497  & 0.514  &      & 0.482  & 0.429  &      & 0.466  & 0.500  \\
         & \multicolumn{1}{r}{336} & & 0.323 & 0.374 &      &  {0.345} &  {0.381} &      & 0.562  & 0.498  &      & 0.507  & 0.522  &      & 0.532  & 0.466  &      & 0.476  & 0.509  \\
         & \multicolumn{1}{r}{720} & & 0.346 & 0.405 &      &  {0.388} &  {0.422} &      & 0.558  & 0.506  &      & 0.572  & 0.557  &      & 0.525  & 0.474  &      & 0.542  & 0.548  \\
         & \multicolumn{1}{r}{avg} & & 0.308 & 0.362 &      &  {0.333} &  {0.375} &      & 0.505  & 0.468  &      & 0.516  & 0.525  &      & 0.483  & 0.437  &      & 0.485  & 0.513  \\
    \midrule
    \multirow{5}[1]{*}{\rotatebox{90}{$ETTm1$}} 
    & \multicolumn{1}{r}{96} & & 0.343 & 0.345 &      &  {0.341} &  {0.347} &      & 1.204  & 0.659  &      & 0.702  & 0.568  &      & 0.423  & 0.387  &      & 0.369  & 0.399  \\
         & \multicolumn{1}{r}{192} & & 0.358 & 0.357 &      &  {0.360} &  {0.360} &      & 1.251  & 0.685  &      & 0.704  & 0.570  &      & 0.463  & 0.406  &      & 0.374  & 0.402  \\
         & \multicolumn{1}{r}{336} & & 0.372 & 0.364 &      &  {0.377} &  {0.374} &      & 1.276  & 0.702  &      & 0.709  & 0.574  &      & 0.496  & 0.426  &      & 0.382  & 0.407  \\
         & \multicolumn{1}{r}{720} & & 0.401 & 0.393 &      & 0.416  &  {0.405} &      & 1.311  & 0.724  &      & 0.713  & 0.580  &      & 0.574  & 0.464  &      &  {0.394} & 0.416  \\
         & \multicolumn{1}{r}{avg} & & 0.368 & 0.365 &      &  {0.374} &  {0.372} &      & 1.261  & 0.693  &      & 0.707  & 0.573  &      & 0.489  & 0.421  &      & 0.380  & 0.406  \\
    \midrule
    \multirow{5}[1]{*}{\rotatebox{90}{$ETTm2$}} 
    & \multicolumn{1}{r}{96} & & 0.220 & 0.264 &      &  {0.228} &  {0.282} &      & 0.257  & 0.324  &      & 0.397  & 0.434  &      & 0.263  & 0.301  &      & 0.365  & 0.411  \\
         & \multicolumn{1}{r}{192} & & 0.244 & 0.276 &      &  {0.262} &  {0.305} &      & 0.331  & 0.366  &      & 0.402  & 0.436  &      & 0.321  & 0.337  &      & 0.369  & 0.414  \\
         & \multicolumn{1}{r}{336} & & 0.268 & 0.293 &      &  {0.293} &  {0.328} &      & 0.402  & 0.406  &      & 0.407  & 0.439  &      & 0.376  & 0.370  &      & 0.375  & 0.418  \\
         & \multicolumn{1}{r}{720} & & 0.325 & 0.331 &      &  {0.343} &  {0.370} &      & 0.512  & 0.462  &      & 0.413  & 0.443  &      & 0.471  & 0.422  &      & 0.380  & 0.423  \\
         & \multicolumn{1}{r}{avg} & & 0.264 & 0.291 &      &  {0.282} &  {0.321} &      & 0.376  & 0.390  &      & 0.405  & 0.438  &      & 0.358  & 0.357  &      & 0.372  & 0.417  \\
    \midrule
    \multicolumn{2}{c}{ {Average}} & & 0.330 & 0.356 &      &  {0.344} &  {0.370} &      & 0.864  & 0.571  &      & 0.635  & 0.567  &      & 0.482  & 0.424  &      & 0.463  & 0.484  \\
    \bottomrule
    \end{tabular}%
    }
\end{table*}%

\paragraph{Comparison of traditional methods.}
Besides deep learning-based models, we compared the performance of traditional machine learning approaches on zero-shot Long-Term TSF scenarios. Table 1 demonstrates multiple baselines, including classical approaches such as ARIMA and ETS. Furthermore, we incorporated two periodicity-based methods~\citep{chen2024visionts}: Seasonal Naive and Average.

\begin{table*}[htpb]
  \centering
  \caption{Full zero-shot forecasting results (MAE) on the Monash TSF benchmark. We reported the reproduction results of LLMTime based on the GPT3.5 API from Moirai~\citep{woo2024unified}.}\label{tab:task3_monash}
    \resizebox{\linewidth}{!}{
    \begin{tabular}{lcccccccccccccccc}
    \toprule
    & {TimeArtist} & {VisionTS} & {LLMTime} &  {Moirai(S)} &  {Naive} &  {SES} &  {Theta} &  {TBATS} &  {ETS} &  {(DHR-)ARIMA} & {PR} &  {CatBoost} &  {FFNN} &  {DeepAR} &  {N-BEATS} &  {WaveNet} \\
    \midrule
    
    M1 Monthly & 1894.73 & 1987.69 & 2562.84 & 2082.26 & 2707.75 & 2259.04 & 2166.18 & 2237.5 & 1905.28 & 2080.13 & 2088.25 & 2052.32 & 2162.58 & 1860.81 &  {1820.37} & 2184.42 \\
    
    M3 Monthly & 698.52 & 737.93 & 877.97 & 713.41 & 837.14 & 743.41 &  {623.71} & 630.59 & 626.46 & 654.8 & 692.97 & 732  & 692.48 & 728.81 & 648.6 & 699.3 \\
    
    M3 Other & 208.47 & 315.85 & 300.3 & 263.54 & 278.43 & 277.83 & 215.35 &  {189.42} & 194.98 & 193.02 & 234.43 & 318.13 & 240.17 & 247.56 & 221.85 & 245.29 \\
    
    M4 Monthly & 489.3 & 666.54 & 728.27 & 597.6 & 671.27 & 625.24 &  {563.58} & 589.52 & 582.6 & 575.36 & 596.19 & 611.69 & 612.52 & 615.22 & 578.48 & 655.51 \\
    
    M4 Weekly & 506.8 & 404.23 & 518.44 & 339.76 & 347.99 & 336.82 & 333.32 & 296.15 & 335.66 & 321.61 & 293.21 & 364.65 & 338.37 & 351.78 &  {277.73} & 359.46 \\
    
    M4 Daily & 223.6 & 215.63 & 266.52 & 189.1 & 180.83 & 178.27 & 178.86 &  {176.6} & 193.26 & 179.67 & 181.92 & 231.36 & 177.91 & 299.79 & 190.44 & 189.47 \\
    
    M4 Hourly & 237.5 & 288.37 & 576.06 & 268.04 & 1218.06 & 1218.06 & 1220.97 & 386.27 & 3358.1 & 1310.85 &  {257.39} & 285.35 & 385.49 & 886.02 & 425.75 & 393.63 \\
    
    Tourism Quarterly & 10559.62 & 12931.88 & 16918.86 & 18352.44 & 15845.1 & 15014.19 &  {7656.49} & 9972.42 & 8925.52 & 10475.47 & 9092.58 & 10267.97 & 8981.04 & 9511.37 & 8640.56 & 9137.12 \\
    
    Tourism Monthly & 1593.42 & 2560.19 & 5608.61 & 3569.85 & 5636.83 & 5302.1 & 2069.96 & 2940.08 & 2004.51 & 2536.77 & 2187.28 & 2537.04 & 2022.21 &  {1871.69} & 2003.02 & 2095.13 \\
    
    CIF 2016 & 514783.84 & 570907.24  & 599313.8 & 655888.58 & 578596.5 & 581875.97 & 714818.6 & 855578.4 & 642421.4 & {469059} & 563205.57 & 603551.3 & 1495923 & 3200418 & 679034.8 & 5998225 \\
    
    Aus. Elec. Demand & 207.39 & 237.44 & 760.81 & 266.57 & 659.6 & 659.6 & 665.04 & 370.74 & 1282.99 & 1045.92 & 247.18 & 241.77 & 258.76 & 302.41 &  {213.83} & 227.5 \\
    
    Bitcoin & 1.07E+18 & 2.33E+18 & 1.74E+18 & 1.76E+18 & 7.78E+17 & 5.33E+18 & 5.33E+18 & 9.9E+17 & 1.1E+18 & 3.62E+18 &  {6.66E+17} & 1.93E+18 & 1.45E+18 & 1.95E+18 & 1.06E+18 & 2.46E+18 \\
    
    Pedestrian Counts & 87.46 & 52.01 & 97.77 & 54.88 & 170.88 & 170.87 & 170.94 & 222.38 & 216.5 & 635.16 & 44.18 &  {43.41} & 46.41 & 44.78 & 66.84 & 46.46 \\
    
    Vehicle Trips & 35.29 & 22.08 & 31.48 & 24.46 & 31.42 & 29.98 & 30.76 &  {21.21} & 30.95 & 30.07 & 27.24 & 22.61 & 22.93 & 22 & 28.16 & 24.15 \\
    
    KDD cup & 49.08 & 38.16 & 42.72 & 39.81 & 42.13 & 42.04 & 42.06 & 39.2 & 44.88 & 52.2 & 36.85 &  {34.82} & 37.16 & 48.98 & 49.1 & 37.08 \\
    
    Weather & 2.25 & 2.06 & 2.17 & {1.96} & 2.36 & 2.24 & 2.51 & 2.3  & 2.35 & 2.45 & 8.17 & 2.51 & 2.09 & 2.02 & 2.34 & 2.29 \\
    
    NN5 Daily & 3.25 & {3.51} & 7.1  & 5.37 & 8.26 & 6.63 & 3.8  & 3.7  & 3.72 & 4.41 & 5.47 & 4.22 & 4.06 & 3.94 & 4.92 & 3.97 \\
    
    NN5 Weekly & 13.73 & 14.67 & 15.76 & 15.07 & 16.71 & 15.66 & 15.3 & 14.98 & 15.7 & 15.38 & 14.94 & 15.29 & 15.02 & 14.69 &  {14.19} & 19.34 \\
    
    Carparts & 0.62 & 0.58 & 0.44 & 0.53 & 0.65 & 0.55 & 0.53 & 0.58 & 0.56 & 0.56 & 0.41 & 0.53 &  {0.39} &  {0.39} & 0.98 & 0.4 \\
    
    FRED-MD & 2138.54 & {1893.67} & 2804.64 & 2568.48 & 2825.67 & 2798.22 & 3492.84 & 1989.97 & 2041.42 & 2957.11 & 8921.94 & 2475.68 & 2339.57 & 4264.36 & 2557.8 & 2508.4 \\
    
    Traffic Hourly & 0.01 &  {0.01} & 0.03 & 0.02 & 0.03 & 0.03 & 0.03 & 0.04 & 0.03 & 0.04 & 0.02 & 0.02 &  {0.01} &  {0.01} & 0.02 & 0.02 \\
    
    Traffic Weekly & 1.12 & 1.14 & 1.15 & 1.17 & 1.19 & 1.12 & 1.13 & 1.17 & 1.14 & 1.22 & 1.13 & 1.17 & 1.15 & 1.18 &  {1.11} & 1.2 \\
    
    Rideshare & 2.59 & 5.92 & 6.28 & {1.35} & 6.29 & 6.29 & 7.62 & 6.45 & 6.29 & 3.37 & 6.3  & 6.07 & 6.59 & 6.28 & 5.55 & 2.75 \\
    
    Hospital & 18.04 & 19.36 & 25.68 & 23   & 24.07 & 21.76 & 18.54 &  {17.43} & 17.97 & 19.6 & 19.24 & 19.17 & 22.86 & 18.25 & 20.18 & 19.35 \\
    
    COVID Deaths & 269.7 & 137.51 & 653.31 & 124.32 & 353.71 & 353.71 & 321.32 & 96.29 &  {85.59} & 85.77 & 347.98 & 475.15 & 144.14 & 201.98 & 158.81 & 1049.48  \\
    
    Temperature Rain & 5.83 & 6.37 & 6.37 & 5.3  & 9.39 & 8.18 & 8.22 & 7.14 & 8.21 & 7.19 & 6.13 & 6.76 & 5.56 & 5.37 & 7.28 & 5.81 \\
    
    Sunspot & 2.72 & 2.81 & 5.07 &  {0.11} & 3.93 & 4.93 & 4.93 & 2.57 & 4.93 & 2.57 & 3.83 & 2.27 & 7.97 & 0.77 & 14.47 & 0.17 \\
    
    Saugeen River Flow & 23.70 & 30.22 & 34.84 & 24.07 & 21.5 & 21.5 & 21.49 & 22.26 & 30.69 & 22.38 & 25.24 &  {21.28} & 22.98 & 23.51 & 27.92 & 22.17 \\
    
    US Births & 473.6 & 519.94 & 1374.99 & 872.51 & 1152.67 & 1192.2 & 586.93 &  {399} & 419.73 & 526.33 & 574.93 & 441.7 & 557.87 & 424.93 & 422  & 504.4 \\
    
    \midrule
    
     {Normalized MAE} & 0.680 & 0.729  & 1.041  & 0.657  & 1.000  & 1.028  & 0.927  & 0.758  & 0.872  & 0.898  & 0.785  & 0.760  & 0.741  & 0.759  & 0.783  & 0.749 \\
    
     {Rank} & 2 & 3    & 16   & 1    & 14   & 15   & 13   & 6    & 11   & 12   & 10   & 8    & 4    & 7    & 9    & 5  \\
    \bottomrule
    \end{tabular}%
}
\end{table*}%

\paragraph{Zero-shot forecasting results of the Monash TSF benchmark.}
In the table~\ref{tab:task3_monash}, we evaluated the MAE results of \myformer and other baselines on the Monash~\citep{godahewa2021monash} dataset. 
Additionally, we also presented the standardized MAE metric, which is calculated by dividing the MAE of each dataset by the MAE value of the simple prediction.

\begin{table*}[htpb]
  \centering
  \caption{Configurations of \myformer utilized in our full-shot TSF by outpainting.}\label{tab:task3_full_hyper}
  \small
    \begin{tabular}{ccccccc}
    \toprule
    & \textbf{ETTh1} & \textbf{ETTh2} & \textbf{ETTm1} & \textbf{ETTm2} & \textbf{Weather} & \textbf{Electricity} \\
    \midrule
    Observation-image Width & 256 & 256 & 256 & 256 & 256 & 256 \\
    Forecasting-image Width & 768 & 768 & 768 & 768 & 768 & 768 \\
    Context length $L$ & 1152 & 1152 & 2304 & 1152 & 576 & 1152 \\
    \bottomrule
    \end{tabular}
\end{table*}%

\paragraph{Hyperparameters of full-shot forecasting.}
In the full sample prediction scenario, we froze the two-stage pre-trained dual-autoencoder and shared-quantizer components as well as the visual model, and only used the dataset from the evaluation domain to perform end-to-end full-scale fine-tuning of the alignment component. Thanks to the freezing of most parameters, the carefully trained representation space was retained in the downstream tasks. We only trained the alignment component to ensure that the visual model could learn meaningful visual semantic information from the observed sequence. Additionally, in the full sample prediction, we followed the channel-independent design, set the learning rate to E-4, the batch size to 256, and adopted the Adam optimizer to train the alignment. We presented the results of three repetitions in table~\ref{tab:task3_full_hyper}.

\begin{table*}[htpb]
\centering
\caption{Full-shot forecasting performance on the long-term TSF benchmark.}\label{tab:task3_full_longtsf_all}
\tabcolsep=0.7mm
\renewcommand\arraystretch{1.5}
\resizebox{\linewidth}{!}{
\begin{tabular}{cccccccccccccccccccccccccccccccccccccc}
\toprule
\multicolumn{2}{c}{ {Pretrain}} & & 
\multicolumn{5}{c}{\textit{ {Images}}} & & 
\multicolumn{5}{c}{\textit{ {Text}}} & & 
\multicolumn{23}{c}{\textit{ {No Pretrain}}} \\

\cmidrule{4-8}\cmidrule{10-14}\cmidrule{16-38}

\multicolumn{2}{c}{ {Method}} & & 
\multicolumn{2}{c}{ {\myformer}} & & 
\multicolumn{2}{c}{ {VisionTS}} & & 
\multicolumn{2}{c}{ {Time-LLM}} & & 
\multicolumn{2}{c}{ {GPT4TS}} & & 
\multicolumn{2}{c}{ {DLinear}} & & 
\multicolumn{2}{c}{ {PatchTST}} & & 
\multicolumn{2}{c}{ {TimesNet}} & & 
\multicolumn{2}{c}{ {FEDformer}} & & 
\multicolumn{2}{c}{ {Autoformer}} & & 
\multicolumn{2}{c}{ {Stationary}} & & 
\multicolumn{2}{c}{ {ETSformer}} & & 
\multicolumn{2}{c}{ {Informer}} \\

\multicolumn{2}{c}{ {Metric}} & & MSE & MAE & & MSE & MAE & & MSE & MAE & & MSE & MAE & & MSE & MAE & & MSE & MAE & & MSE & MAE & & MSE & MAE & & MSE & MAE & & MSE & MAE & & MSE & MAE & & MSE & MAE \\
\midrule
\multirow{5}[2]{*}{\rotatebox{90}{$ETTh1$}} & \multicolumn{1}{r}{96} & & {0.339} & {0.371} &      &  {0.347} &  {0.376} &      & 0.376  & 0.402  &      & 0.370  & 0.389  &      & 0.375  & 0.399  &      & 0.370  & 0.399  &      & 0.384  & 0.402  &      & 0.376  & 0.419  &      & 0.449  & 0.459  &      & 0.513  & 0.491  &      & 0.494  & 0.479  &      & 0.865  & 0.713  \\
     & \multicolumn{1}{r}{192} & & {0.369} & {0.385} &      &  {0.385} &  {0.400} &      & 0.407  & 0.421  &      & 0.412  & 0.413  &      & 0.405  & 0.416  &      & 0.413  & 0.421  &      & 0.436  & 0.429  &      & 0.420  & 0.448  &      & 0.500  & 0.482  &      & 0.534  & 0.504  &      & 0.538  & 0.504  &      & 1.008  & 0.792  \\
     & \multicolumn{1}{r}{336} & & {0.387} & {0.393} &      &  {0.407} &  {0.415} &      & 0.430  & 0.438  &      & 0.448  & 0.431  &      & 0.439  & 0.443  &      & 0.422  & 0.436  &      & 0.491  & 0.469  &      & 0.459  & 0.465  &      & 0.521  & 0.496  &      & 0.588  & 0.535  &      & 0.574  & 0.521  &      & 1.107  & 0.809  \\
     & \multicolumn{1}{r}{720} & & {0.390} & {0.421} &      &  {0.439} &  {0.443} &      & 0.457  & 0.468  &      & 0.441  & 0.449  &      & 0.472  & 0.490  &      & 0.447  & 0.466  &      & 0.521  & 0.500  &      & 0.506  & 0.507  &      & 0.514  & 0.512  &      & 0.643  & 0.616  &      & 0.562  & 0.535  &      & 1.181  & 0.865  \\
     & \multicolumn{1}{r}{avg} & & {0.371} & {0.393} &      &  {0.395} &  {0.409} &      & 0.418  & 0.432  &      & 0.418  & 0.421  &      & 0.423  & 0.437  &      & 0.413  & 0.431  &      & 0.458  & 0.450  &      & 0.440  & 0.460  &      & 0.496  & 0.487  &      & 0.570  & 0.537  &      & 0.542  & 0.510  &      & 1.040  & 0.795  \\
\midrule
\multirow{5}[2]{*}{\rotatebox{90}{$ETTh2$}} & \multicolumn{1}{r}{96} & & {0.260} & {0.309} &      &  {0.269} &  {0.328} &      & 0.286  & 0.346  &      & 0.280  & 0.335  &      & 0.289  & 0.353  &      & 0.274  & 0.336  &      & 0.340  & 0.374  &      & 0.358  & 0.397  &      & 0.346  & 0.388  &      & 0.476  & 0.458  &      & 0.340  & 0.391  &      & 3.755  & 1.525  \\
     & \multicolumn{1}{r}{192} & & {0.292} & {0.348} &      &  {0.332} &  {0.374} &      & 0.361  & 0.391  &      & 0.348  & 0.380  &      & 0.383  & 0.418  &      & 0.339  & 0.379  &      & 0.402  & 0.414  &      & 0.429  & 0.439  &      & 0.456  & 0.452  &      & 0.512  & 0.493  &      & 0.430  & 0.439  &      & 5.602  & 1.931  \\
     & \multicolumn{1}{r}{336} & & {0.334} & {0.365} &      & 0.351  & 0.395  &      & 0.390  & 0.414  &      & 0.380  & 0.405  &      & 0.448  & 0.465  &      &  {0.329} &  {0.380} &      & 0.452  & 0.452  &      & 0.496  & 0.487  &      & 0.482  & 0.486  &      & 0.552  & 0.551  &      & 0.485  & 0.479  &      & 4.721  & 1.835  \\
     & \multicolumn{1}{r}{720} & & {0.357} & {0.398} &      & 0.390  & 0.430  &      & 0.405  & 0.434  &      & 0.406  & 0.436  &      & 0.605  & 0.551  &      &  {0.379} &  {0.422} &      & 0.462  & 0.468  &      & 0.463  & 0.474  &      & 0.515  & 0.511  &      & 0.562  & 0.560  &      & 0.500  & 0.497  &      & 3.647  & 1.625  \\
     & \multicolumn{1}{r}{avg} & & {0.311} & {0.355} &      & 0.336  & 0.382  &      & 0.361  & 0.396  &      & 0.354  & 0.389  &      & 0.431  & 0.447  &      &  {0.330} &  {0.379} &      & 0.414  & 0.427  &      & 0.437  & 0.449  &      & 0.450  & 0.459  &      & 0.526  & 0.516  &      & 0.439  & 0.452  &      & 4.431  & 1.729  \\
\midrule
\multirow{5}[2]{*}{\rotatebox{90}{$ETTm1$}} & \multicolumn{1}{r}{96} & & {0.263} & {0.294} &      &  {0.281} &  {0.322} &      & 0.291  & 0.341  &      & 0.300  & 0.340  &      & 0.299  & 0.343  &      & 0.290  & 0.342  &      & 0.338  & 0.375  &      & 0.379  & 0.419  &      & 0.505  & 0.475  &      & 0.386  & 0.398  &      & 0.375  & 0.398  &      & 0.672  & 0.571  \\
     & \multicolumn{1}{r}{192} & & {0.307} & {0.318} &      &  {0.322} &  {0.353} &      & 0.341  & 0.369  &      & 0.343  & 0.368  &      & 0.335  & 0.365  &      & 0.332  & 0.369  &      & 0.374  & 0.387  &      & 0.426  & 0.441  &      & 0.553  & 0.496  &      & 0.459  & 0.444  &      & 0.408  & 0.410  &      & 0.795  & 0.669  \\
     & \multicolumn{1}{r}{336} & & {0.339} & {0.346} &      &  {0.356} &  {0.379} &      & 0.359  &  {0.379} &      & 0.376  & 0.386  &      & 0.369  & 0.386  &      & 0.366  & 0.392  &      & 0.410  & 0.411  &      & 0.445  & 0.459  &      & 0.621  & 0.537  &      & 0.495  & 0.464  &      & 0.435  & 0.428  &      & 1.212  & 0.871  \\
     & \multicolumn{1}{r}{720} & & {0.372} & {0.380} &      &  {0.391} &  {0.413} &      & 0.433  & 0.419  &      & 0.431  & 0.416  &      & 0.425  & 0.421  &      & 0.416  & 0.420  &      & 0.478  & 0.450  &      & 0.543  & 0.490  &      & 0.671  & 0.561  &      & 0.585  & 0.516  &      & 0.499  & 0.462  &      & 1.166  & 0.823  \\
     & \multicolumn{1}{r}{avg} & & {0.320} & {0.335} &      &  {0.338} &  {0.367} &      & 0.356  & 0.377  &      & 0.363  & 0.378  &      & 0.357  & 0.379  &      & 0.351  & 0.381  &      & 0.400  & 0.406  &      & 0.448  & 0.452  &      & 0.588  & 0.517  &      & 0.481  & 0.456  &      & 0.429  & 0.425  &      & 0.961  & 0.734  \\
\midrule
\multirow{5}[2]{*}{\rotatebox{90}{$ETTm2$}} & \multicolumn{1}{r}{96} & & {0.152} & {0.238} &      & 0.169  & 0.256  &      &  {0.162} &  {0.248} &      & 0.163  & 0.249  &      & 0.167  & 0.269  &      & 0.165  & 0.255  &      & 0.187  & 0.267  &      & 0.203  & 0.287  &      & 0.255  & 0.339  &      & 0.192  & 0.274  &      & 0.189  & 0.280  &      & 0.365  & 0.453  \\
     & \multicolumn{1}{r}{192} & & {0.213} & {0.275} &      & 0.225  & 0.294  &      & 0.235  & 0.304  &      & 0.222  &  {0.291} &      & 0.224  & 0.303  &      &  {0.220} & 0.292  &      & 0.249  & 0.309  &      & 0.269  & 0.328  &      & 0.281  & 0.340  &      & 0.280  & 0.339  &      & 0.253  & 0.319  &      & 0.533  & 0.563  \\
     & \multicolumn{1}{r}{336} & & {0.264} & {0.317} &      & 0.278  & 0.334  &      & 0.280  & 0.329  &      &  {0.273} &  {0.327} &      & 0.281  & 0.342  &      & 0.274  & 0.329  &      & 0.321  & 0.351  &      & 0.325  & 0.366  &      & 0.339  & 0.372  &      & 0.334  & 0.361  &      & 0.314  & 0.357  &      & 1.363  & 0.887  \\
     & \multicolumn{1}{r}{720} & & {0.346} & {0.361} &      & 0.372  & 0.392  &      & 0.366  & 0.382  &      &  {0.357} &  {0.376} &      & 0.397  & 0.421  &      & 0.362  & 0.385  &      & 0.408  & 0.403  &      & 0.421  & 0.415  &      & 0.433  & 0.432  &      & 0.417  & 0.413  &      & 0.414  & 0.413  &      & 3.379  & 1.338  \\
     & \multicolumn{1}{r}{avg} & & {0.244} & {0.298} &      & 0.261  & 0.319  &      & 0.261  & 0.316  &      &  {0.254} &  {0.311} &      & 0.267  & 0.334  &      & 0.255  & 0.315  &      & 0.291  & 0.333  &      & 0.305  & 0.349  &      & 0.327  & 0.371  &      & 0.306  & 0.347  &      & 0.293  & 0.342  &      & 1.410  & 0.810  \\
\midrule
\multirow{5}[2]{*}{\rotatebox{90}{$Illness$}} & \multicolumn{1}{r}{24} & & {1.475} & {0.790} &      & 2.034  & 0.937  &      & 1.792  & 0.807  &      & 1.869  & 0.823  &      & 2.215  & 1.081  &      &  {1.319} &  {0.754} &      & 2.317  & 0.934  &      & 3.228  & 1.260  &      & 3.483  & 1.287  &      & 2.294  & 0.945  &      & 2.527  & 1.020  &      & 5.764  & 1.677  \\
     & \multicolumn{1}{r}{36} & & {1.524} & {0.845} &      & 1.866  & 0.888  &      & 1.833  &  {0.833} &      & 1.853  & 0.854  &      & 1.963  & 0.963  &      &  {1.430} & 0.834  &      & 1.972  & 0.920  &      & 2.679  & 1.080  &      & 3.103  & 1.148  &      & 1.825  & 0.848  &      & 2.615  & 1.007  &      & 4.755  & 1.467  \\
     & \multicolumn{1}{r}{48} & & {1.608} & {0.873} &      & 1.784  & 0.870  &      & 2.269  & 1.012  &      & 1.886  & 0.855  &      & 2.130  & 1.024  &      &  {1.553} &  {0.815} &      & 2.238  & 0.940  &      & 2.622  & 1.078  &      & 2.669  & 1.085  &      & 2.010  & 0.900  &      & 2.359  & 0.972  &      & 4.763  & 1.469  \\
     & \multicolumn{1}{r}{60} & & {1.572} & {0.861} &      & 1.910  & 0.912  &      & 2.177  & 0.925  &      & 1.877  & 0.877  &      & 2.368  & 1.096  &      &  {1.470} &  {0.788} &      & 2.027  & 0.928  &      & 2.857  & 1.157  &      & 2.770  & 1.125  &      & 2.178  & 0.963  &      & 2.487  & 1.016  &      & 5.264  & 1.564  \\
     & \multicolumn{1}{r}{avg} & & {1.545} & {0.842} &      & 1.899  & 0.902  &      & 2.018  & 0.894  &      & 1.871  & 0.852  &      & 2.169  & 1.041  &      &  {1.443} &  {0.798} &      & 2.139  & 0.931  &      & 2.847  & 1.144  &      & 3.006  & 1.161  &      & 2.077  & 0.914  &      & 2.497  & 1.004  &      & 5.137  & 1.544  \\
\midrule
\multirow{5}[2]{*}{\rotatebox{90}{$Weather$}} & \multicolumn{1}{r}{96} & & {0.138} & {0.155} &      &  {0.142} & 0.192 &      & 0.155  & 0.199  &      & 0.148  &  {0.188} &      & 0.176  & 0.237  &      & 0.149  & 0.198  &      & 0.172  & 0.220  &      & 0.217  & 0.296  &      & 0.266  & 0.336  &      & 0.173  & 0.223  &      & 0.197  & 0.281  &      & 0.300  & 0.384  \\
     & \multicolumn{1}{r}{192} & & {0.177} & {0.213} &      &  {0.191} & 0.238 &      & 0.223  & 0.261  &      & 0.192  &  {0.230} &      & 0.220  & 0.282  &      & 0.194  & 0.241  &      & 0.219  & 0.261  &      & 0.276  & 0.336  &      & 0.307  & 0.367  &      & 0.245  & 0.285  &      & 0.237  & 0.312  &      & 0.598  & 0.544  \\
     & \multicolumn{1}{r}{336} & & {0.221} & {0.266} &      & 0.246 & 0.282 &      & 0.251  & 0.279  &      & 0.246  &  {0.273} &      & 0.265  & 0.319  &      &  {0.245} & 0.282  &      & 0.280  & 0.306  &      & 0.339  & 0.380  &      & 0.359  & 0.395  &      & 0.321  & 0.338  &      & 0.298  & 0.353  &      & 0.578  & 0.523  \\
     & \multicolumn{1}{r}{720} & & {0.284} & {0.311} &      & 0.328 & 0.337 &      & 0.345  & 0.342  &      & 0.320  &  {0.328} &      & 0.333  & 0.362  &      &  {0.314} & 0.334  &      & 0.365  & 0.359  &      & 0.403  & 0.428  &      & 0.419  & 0.428  &      & 0.414  & 0.410  &      & 0.352  & 0.388  &      & 1.059  & 0.741  \\
     & \multicolumn{1}{r}{avg} & & {0.205} & {0.236} &      & 0.227  & 0.262  &      & 0.244  & 0.270  &      & 0.227  &  {0.255} &      & 0.249  & 0.300  &      &  {0.226} & 0.264  &      & 0.259  & 0.287  &      & 0.309  & 0.360  &      & 0.338  & 0.382  &      & 0.288  & 0.314  &      & 0.271  & 0.334  &      & 0.634  & 0.548  \\
\midrule
\multirow{5}[2]{*}{\rotatebox{90}{$Traffic$}} & \multicolumn{1}{r}{96} & & {0.359} & {0.247} &      &  {0.344} &  {0.236} &      & 0.392  & 0.267  &      & 0.396  & 0.264  &      & 0.410  & 0.282  &      & 0.360  & 0.249  &      & 0.593  & 0.321  &      & 0.587  & 0.366  &      & 0.613  & 0.388  &      & 0.612  & 0.338  &      & 0.607  & 0.392  &      & 0.719  & 0.391  \\
     & \multicolumn{1}{r}{192} & & {0.385} & {0.261} &      &  {0.372} &  {0.249} &      & 0.409  & 0.271  &      & 0.412  & 0.268  &      & 0.423  & 0.287  &      & 0.379  & 0.256  &      & 0.617  & 0.336  &      & 0.604  & 0.373  &      & 0.616  & 0.382  &      & 0.613  & 0.340  &      & 0.621  & 0.399  &      & 0.696  & 0.379  \\
     & \multicolumn{1}{r}{336} & & {0.401} & {0.277} &      &  {0.383} &  {0.257} &      & 0.434  & 0.296  &      & 0.421  & 0.273  &      & 0.436  & 0.296  &      & 0.392  & 0.264  &      & 0.629  & 0.336  &      & 0.621  & 0.383  &      & 0.622  & 0.337  &      & 0.618  & 0.328  &      & 0.622  & 0.396  &      & 0.777  & 0.420  \\
     & \multicolumn{1}{r}{720} & & {0.418} & {0.272} &      &  {0.422} &  {0.280} &      & 0.451  & 0.291  &      & 0.455  & 0.291  &      & 0.466  & 0.315  &      & 0.432  & 0.286  &      & 0.640  & 0.350  &      & 0.626  & 0.382  &      & 0.660  & 0.408  &      & 0.653  & 0.355  &      & 0.632  & 0.396  &      & 0.864  & 0.472  \\
     & \multicolumn{1}{r}{avg} & & {0.391} & {0.264} &      &  {0.380} &  {0.256} &      & 0.422  & 0.281  &      & 0.421  & 0.274  &      & 0.434  & 0.295  &      & 0.391  & 0.264  &      & 0.620  & 0.336  &      & 0.610  & 0.376  &      & 0.628  & 0.379  &      & 0.624  & 0.340  &      & 0.621  & 0.396  &      & 0.764  & 0.416  \\
\midrule
\multirow{5}[2]{*}{\rotatebox{90}{$Electricity$}} & \multicolumn{1}{r}{96} & & {0.113} & {0.200} &      &  {0.126} &  {0.218} &      & 0.137  & 0.233  &      & 0.141  & 0.239  &      & 0.140  & 0.237  &      & 0.129  & 0.222  &      & 0.168  & 0.272  &      & 0.193  & 0.308  &      & 0.201  & 0.317  &      & 0.169  & 0.273  &      & 0.187  & 0.304  &      & 0.274  & 0.368  \\
     & \multicolumn{1}{r}{192} & & {0.131} & {0.219} &      &  {0.144} &  {0.237} &      & 0.152  & 0.247  &      & 0.158  & 0.253  &      & 0.153  & 0.249  &      & 0.157  & 0.240  &      & 0.184  & 0.289  &      & 0.201  & 0.315  &      & 0.222  & 0.334  &      & 0.182  & 0.286  &      & 0.199  & 0.315  &      & 0.296  & 0.386  \\
     & \multicolumn{1}{r}{336} & & {0.159} & {0.234} &      &  {0.162} &  {0.256} &      & 0.169  & 0.267  &      & 0.172  & 0.266  &      & 0.169  & 0.267  &      & 0.163  & 0.259  &      & 0.198  & 0.300  &      & 0.214  & 0.329  &      & 0.231  & 0.338  &      & 0.200  & 0.304  &      & 0.212  & 0.329  &      & 0.300  & 0.394  \\
     & \multicolumn{1}{r}{720} & & {0.170} & {0.258} &      &  {0.192} &  {0.286} &      & 0.200  & 0.290  &      & 0.207  & 0.293  &      & 0.203  & 0.301  &      & 0.197  & 0.290  &      & 0.220  & 0.320  &      & 0.246  & 0.355  &      & 0.254  & 0.361  &      & 0.222  & 0.321  &      & 0.233  & 0.345  &      & 0.373  & 0.439  \\
     & \multicolumn{1}{r}{avg} & & {0.143} & {0.228} &      &  {0.156} &  {0.249} &      & 0.165  & 0.259  &      & 0.170  & 0.263  &      & 0.166  & 0.264  &      & 0.162  & 0.253  &      & 0.193  & 0.295  &      & 0.214  & 0.327  &      & 0.227  & 0.338  &      & 0.193  & 0.296  &      & 0.208  & 0.323  &      & 0.311  & 0.397  \\
\midrule
\bottomrule
\end{tabular}%
}
\end{table*}%

\paragraph{Full-shot forecasting results on the long-term TSF benchmark.}
Table~\ref{tab:task3_full_longtsf_all} demostrates the performance comparison of the full-shot time series forecasting on the Long-term TSF benchmark.

\begin{table*}[htbp]
    \centering
    \small
    \tabcolsep=0.0mm
    \renewcommand\arraystretch{1.7}
    \caption{Full results for the classification task. We report the classification accuracy (\%) as the result. The standard deviation is within 0.1\%.}\label{tab:classification}
    \resizebox{\textwidth}{!}{
    \begin{tabular}{c|ccccccccccccccccccccccccccccccccccc}
    \toprule
    \hline

    \multirow{2}{*}{{Datasets/Models}} & 
    \multicolumn{3}{c}{{Classical methods}} & 
    \multicolumn{3}{c}{{RNN}} &
    \multicolumn{1}{c}{{TCN}} &
    \multicolumn{10}{c}{{Transformers}} & 
    \multicolumn{3}{c}{{MLP}} & 
    \multicolumn{1}{c}{{CNN}}\\
    
    \cmidrule(lr){2-4}\cmidrule(lr){5-7}\cmidrule(lr){8-8}\cmidrule(lr){9-18}\cmidrule(lr){19-21}\cmidrule(lr){22-22} &
    
    \scalebox{0.8}{DTW} & \scalebox{0.8}{XGBoost} & \scalebox{0.8}{Rocket} & \scalebox{0.8}{LSTM} & \scalebox{0.8}{LSTNet} & 
    \scalebox{0.8}{LSSL} & \scalebox{0.8}{TCN} & \scalebox{0.9}{Trans.} & \scalebox{0.9}{Re.} & \scalebox{0.9}{In.} & 
    \scalebox{0.9}{Pyra.} & \scalebox{0.9}{Auto.} & \scalebox{0.9}{Stat.} & \scalebox{0.9}{FED.} & \scalebox{0.9}{ETS.} & 
    \scalebox{0.9}{Flow.} & \scalebox{0.8}{iTrans.} & \scalebox{0.8}{DLinear} & \scalebox{0.8}{LightTS.} & 
    \scalebox{0.8}{TiDE} & \scalebox{0.8}{TimesNet} & { {\myformer}} \\
    
    \hline

	{EthanolConcentration} &  {32.3} &  {43.7} &  {45.2} &  {32.3} &  {39.9} &  {31.1}&   {28.9} &  {32.7} & {31.9} & {31.6}   & {30.8} & {31.6} & {32.7} &  {28.1}& {31.2}  &  {33.8} &  {28.1}&  {32.6} & {29.7} &  {27.1}& {35.7} &  {37.5}\\
  
	{FaceDetection} &  {52.9} &  {63.3} &  {64.7} &  {57.7} &  {65.7} &  {66.7} &  {52.8} &  {67.3} &  {68.6} & {67.0} & {65.7} & {68.4} & {68.0} & {66.0} &  {66.3} &  {67.6} &  {66.3}& {68.0} & {67.5} &  {65.3}&  {68.6} &  {72.1}  \\
  
    {Handwriting} &  {28.6} &  {15.8} &  {58.8} &  {15.2} &  {25.8} &  {24.6} &  {53.3} &  {32.0} &  {27.4} & {32.8} & {29.4} & {36.7} & {31.6} & {28.0} &   {32.5} &  {33.8} &  {24.2}&  {27.0} & {26.1} &  {23.2}&  {32.1} &  {30.5} \\
  
	{Heartbeat} &  {71.7}  &  {73.2} &  {75.6} &  {72.2} &  {77.1} &  {72.7}&  {75.6} &  {76.1} &  {77.1} & {80.5} & {75.6} & {74.6} & {73.7} & {73.7} &   {71.2} &  {77.6} &  {75.6}&  {75.1} & {75.1} &  {74.6}&  {78.0}  &  {78.5}    \\
  
	{JapaneseVowels} &  {94.9} &  {86.5} &  {96.2} &  {79.7} &  {98.1} &  {98.4} &  {98.9} &  {98.7} &  {97.8} & {98.9} & {98.4} & {96.2} & {99.2} & {98.4} &  {95.9} &   {98.9} &  {96.6}&  {96.2} & {96.2} &  {95.6}&  {98.4} &  {99.4}  \\
  
	{PEMS-SF} &  {71.1} &  {98.3} &  {75.1} &  {39.9} &  {86.7} &  {86.1}&  {68.8} &  {82.1} &  {82.7} & {81.5} & {83.2} & {82.7} & {87.3} & {80.9} &  {86.0} &   {83.8} &  {87.9}&  {75.1} & {88.4} &  {86.9}&  {89.6} &  {94.0} \\
  
	{SelfRegulationSCP1} &  {77.7}  &  {84.6} &  {90.8} &  {68.9} &  {84.0} &  {90.8} &  {84.6} &  {92.2} &  {90.4} & {90.1} & {88.1} & {84.0} & {89.4} & {88.7} &  {89.6} &  {92.5} &  {90.2}&  {87.3} & {89.8} &  {89.2}&  {91.8} &  {87.8} \\
  
    {SelfRegulationSCP2} &  {53.9} &  {48.9} &  {53.3} &  {46.6} &  {52.8} &  {52.2} &  {55.6} &  {53.9} &  {56.7} & {53.3} & {53.3} & {50.6} & {57.2} & {54.4} &  {55.0} &   {56.1} &  {54.4}&  {50.5} & {51.1} &  {53.4}&  {57.2} &  {57.0}  \\
     
    {SpokenArabicDigits} &  {96.3} &  {69.6} &  {71.2} &  {31.9} &  {100.0} &  {100.0} &  {95.6} &  {98.4} &  {97.0} & {100.0} & {99.6} & {100.0} & {100.0} & {100.0} &  {100.0} &   {98.8} &  {96.0}&  {81.4} & {100.0} &  {95.0}&  {99.0} &  {100.0} \\
     
    {UWaveGestureLibrary} &  {90.3} &  {75.9} &  {94.4} &  {41.2} &  {87.8} &  {85.9} &  {88.4} &  {85.6} &  {85.6} & {85.6} & {83.4} & {85.9} & {87.5} & {85.3} &  {85.0} &   {86.6} &  {85.9}&  {82.1} & {80.3} &  {84.9}&  {85.3} &  {92.5} \\
    
    \hline
    
    {Average Accuracy} &  {67.0} &  {66.0} &  {72.5} &  {48.6} &  {71.8} &  {70.9} &  {70.3} &  {71.9} &  {71.5} & {72.1} & {70.8} & {71.1} & {72.7} & {70.7} &  {71.0} &{ {73.0}} &  {70.5}&  {67.5} & {70.4} &  {69.5}& {73.6} &  {{75.2}} \\
    
    \hline
    \bottomrule
    \end{tabular}
    } 
\end{table*}

\paragraph{Complete classification results on the UEA archive.}
Table~\ref{tab:classification} presents the comprehensive performance evaluation of TimeArtist across the UEA classification datasets.

\begin{table*}[htpb]
  \vspace{-10pt}
  \caption{To verify the sensitivity of \myformer to hyperparameter selection, we show the performance of \myformer with different parameter scales on multiple benchmarks, with a fixed forecasting window of 96, and a performance metric of MSE.}\label{tab:sensitivity}
  \centering
  \resizebox{1.7\columnwidth}{!}{
  \begin{small}
  \renewcommand{\multirowsetup}{\centering}
  \tabcolsep=0.4cm
  \renewcommand\arraystretch{1.3}
  \begin{tabular}{ccccccccc}
    \toprule
    \hline
    
    \multicolumn{2}{c}{\multirow{1}{*}{\scalebox{1.0}{Hyperparameter}}} & 
    \multicolumn{1}{c}{\rotatebox{0}{\scalebox{1.0}{ETTh1}}} & 
    \multicolumn{1}{c}{\rotatebox{0}{\scalebox{1.0}{ETTh2}}} & 
    \multicolumn{1}{c}{\rotatebox{0}{\scalebox{1.0}{ETTm1}}} & 
    \multicolumn{1}{c}{\rotatebox{0}{\scalebox{1.0}{ETTm2}}} & 
    \multicolumn{1}{c}{\rotatebox{0}{\scalebox{1.0}{Weather}}} & 
    \multicolumn{1}{c}{\rotatebox{0}{\scalebox{1.0}{ECL}}} & 
    \multicolumn{1}{c}{\rotatebox{0}{\scalebox{1.0}{Traffic}}} \\
    \hline

    \multicolumn{1}{c}{\multirow{3}{*}{\rotatebox{0}{N=32}}} &
    \multirow{1}{*}{\scalebox{1.0}{\shortstack{L=3,D=16}}} &
    0.356 & 0.281 & 0.287 & 0.160 & 0.148 & 0.117 & 0.371 \\
    \cline{2-9}

    & \multirow{1}{*}{\scalebox{1.0}{\shortstack{L=4,D=32}}} &
    0.350 & 0.275 & 0.273 & 0.165 & 0.145 & 0.120 & 0.389 \\
    \cline{2-9}
    
    & \multirow{1}{*}{\scalebox{1.0}{\shortstack{L=6,D=128}}} &
    0.352 & 0.285 & 0.279 & 0.163 & 0.152 & 0.126 & 0.393 \\
    \hline

    \multicolumn{1}{c}{\multirow{3}{*}{\rotatebox{0}{N=64}}} &
    \multirow{1}{*}{\scalebox{1.0}{\shortstack{L=3,D=16}}} &
    0.334 & {0.246} & 0.269 & 0.158 & 0.140 & 0.118 & 0.361 \\
    \cline{2-9}
    
    & \multirow{1}{*}{\scalebox{1.0}{\shortstack{L=4,D=32}}} &
    0.339 & 0.260 & 0.263 & 0.152 & 0.138 & 0.113 & 0.359 \\
    \cline{2-9}

    & \multirow{1}{*}{\scalebox{1.0}{\shortstack{L=6,D=128}}} &
    0.340 & 0.273 & 0.282 & 0.159 & 0.146 & 0.119 & 0.370 \\
    \hline
    
    \multicolumn{1}{c}{\multirow{3}{*}{\rotatebox{0}{N=128}}} &
    \multirow{1}{*}{\scalebox{1.0}{\shortstack{L=3,D=16}}} &
     {0.337} & 0.254 & {0.268} & 0.152 & 0.136 & 0.112 & 0.356 \\
    \cline{2-9}
    
    & \multirow{1}{*}{\scalebox{1.0}{\shortstack{L=4,D=32}}} &
    0.346 & 0.289 & 0.275 & 0.157 & 0.148 & 0.122 & 0.365 \\
    \cline{2-9}

    & \multirow{1}{*}{\scalebox{1.0}{\shortstack{L=6,D=128}}} &
    0.344 & 0.285 & 0.272 & 0.162 & 0.150 & 0.119 & 0.373 \\
    \hline
    
    \bottomrule
  \end{tabular}
  \end{small}
}
\end{table*}

\begin{table*}[htpb]
  \caption{The comparison of the memory usage and time consumption of the proposed \myformer and other baselines during training (full fine-tuning) on the ETTh1-96 task, in the full-shot forecasting scenario.}\label{tab:full_computational}
  \centering
  \resizebox{1.2\columnwidth}{!}{
  \begin{small}
  \renewcommand{\multirowsetup}{\centering}
  \tabcolsep=0.3cm
  \renewcommand\arraystretch{1.2}
  \begin{tabular}{c|ccc}
    \toprule
    \hline
    
    {Metrics(ETTh1-96)} &
    {Total Param. (M)} &
    {Mem. (MiB)} &
    {Speed (s/iter)} \\
    \hline

    UniTime	& 439.52 & 2074 & 0.335 \\
    TimeLLM & 3623.71 & 4537 & 0.184 \\
    \myformer & 313.79 & 1809 & 0.270 \\

    \hline
    \bottomrule
  \end{tabular}
  \end{small}
}
\end{table*}

\section{E. Hyperparameter Sensitivity and Ablation}
In the concrete implementation of \myformer, both architectural hyperparameters of the model and vector quantization hyperparameters are specified. 
The former encompasses structural configurations of the dual-autoencoder and alignment components, while the latter defines architectural details of the shared-quantizer module. 
Regarding the model architecture, we denote the number of transformer layers in the encoder, decoder, and alignment components as $L$, and represent the dimensionality of the latent space as $D$. 
For the quantization operation, we define each image or time series sample being quantized into $N$ distinct code vectors.

In order to verify the sensitivity of \myformer to hyperparameter selection, the Table~\ref{tab:sensitivity} shows the performance of \myformer with different parameter scales on full-shot forecasting task. 
Where the model parameter combinations include \begin{small}$(L,D) \textit{=} (3,16), (4,32), (6,128)$\end{small} and \begin{small}$N \textit{=} 32, 64, 128$\end{small}, with a fixed forecasting window of $96$, and a performance metric of MSE. 
By default, the hyperparameters of \myformer are fixed to $L\textit{=}4$, $D\textit{=}32$ and $N\textit{=}64$, all experimental results presented follow this setting. Furthermore, in the table~\ref{tab:full_computational}, we also present the comparison of the memory usage and time consumption during training (full fine-tuning) of the proposed \myformer and other baseline models in the ETTh1-96 task. This can serve as a reference for the computational complexity of model fine-tuning in the full-shot forecasting scenario.

\begin{table*}[htpb]
  \centering
  \caption{To investigate the impact of predefined label-category pairs on temporal-visual alignment training in classification tasks. We conducted an ablation experiment on image category pairs and demonstrated the accuracy rate on the UEA dataset.}\label{tab:ablation_tsc}
  \small
    \begin{tabular}{ccccc}
    \toprule
    \textbf{category pairs} & \textbf{FaceDetection(\%)} & \textbf{Heartbeat(\%)} & \textbf{SelfRegulationSCP1(\%)} & \textbf{SelfRegulationSCP2(\%)} \\
    \midrule
    Persian Cat $vs.$ School Bus & 71.9 & 79.3 & 88.1 & 56.8 \\
    Golden Retriever $vs.$ Siamese Cat & 72.1 & 78.5 & 87.8 & 57.0 \\
    Gray Wolf $vs.$ Siberian Husky & 73.4 & 78.8 & 87.6 & 57.2 \\
    Coffee Cup $vs.$ Mug & 72.8 & 79.1 & 87.3 & 57.0 \\
    Elephant $vs.$ Ant & 71.2 & 78.5 & 87.9 & 58.9 \\
    \bottomrule
    \end{tabular}
\end{table*}%

In time-series classification, we establish alignment relationships between temporal and visual samples based on predefined label-category pairs. In this section, we design five groups of image category pairs derived from ImageNet-1k, encompassing category pairs with varying difficulty levels (from extremely easy to distinguish to relatively difficult) and covering diverse semantic hierarchies (animals, man-made objects, natural objects, etc.). 
Table~\ref{tab:ablation_tsc} presents our ablation study on image category pair selection conducted on the UAE benchmark. Comprehensive experiments demonstrate that different visual semantic disparities do not significantly impact classification performance, verifying that the model can effectively learn distinguishable visual semantic differences during the alignment phase.

\section{F. Visualization}~\label{sec:visual}
\begin{figure*}[ht]
\begin{center}
\centerline{\includegraphics[width=1.9\columnwidth]{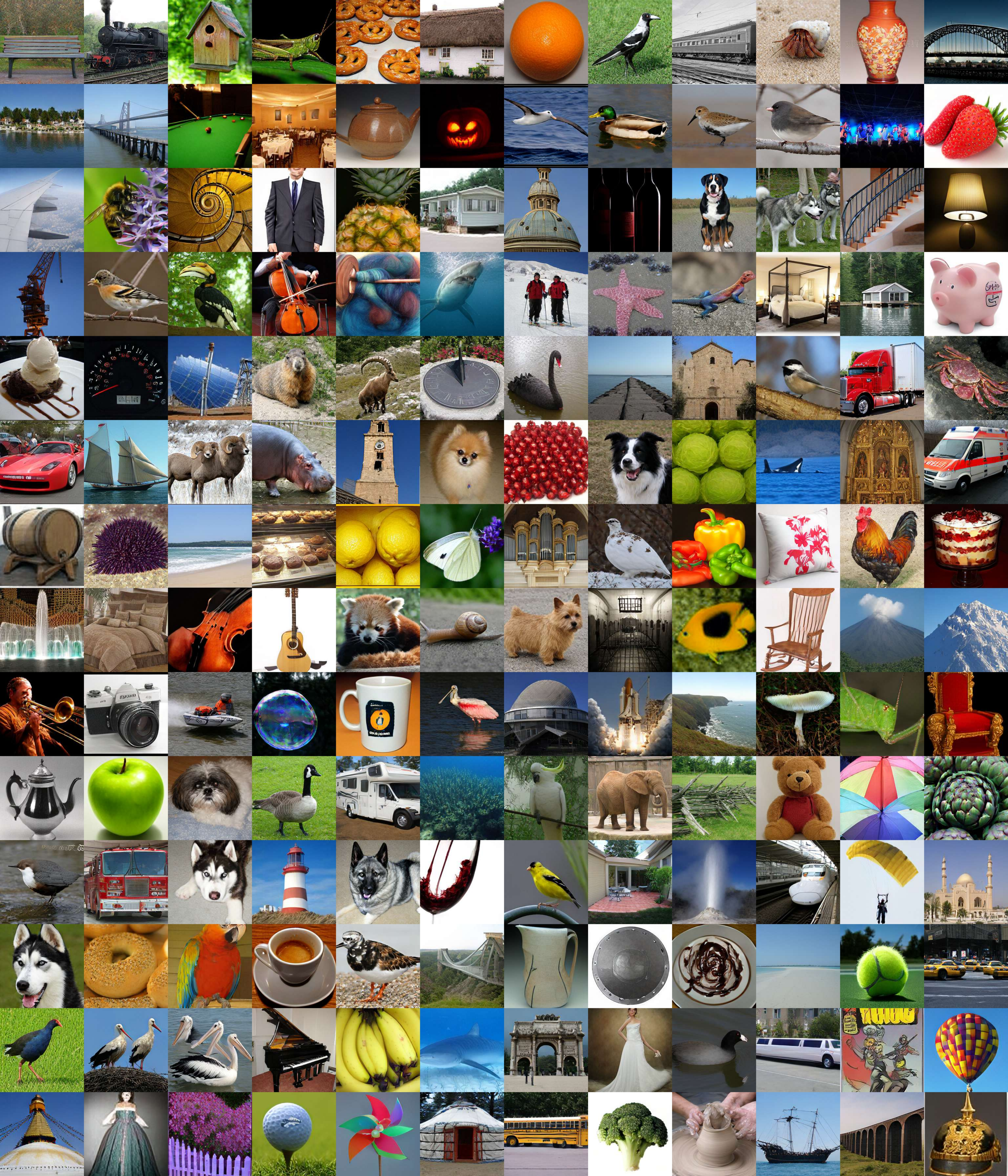}}
\vspace{20pt}
\caption{
Visualization of stitched images generated from time series.
}\label{fig:concat_1}
\end{center}
\vspace{-10pt}
\end{figure*}

\begin{figure*}[ht]
\begin{center}
\centerline{\includegraphics[width=1.9\columnwidth]{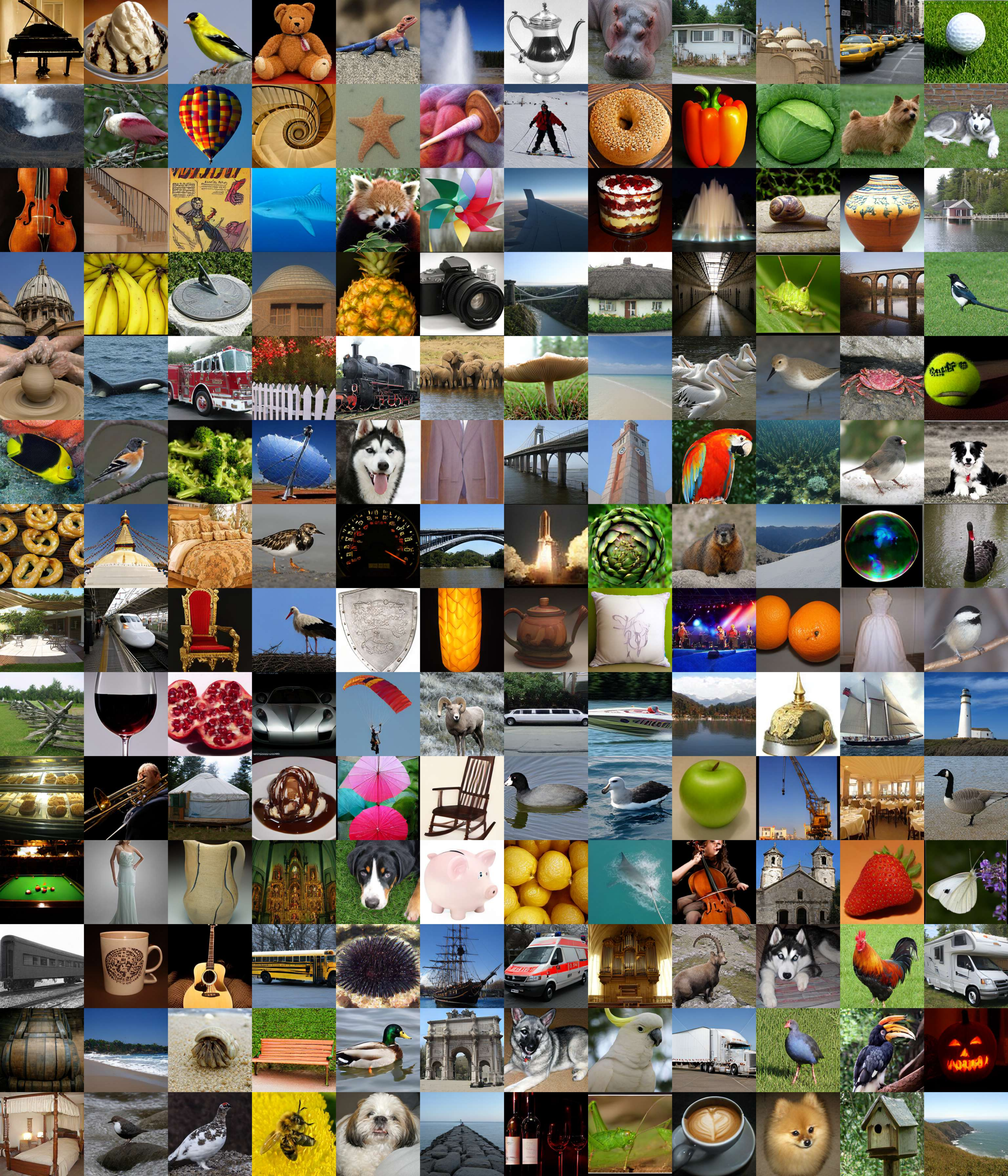}}
\vspace{20pt}
\caption{
Visualization of stitched images generated from time series.
}\label{fig:concat_2}
\end{center}
\vspace{-10pt}
\end{figure*}

\vspace{-10pt}
\paragraph{Convert the time series into an image.}
In the figure~\ref{fig:concat_1} and \ref{fig:concat_2}, we present the images generated from the time series.

\begin{figure*}[ht]
\begin{center}
	\centerline{\includegraphics[width=1.7\columnwidth]{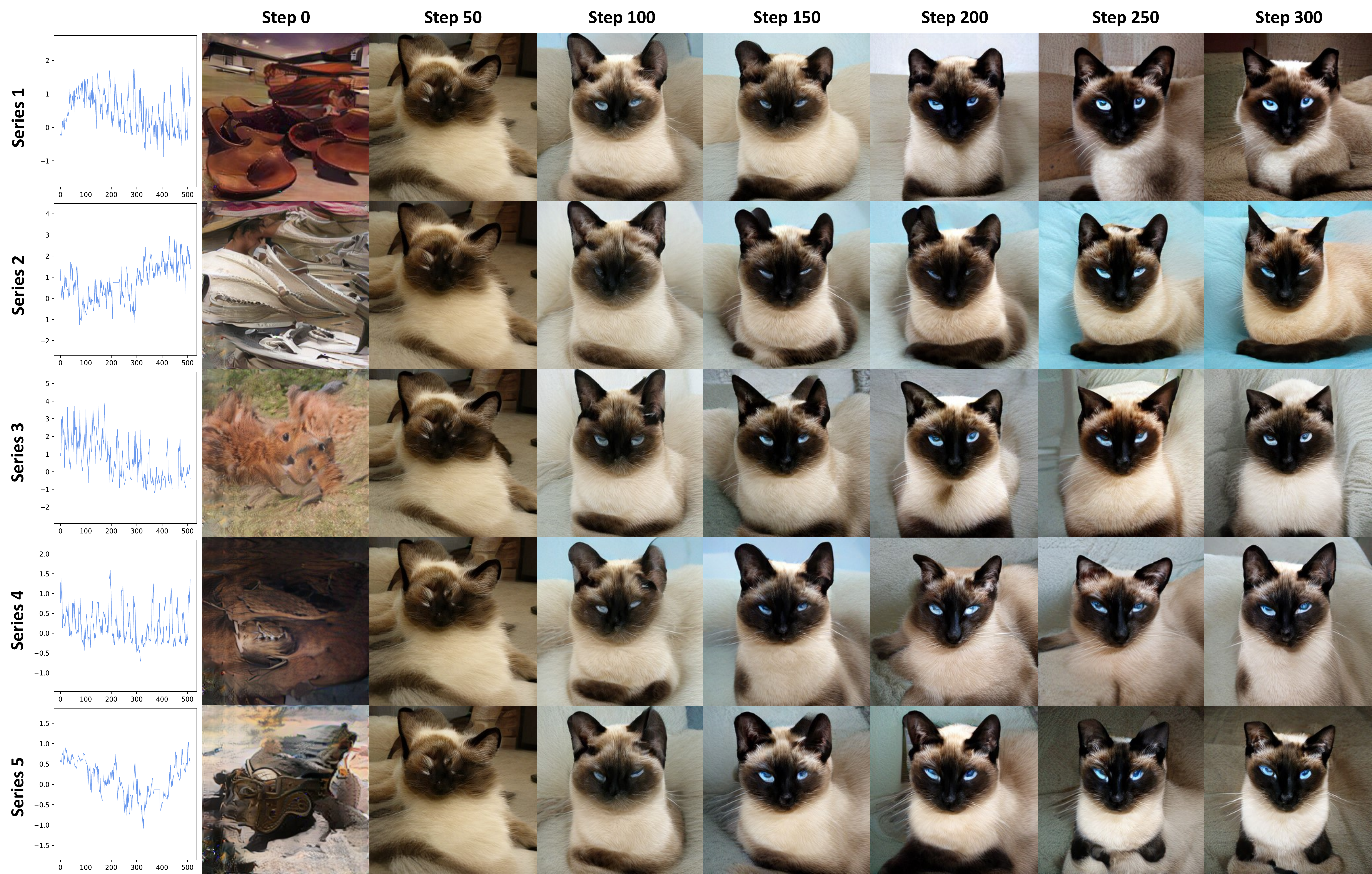}}
        \vspace{20pt}
	\centerline{\includegraphics[width=1.7\columnwidth]{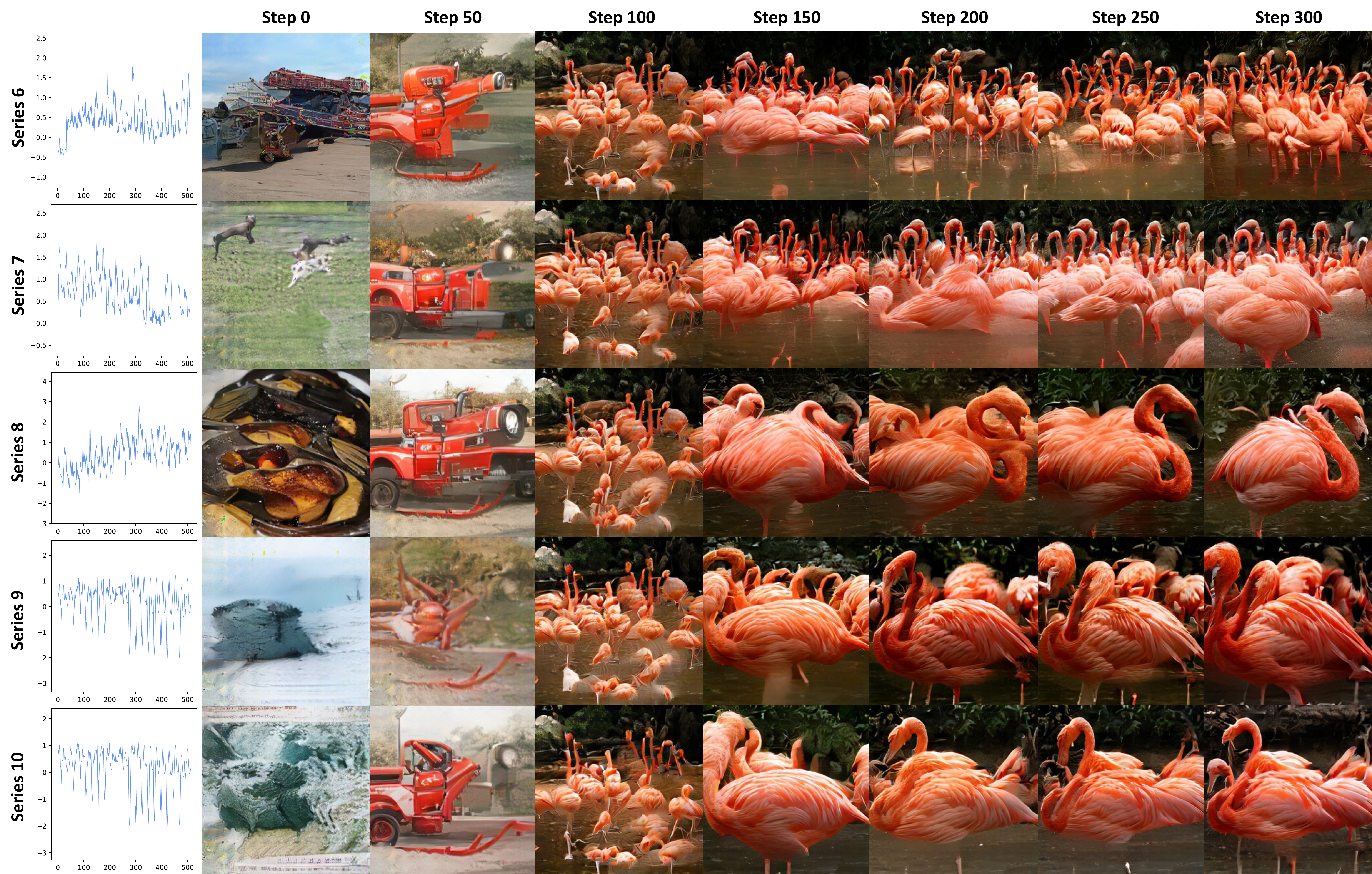}}
        \vspace{20pt}
	\caption{
As the training steps increases, \myformer initially generates standardized semantic representations—images conforming to universal cognitive patterns of objects like cats or dogs. 
Subsequently, it progressively learns fine-grained mapping relationships between individual time series and specific images, thereby producing diversified visual outputs.
}\label{fig:task1_1}
\end{center}
\vspace{-10pt}
\end{figure*}

\begin{figure*}[ht]
\begin{center}
	\centerline{\includegraphics[width=1.7\columnwidth]{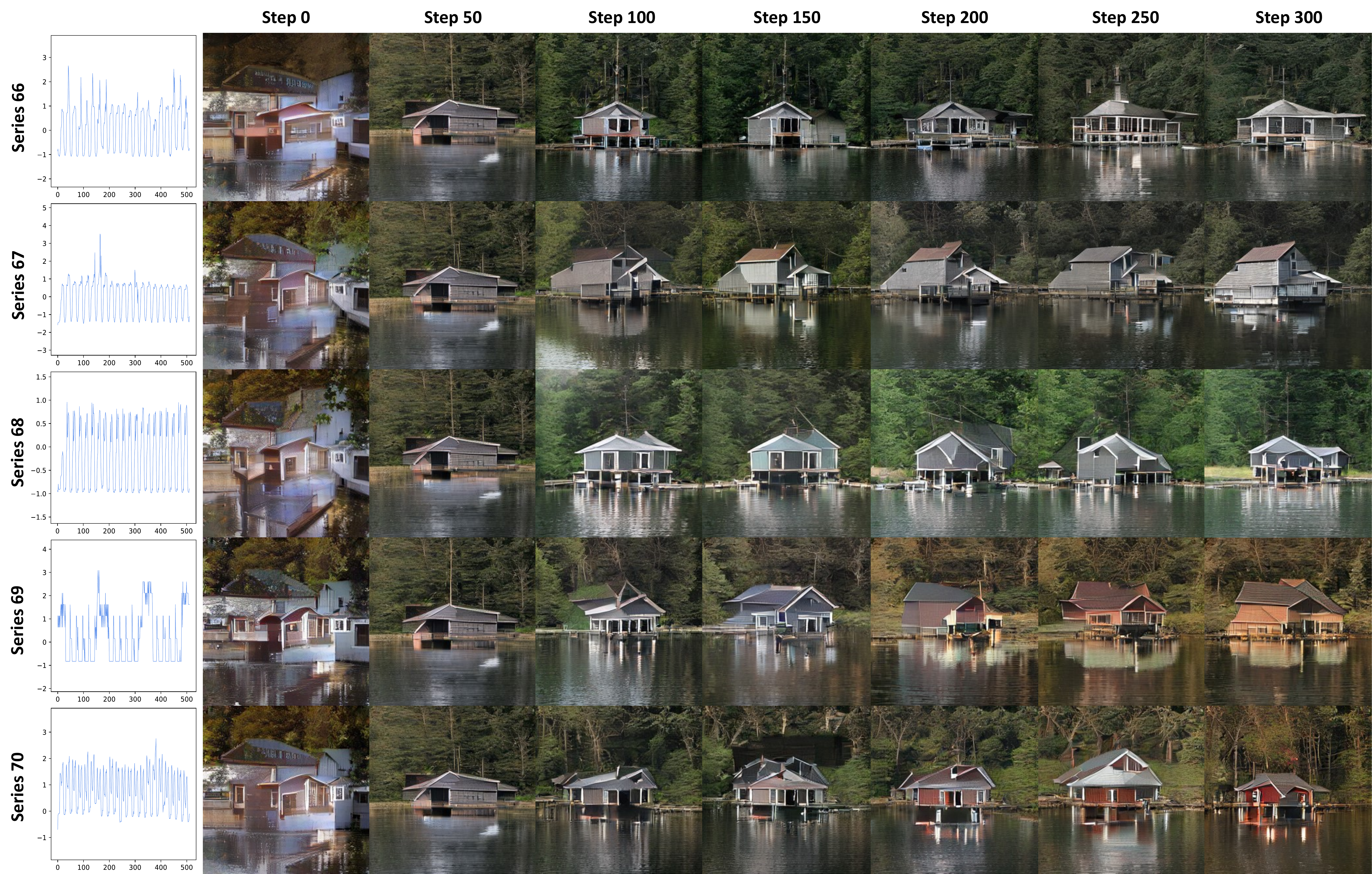}}
        \vspace{20pt}
	\centerline{\includegraphics[width=1.7\columnwidth]{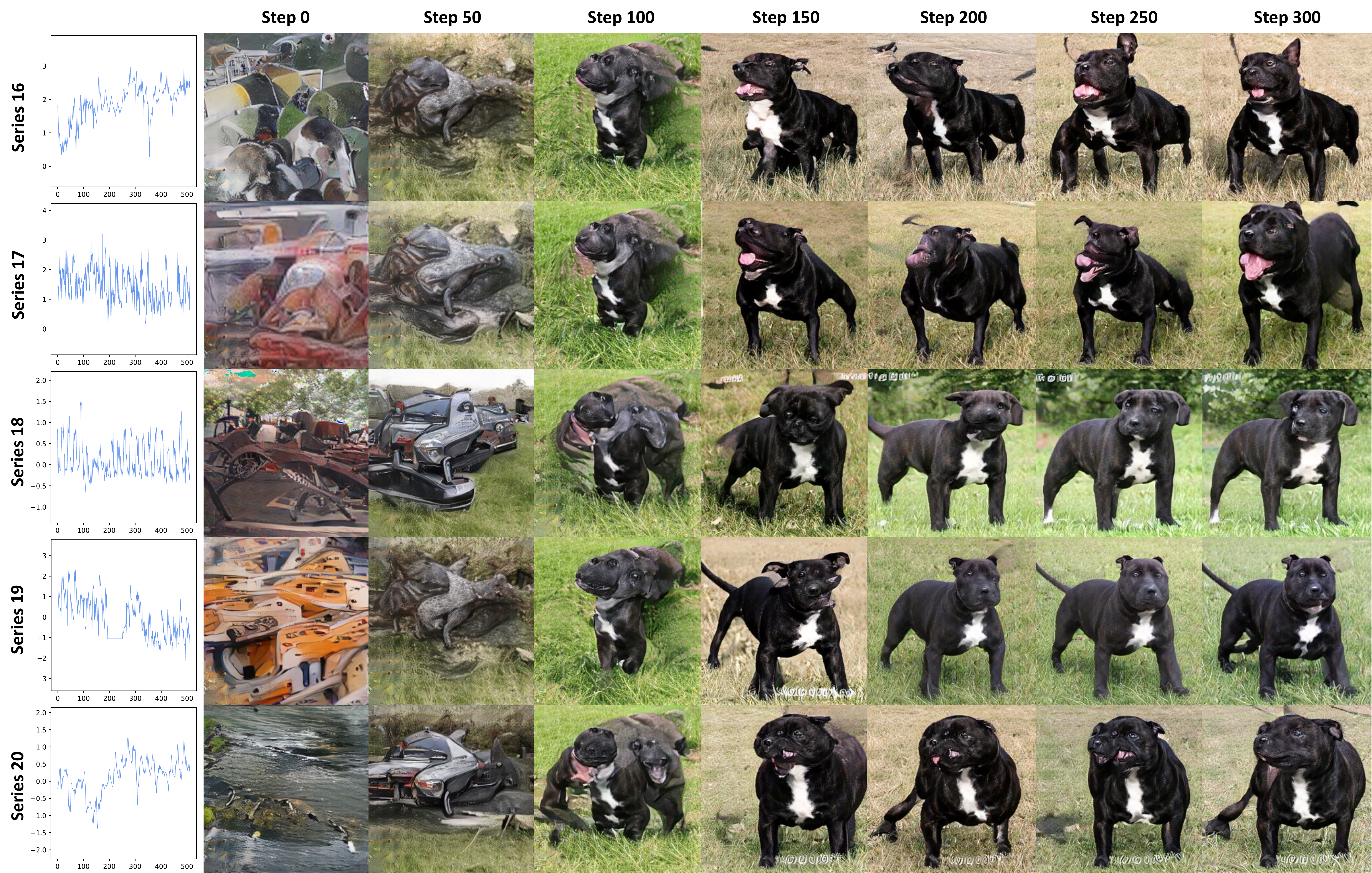}}
        \vspace{20pt}
	\caption{
As the training steps increases, \myformer initially generates standardized semantic representations—images conforming to universal cognitive patterns of objects like cats or dogs. 
Subsequently, it progressively learns fine-grained mapping relationships between individual time series and specific images, thereby producing diversified visual outputs.
}\label{fig:task1_2}
\end{center}
\vspace{-10pt}
\end{figure*}

\begin{figure*}[ht]
\begin{center}
	\centerline{\includegraphics[width=1.7\columnwidth]{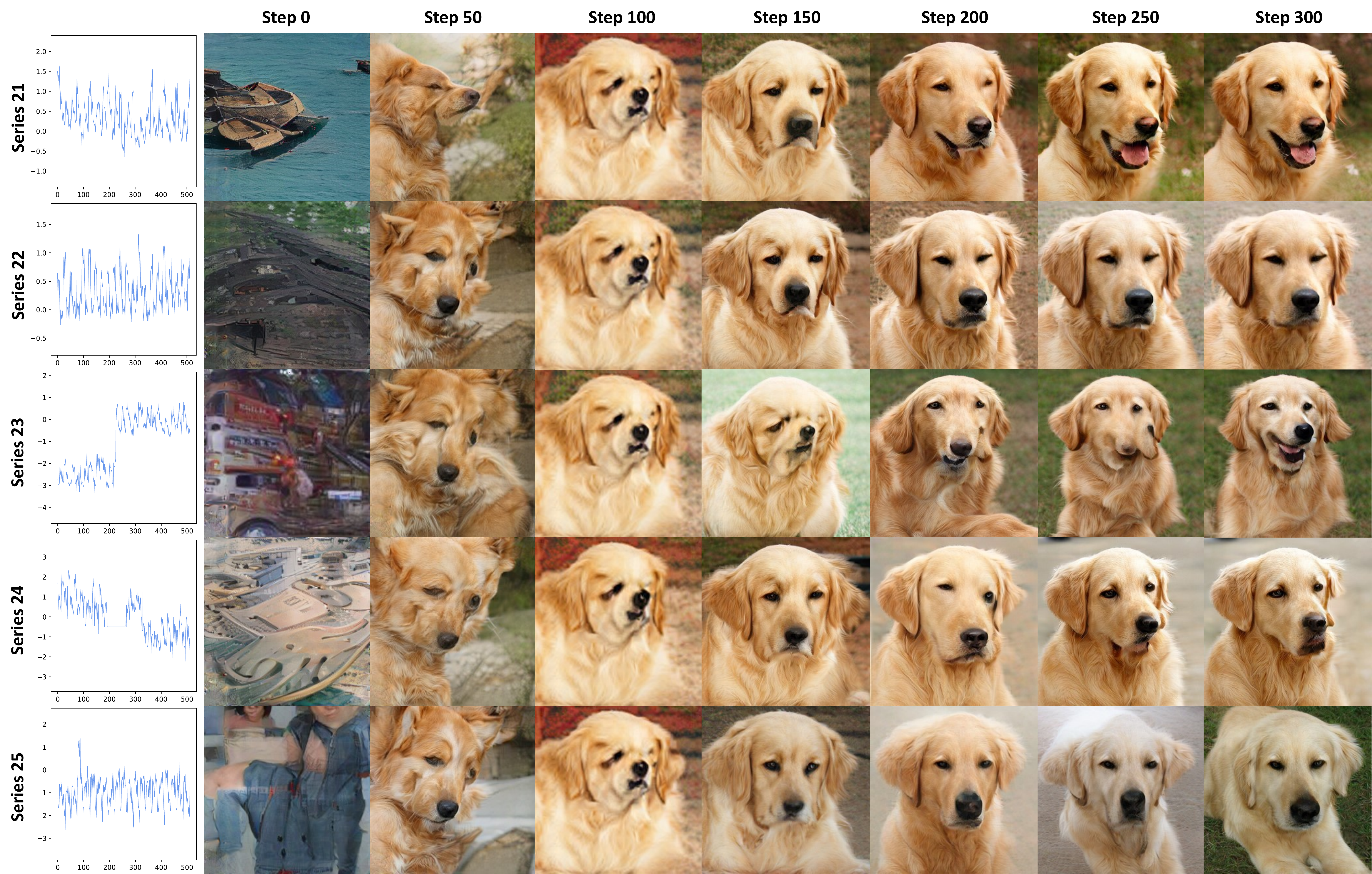}}
        \vspace{20pt}
	\centerline{\includegraphics[width=1.7\columnwidth]{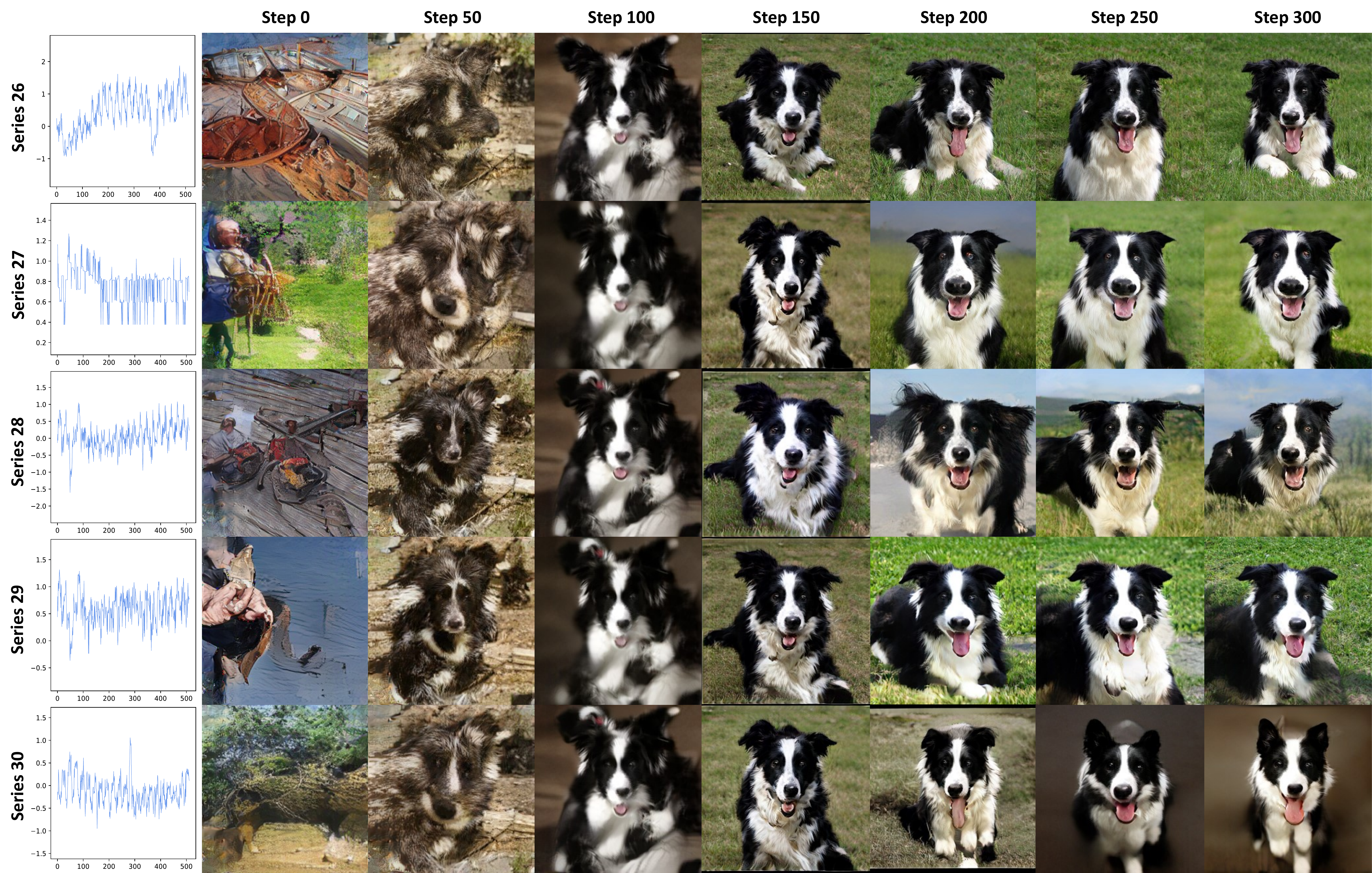}}
        \vspace{20pt}
	\caption{
As the training steps increases, \myformer initially generates standardized semantic representations—images conforming to universal cognitive patterns of objects like cats or dogs. 
Subsequently, it progressively learns fine-grained mapping relationships between individual time series and specific images, thereby producing diversified visual outputs.
}\label{fig:task1_3}
\end{center}
\vspace{-10pt}
\end{figure*}

\begin{figure*}[ht]
\begin{center}
	\centerline{\includegraphics[width=1.7\columnwidth]{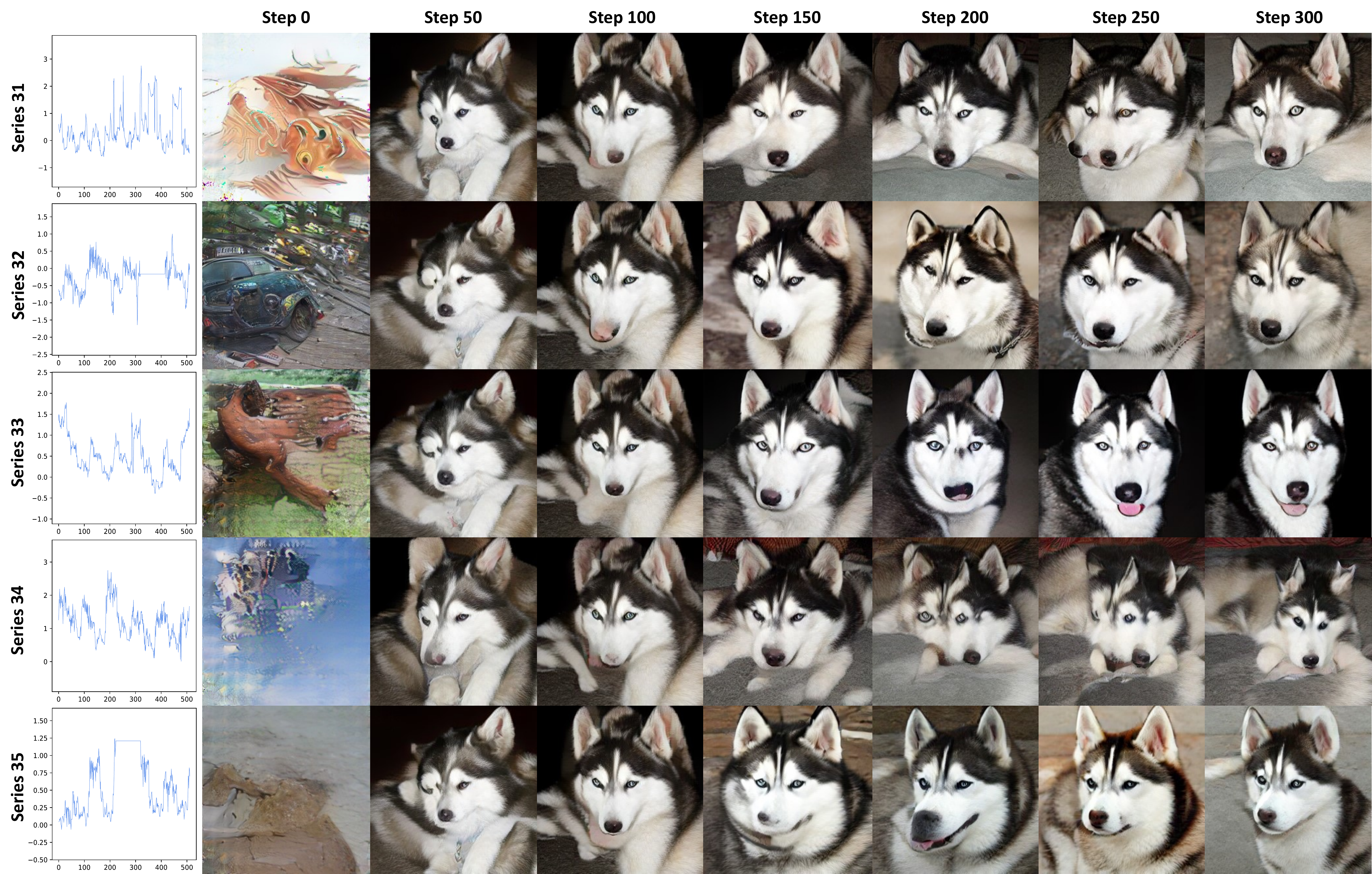}}
        \vspace{20pt}
	\centerline{\includegraphics[width=1.7\columnwidth]{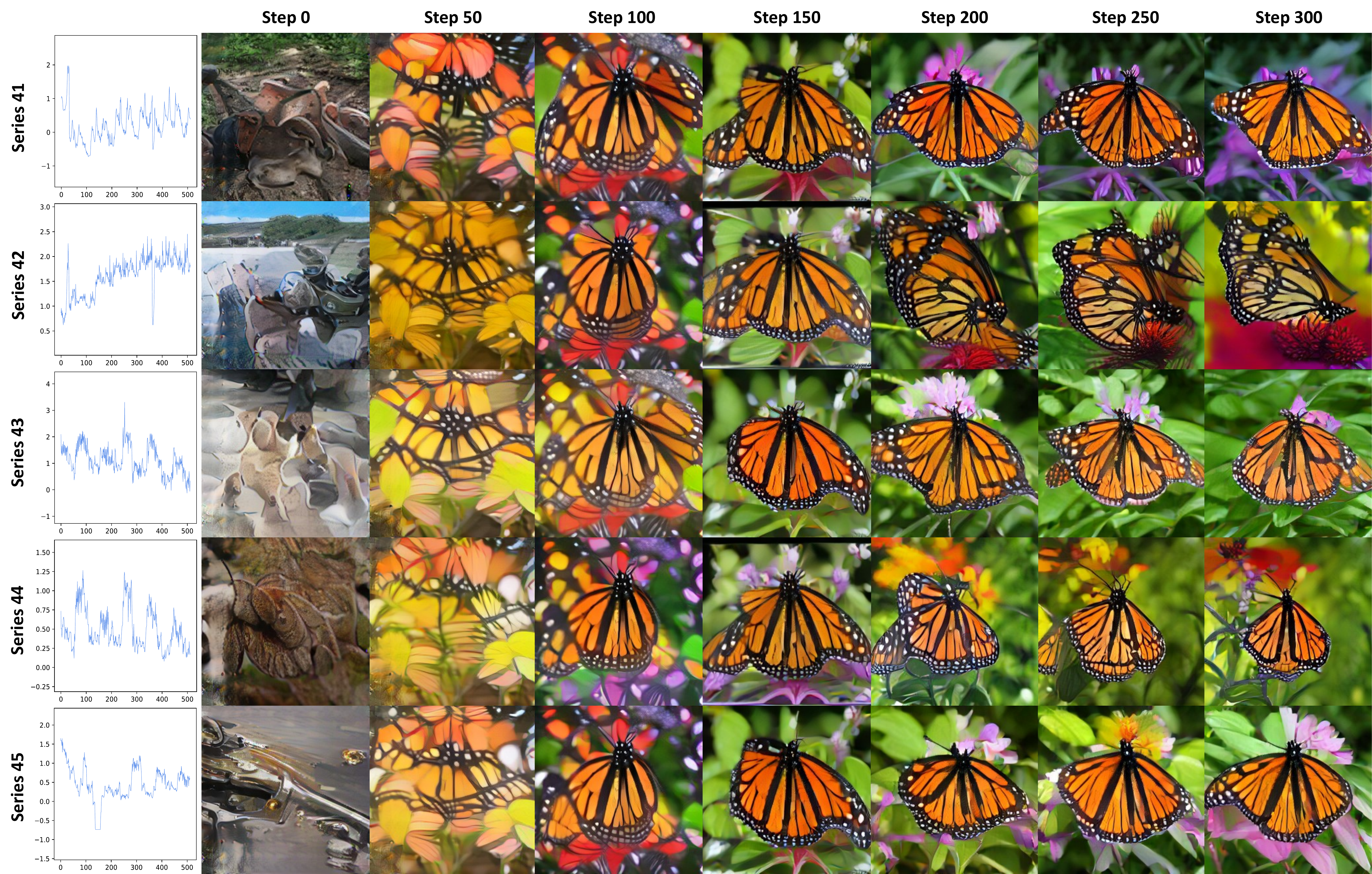}}
        \vspace{20pt}
	\caption{
As the training steps increases, \myformer initially generates standardized semantic representations—images conforming to universal cognitive patterns of objects like cats or dogs. 
Subsequently, it progressively learns fine-grained mapping relationships between individual time series and specific images, thereby producing diversified visual outputs.
}\label{fig:task1_4}
\end{center}
\vspace{-10pt}
\end{figure*}

\paragraph{Time series gradually transforms into images.} Figure~\ref{fig:task1_1},~\ref{fig:task1_2},~\ref{fig:task1_3}, and~\ref{fig:task1_4} illustrate the visualizations of images generated from time series throughout the progressive increase of training steps. 
Here, Stage 0 represents direct image decoding from time series representation without employing the alignment module. 
At earlier stages (e.g., Stage 50), \myformer demonstrates a tendency to learn standardized semantic representations across all images sharing the same label, resulting in nearly identical projections of different time series onto singular objects (e.g., either dogs or cats). 
As training progresses to later stages, the model gradually achieves fine-grained cross-modal alignment, enabling diversified projections where distinct input time series are mapped to varied and specific images.

\paragraph{Temporal-rendered image style transfer.}
After establishing the alignment at the representation level, we naturally explore whether temporal fluctuation patterns can be integrated into real images within the unified latent space, such as merging the dubbing of a movie into a poster:
\begin{small}
\begin{equation}\label{eq5}
  \begin{split}
    x_{recon}^{image}\textit{=}VisuDec(Q&uant(TempEnc(x^{time})) \\
    \textit{+}Q&uant(VisuEnc(x^{image}))).
  \end{split}
\end{equation}
\end{small}
Benefiting from large-scale temporal-visual alignment training, \myformer synthesizes novel temporally-styled images under zero-shot conditions, by capturing rich temporal information from real time series. 
As illustrated in Figure, long-term trends are transformed into lighting or shadow patterns in synthesized images, while periodic fluctuations are converted into textural striations. 
Without requiring finetuning, \myformer utilize ubiquitous real signals to generate artistically styled images through learned representations.

\begin{figure*}[ht]
\begin{center}
	\centerline{\includegraphics[width=2.0\columnwidth]{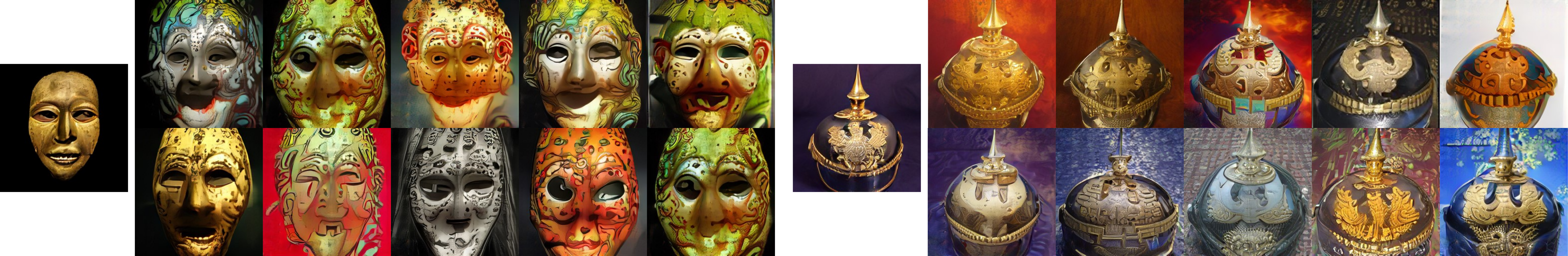}}
        \vspace{10pt}
	\centerline{\includegraphics[width=2.0\columnwidth]{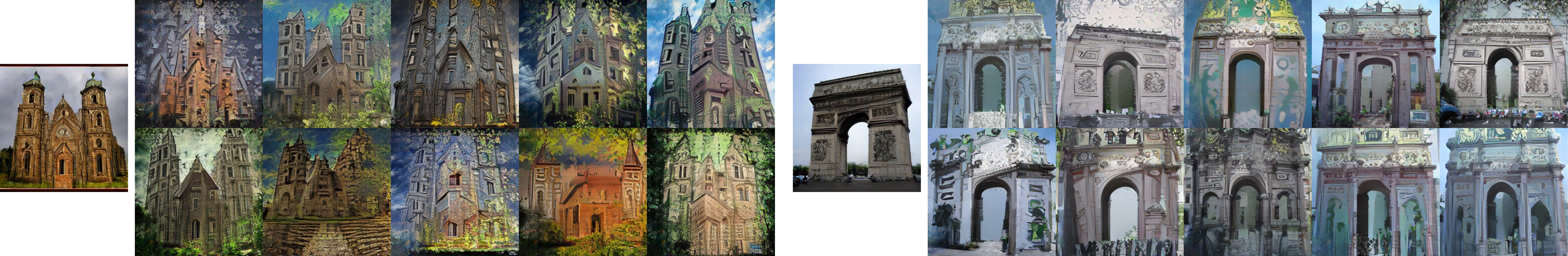}}
        \vspace{10pt}
        \centerline{\includegraphics[width=2.0\columnwidth]{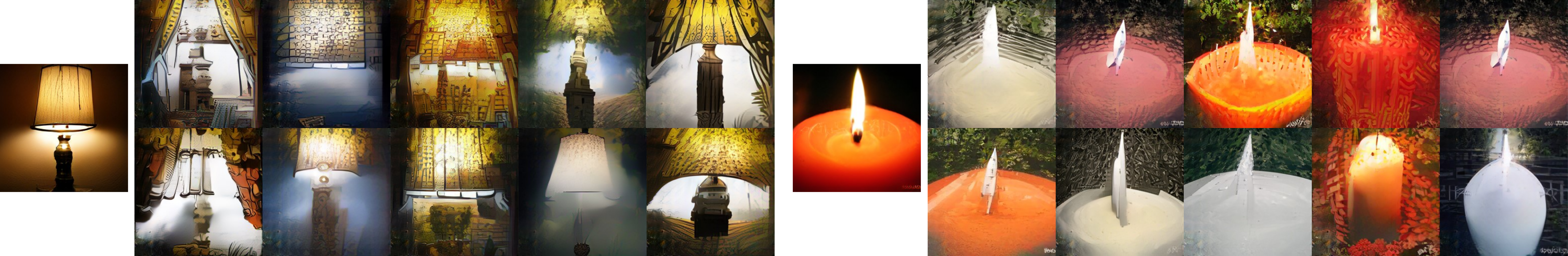}}
        \vspace{10pt}
	\centerline{\includegraphics[width=2.0\columnwidth]{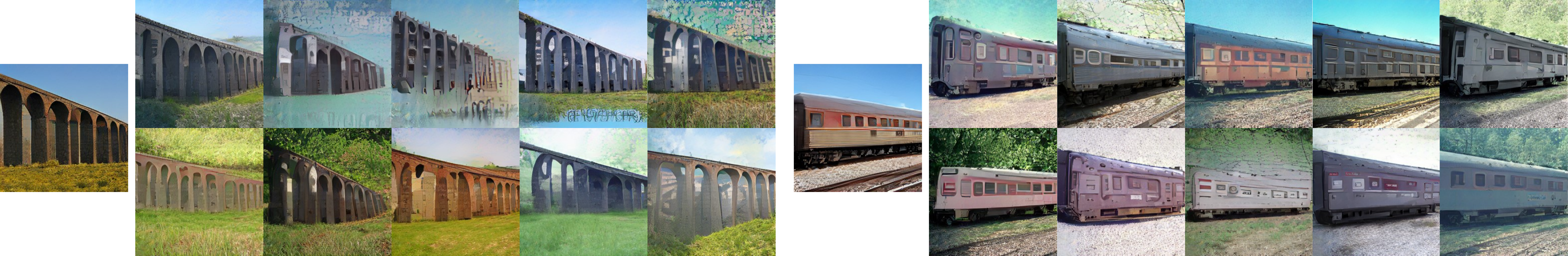}}
        \vspace{10pt}
        \centerline{\includegraphics[width=2.0\columnwidth]{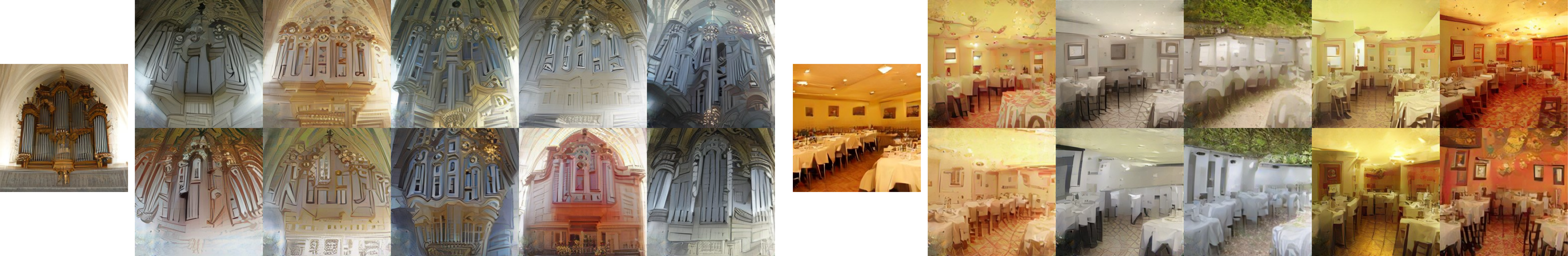}}
        \vspace{10pt}
	\centerline{\includegraphics[width=2.0\columnwidth]{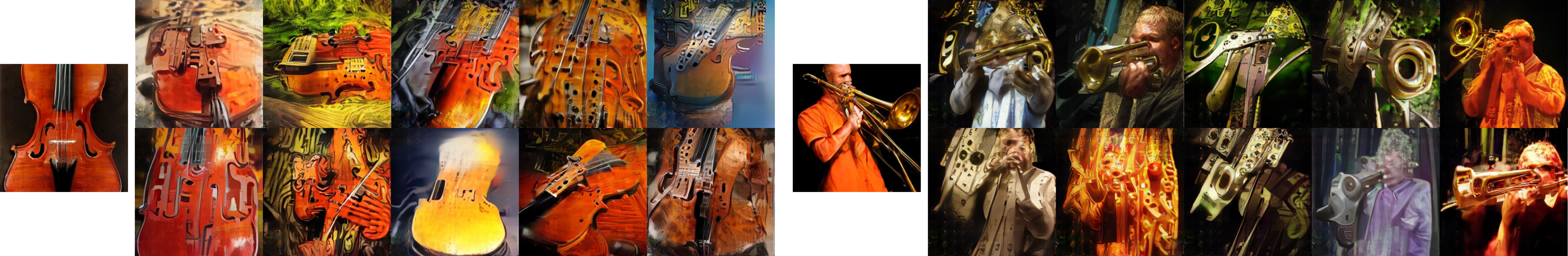}}
        \vspace{10pt}
	\caption{
The visualization results of time-series-based image style transfer are demostrated. 
For each subplot, the leftmost position displays the original image, followed by 10 synthesized images incorporating temporal representations.
}\label{fig:task2_1}
\end{center}
\vspace{-10pt}
\end{figure*}

\begin{figure*}[ht]
\begin{center}
	\centerline{\includegraphics[width=2.0\columnwidth]{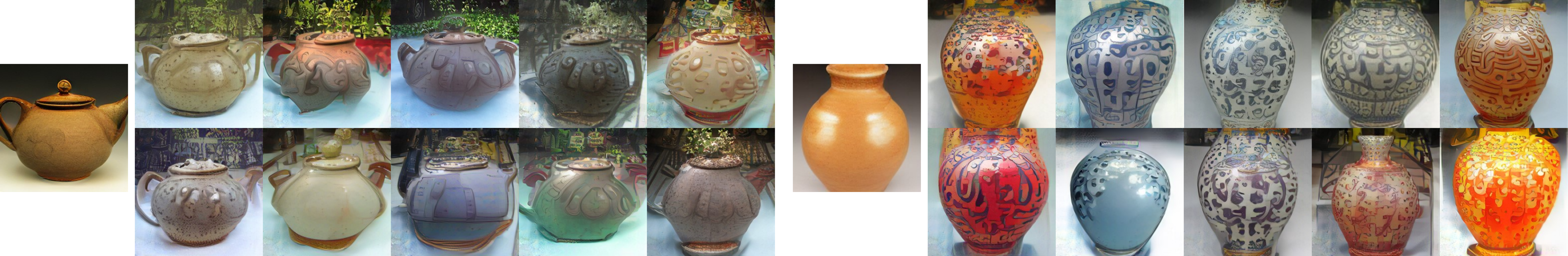}}
        \vspace{10pt}
	\centerline{\includegraphics[width=2.0\columnwidth]{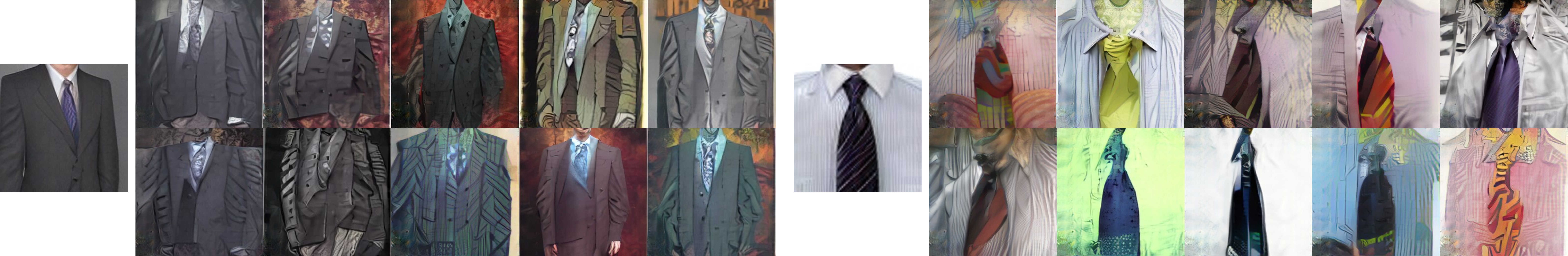}}
        \vspace{10pt}
        \centerline{\includegraphics[width=2.0\columnwidth]{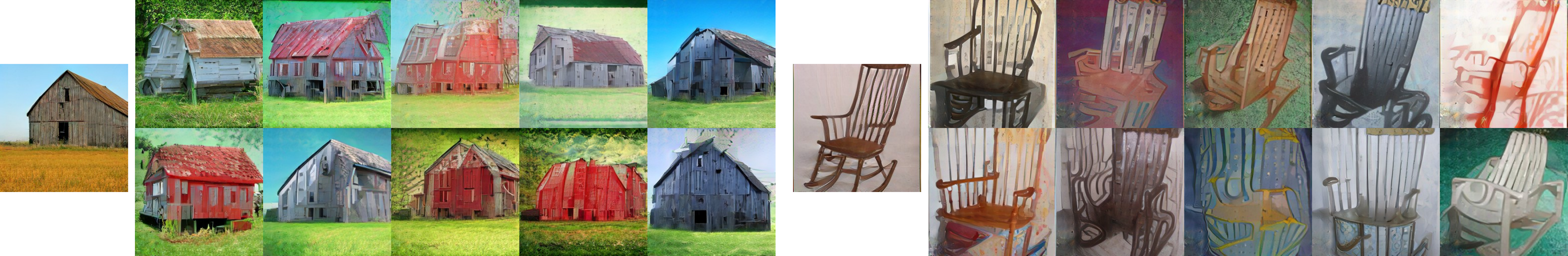}}
        \vspace{10pt}
	\centerline{\includegraphics[width=2.0\columnwidth]{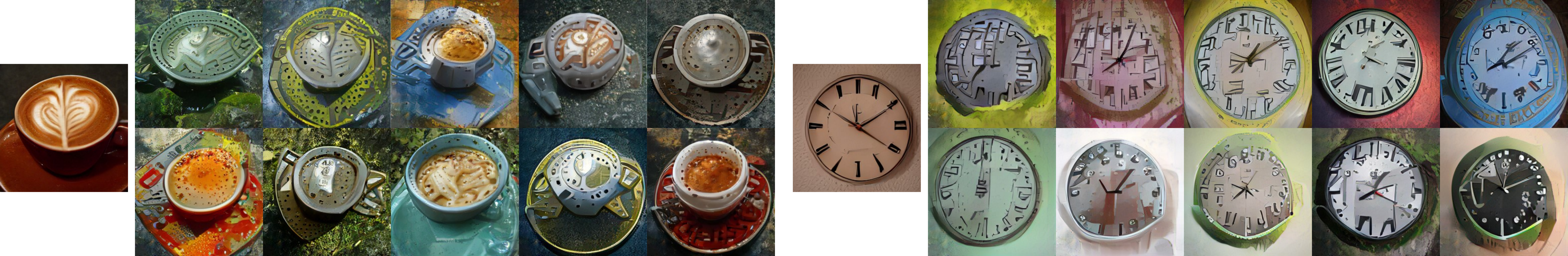}}
        \vspace{10pt}
        \centerline{\includegraphics[width=2.0\columnwidth]{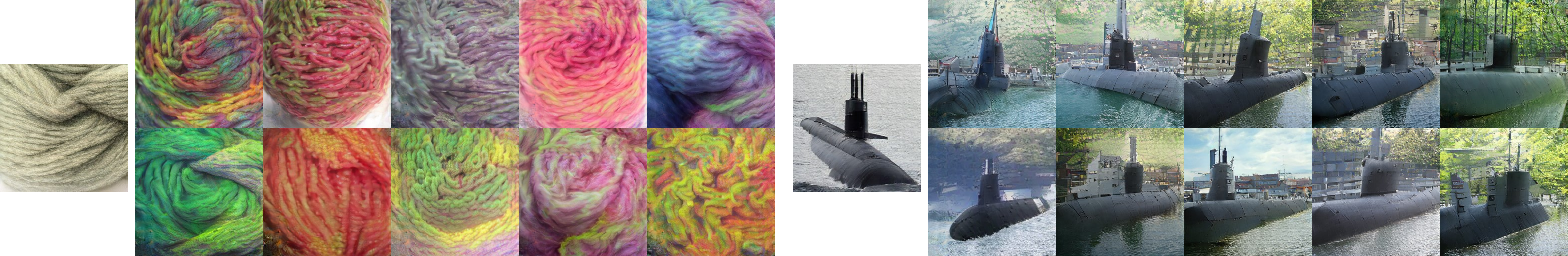}}
        \vspace{10pt}
	\centerline{\includegraphics[width=2.0\columnwidth]{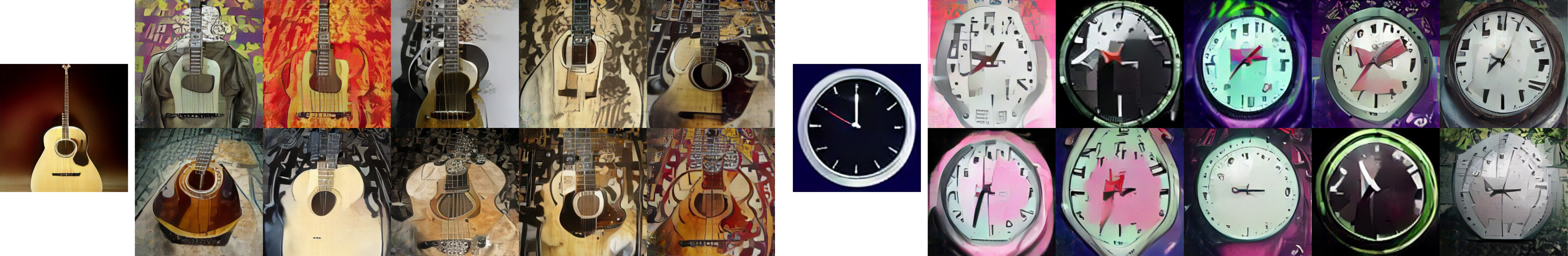}}
        \vspace{10pt}
	\caption{
The visualization results of time-series-based image style transfer are demostrated. 
For each subplot, the leftmost position displays the original image, followed by 10 synthesized images incorporating temporal representations.
}\label{fig:task2_2}
\end{center}
\vspace{-10pt}
\end{figure*}

\begin{figure*}[ht]
\begin{center}
	\centerline{\includegraphics[width=2.0\columnwidth]{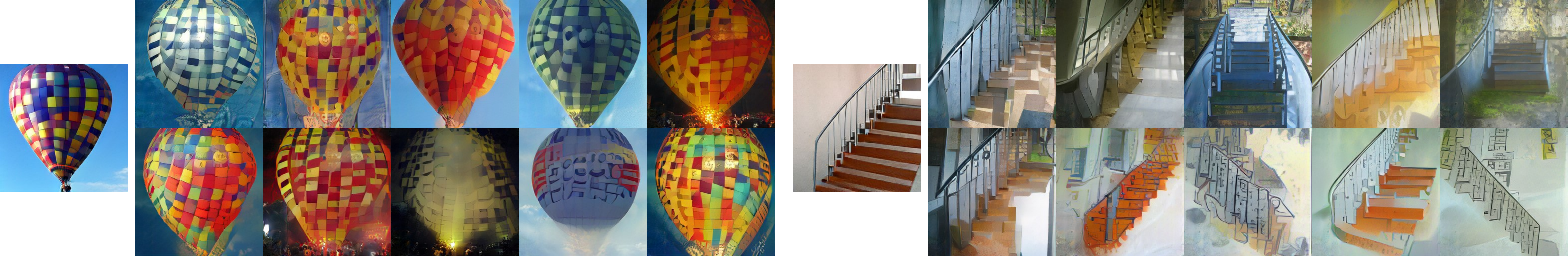}}
        \vspace{10pt}
	\centerline{\includegraphics[width=2.0\columnwidth]{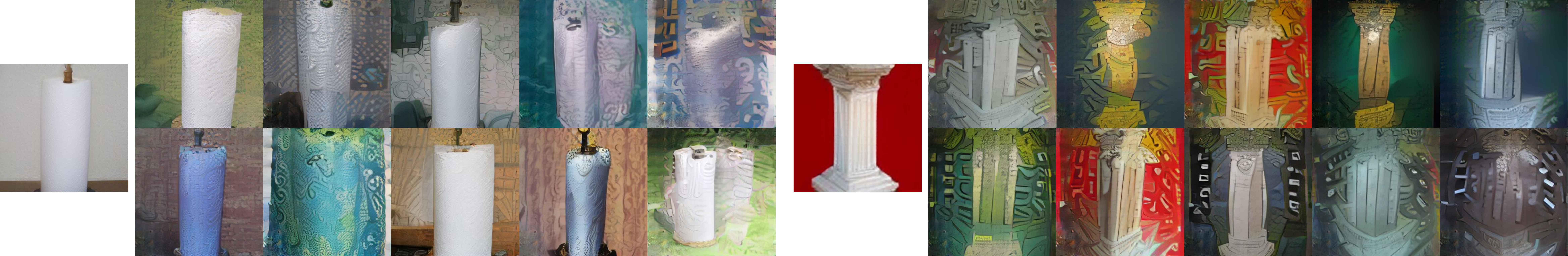}}
        \vspace{10pt}
        \centerline{\includegraphics[width=2.0\columnwidth]{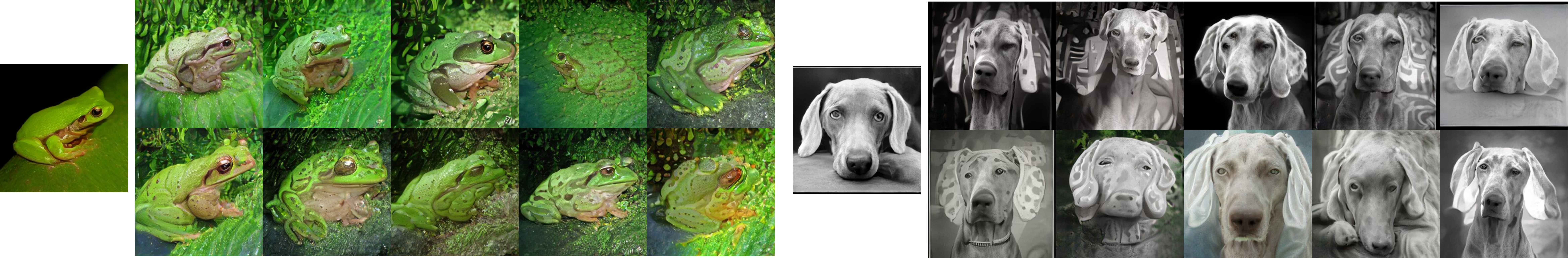}}
        \vspace{10pt}
	\centerline{\includegraphics[width=2.0\columnwidth]{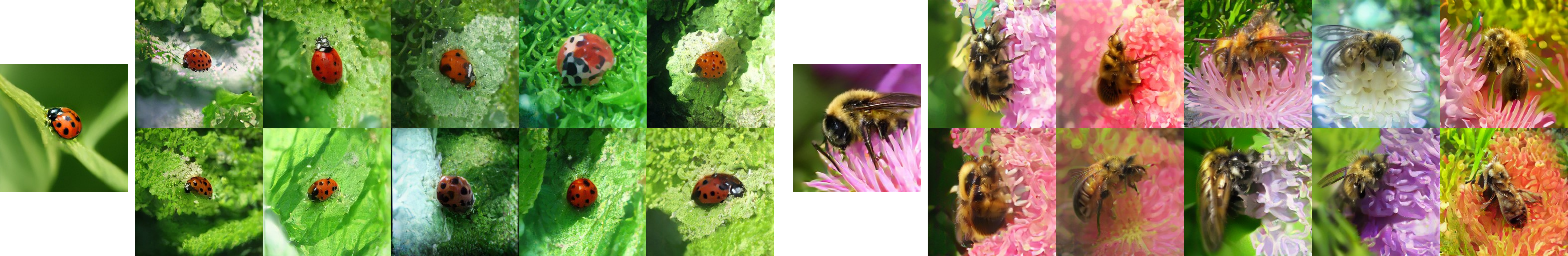}}
        \vspace{10pt}
        \centerline{\includegraphics[width=2.0\columnwidth]{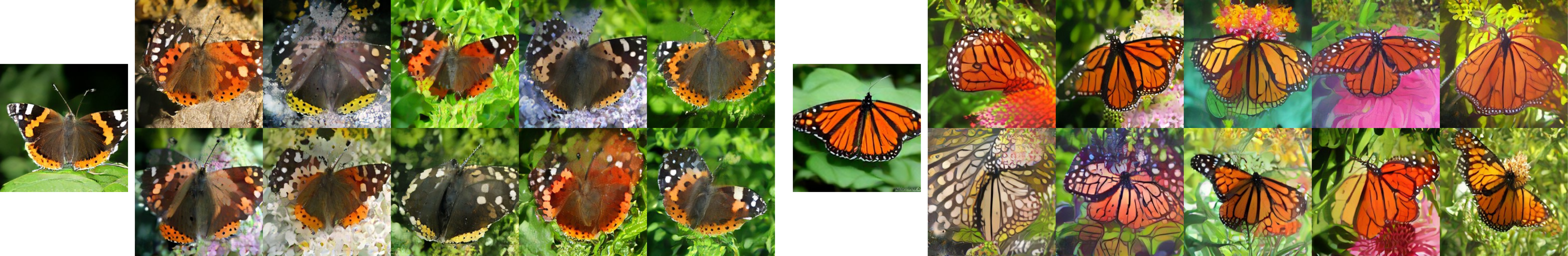}}
        \vspace{10pt}
	\caption{
The visualization results of time-series-based image style transfer are demostrated. 
For each subplot, the leftmost position displays the original image, followed by 10 synthesized images incorporating temporal representations.
}\label{fig:task2_3}
\end{center}
\vspace{-10pt}
\end{figure*}

Through estbalishing a unified discrete representation space, we can fuse real images with arbitrary time series in the latent space, then decode the fused representation into synthesized images. 
By capturing trends, fluctuation patterns from ubiquitous real-world time series, TimeArtist can efficiently generates novel temporal-style images, as shown in figure~\ref{fig:task2_1}, \ref{fig:task2_2} and \ref{fig:task2_3}.

\end{document}